\documentclass[11pt,letterpaper]{article}

\usepackage[utf8]{inputenc}
\usepackage[T1]{fontenc}
\usepackage[english]{babel}
\usepackage[hyphens]{url}
\usepackage{authblk}
\usepackage{booktabs}
\usepackage{hyperref}
\usepackage{amsfonts}
\usepackage{nicefrac}
\usepackage{microtype}
\usepackage{amsmath, amsthm, amssymb}
\usepackage{fancybox}
\usepackage{graphicx}
\usepackage{float}
\usepackage{algorithm}
\usepackage{algpseudocode}
\usepackage{color}
\usepackage{tikz}
\usepackage{array}
\usepackage{subcaption}
\usepackage[textsize=scriptsize]{todonotes}
\usepackage{enumitem}
\usepackage{wrapfig}
\usepackage{enumitem}
\usepackage{etoolbox}
\usepackage{diagbox}
\usepackage{comment}
\usepackage{makecell}
\usepackage{multirow}
\usepackage{multicol}
\usepackage[in]{fullpage}

\usepackage{mathpazo}

\newtheoremstyle{slplain}
  {.4\baselineskip\@plus.1\baselineskip\@minus.1\baselineskip}
  {.3\baselineskip\@plus.1\baselineskip\@minus.1\baselineskip}
  {\itshape}
  {}
  {\bfseries}
  {.}
  { }
  {}
\theoremstyle{slplain} 

\newtheorem*{definition*}{Definition}
\newtheorem*{theorem*}{Theorem}

\makeatletter
\newtheorem*{rep@theorem}{\rep@title}
\newcommand{\newreptheorem}[2]{%
\newenvironment{rep#1}[1]{%
 \def\rep@title{#2 \ref{##1}}%
 \begin{rep@theorem}}%
 {\end{rep@theorem}}}
\makeatother

\newreptheorem{theorem}{Theorem}
\newreptheorem{lemma}{Lemma}
\newreptheorem{claim}{Claim}
\newreptheorem{corollary}{Corollary}
\newreptheorem{proposition}{Proposition}

\theoremstyle{definition}

\theoremstyle{plain} 

\numberwithin{equation}{section}

\newtheoremstyle{etplain}
  {.0\baselineskip\@plus.1\baselineskip\@minus.1\baselineskip}
  {.0\baselineskip\@plus.1\baselineskip\@minus.1\baselineskip}
  {\itshape}
  {}
  {\bfseries}
  {.}
  { }
  {}

\newcommand{\E}{\mathbb{E}}
\newcommand{\R}{\mathbb{R}}

\newcommand{\norm}[1]{\left|\left| #1 \right|\right|}
\newcommand{\Prob}{\mathbb{P}}

\renewcommand\bar\overline

\DeclareMathOperator*{\argmax}{arg\,max}
\DeclareMathOperator*{\argmin}{arg\,min}

\newcolumntype{C}[1]{>{\centering\let\newline\\\arraybackslash\hspace{0pt}}m{#1}}

\newcommand{\mbimagesfig}[3]{
    \begin{figure}
        \centering
        \begin{subfigure}{0.48\textwidth}
            \centering
            \includegraphics[width=0.9\textwidth]{#1/train.png}
            \caption{Domain $A$ images.}
        \end{subfigure}\quad
        \begin{subfigure}{0.48\textwidth}
            \centering
            \includegraphics[width=0.9\textwidth]{#1/val.png}
            \caption{Domain $B$ images.}
        \end{subfigure}
        
        \begin{subfigure}{0.48\textwidth}
            \centering
            \includegraphics[width=0.9\textwidth]{#1/model_based.png}
            \caption{Domain $A$ images from (a) passed through a learned model of natural variation $G(x,\delta)$.}
        \end{subfigure}\quad
        \begin{subfigure}{0.48\textwidth}
            \centering
            \includegraphics[width=0.9\textwidth]{#1/sampled.png}
            \caption{One image $x$ from domain $A$ (left) and six images generated by passing $x$ through $G(x,\delta)$ for randomly sampled $\delta$ vectors.}
        \end{subfigure}
        \caption[#3]{#2}
    \end{figure}
}

\newcommand{\newdataset}[3]{
    \begin{figure}
        \centering
        \begin{subfigure}{0.47\textwidth}
            \centering
            \includegraphics[width=\textwidth]{#1/orig.png}
            \caption{Test images from ImageNet.}
        \end{subfigure}\quad
        \begin{subfigure}{0.47\textwidth}
            \centering
            \includegraphics[width=\textwidth]{#1/combo.png}
            \caption{Corresponding images from new dataset.}
        \end{subfigure}
        \caption[#3]{#2}
    \end{figure}
}

\usepackage[nottoc,notlot,notlof]{tocbibind}
\usepackage[page,toc,titletoc,title]{appendix}

\title{\vspace{-1cm}Model-Based Robust Deep Learning: Generalizing to Natural, Out-of-Distribution Data}

\author{Alexander Robey}
\author{Hamed Hassani}
\author{George J. Pappas}

\affil{Department of Electrical and Systems Engineering \\ University of Pennsylvania}

\date{}

\begin{document}

\maketitle

\vspace{-1cm}
\begin{abstract}

While deep learning has resulted in major breakthroughs in many application domains,  the frameworks commonly used in deep learning remain fragile to artificially-crafted, imperceptible changes in the data.   In response to this fragility, adversarial training has emerged as a principled approach for enhancing the robustness of deep learning with respect to norm-bounded perturbations.  However, there are other sources of fragility for deep learning that are arguably more common and less thoroughly studied.  Indeed, natural variation such as changes in lighting or weather conditions can significantly degrade the accuracy of trained neural networks, proving that such natural variation presents a significant challenge for deep learning.

In this paper, we propose a paradigm shift from perturbation-based adversarial robustness to  {\em model-based robust deep learning}.  Our objective is to provide general training algorithms that can be used to train deep neural networks to be robust against natural variation in data.  Critical to our paradigm is first obtaining a \emph{model of natural variation} which can be used to vary data over a range of natural conditions.  Such models of natural variation may be either known a priori or else learned from data. In the latter case, we show that deep generative models can be used to learn models of natural variation that are consistent with realistic conditions.  We then exploit such models in three novel model-based robust training algorithms in order to enhance the robustness of deep learning with respect to the given model.

Our extensive experiments show that across a variety of naturally-occurring conditions in twelve distinct datasets including MNIST, SVHN, GTSRB, and CURE-TSR, ImageNet, and ImageNet-c, deep neural networks trained with our model-based algorithms significantly outperform classifiers trained via empirical risk minimization, perturbation-based adversarial training, data augmentation methods, and domain adaptation techniques.   Specifically, when training on ImageNet and testing on various subsets of ImageNet-c, our algorithms improve over baseline methods by up to 30 percentage points in top-1 accuracy.  Furthermore, we demonstrate the broad applicability of our paradigm in four challenging robustness applications. (1) Firstly, we show that our algorithms can significantly improve robustness against natural, out-of-distribution data, resulting in accuracy improvements as large as 20-30 percentage points compared to state-of-the-art classifiers.  (2) Secondly, we show that models of natural variation can be effortlessly composed to provide robustness against multiple simultaneous distributional shifts.  To evaluate this property, we curate several new datasets containing multiple sources of natural variation.  (3) Thirdly, we show that models of natural variation trained on one dataset can be applied to new datasets to provide significant levels of robustness against unseen distributional shifts.  (4) Finally, we show that in the setting of unsupervised domain adaptation, our algorithms outperform traditional domain adaptation techniques.

Our results suggest that exploiting models of natural variation can result in  significant improvements in the robustness of deep learning when deployed in natural environments.  This paves the way for a plethora of interesting future research directions, both algorithmic and theoretical, as well as numerous applications in which enhancing the robustness of deep learning will enable it's wider adoption.

Code is available at the following link: \url{https://github.com/arobey1/mbrdl}.

\end{abstract}

\tableofcontents
\listoffigures
\listoftables
\listofalgorithms

\newpage

\section{Introduction}\label{sect:intro}

Over the last decade, we have witnessed unprecedented breakthroughs in deep learning~\cite{lecun2015deep}.  Rapidly growing bodies of work continue to improve the state-of-the-art in generative modeling \cite{zhu2017unpaired,brock2018large,huang2018multimodal}, computer vision \cite{sabour2017dynamic,jaderberg2015spatial,esteves2017polar}, and natural language processing \cite{devlin2018bert,bahdanau2014neural}.  Indeed, the significant progress made in these fields has prompted large-scale integration of deep learning techniques into a myriad of application domains, including autonomous vehicles, medical diagnostics, and robotics \cite{ribeiro2016should,esteva2019guide}.  Importantly, many of these domains are \emph{safety-critical}, meaning that the detections, recommendations, or decisions made by deep learning systems can directly impact the well-being of humans \cite{oakden2020hidden}.  For this reason, it is essential that the deep learning systems used in safety-critical applications are robust and trustworthy \cite{dreossi2019compositional}.

Despite the remarkable progress made toward improving the state-of-the-art in deep learning, it is well-known that many deep learning frameworks including neural networks are fragile to seemingly innocuous and imperceptible changes to their input data \cite{szegedy2013intriguing}.  Well-documented examples of fragility to carefully-designed noise can be found in the context of image detection \cite{hendrycks2019benchmarking}, video analysis \cite{wei2019sparse, shankar2019systematic}, traffic sign misclassification \cite{eykholt2018robust}, machine translation \cite{wallace2020imitation}, clinical trials \cite{papangelou2018toward}, and robotics \cite{melis2017deep}.  In response to this vulnerability, a growing body of work has focused on improving the robustness of deep learning.  More specifically, the literature concerning \emph{adversarial robustness} has sought to improve robustness against small, imperceptible perturbations of data, which have been shown to cause misclassification \cite{szegedy2013intriguing}.  Over the last five years, this literature has included the development of robust training algorithms \cite{madry2017towards,wong2017provable,madaan2019adversarial,prakash2018deflecting,zhang2019theoretically,kurakin2016adversarial,moosavi2016deepfool} and certifiable defenses \cite{raghunathan2018certified,fazlyab2019safety,katz2019marabou}.   In particular, these robust training approaches, i.e.\ the method of \emph{adversarial training} \cite{goodfellow2014explaining}, typically perturb input data via adversarially-chosen, norm-bounded noise in a robust optimization formulation \cite{madry2017towards, wong2017provable}, and have been shown to be effective at improving the robustness of deep learning against norm-bounded perturbations \cite{dong2019benchmarking}. 

While the adversarial training paradigm has provided a rigorous framework for analyzing and improving the robustness of deep learning, the algorithms used in this paradigm have notable limitations.  Specifically, most adversarial training algorithms are only applicable for robustness applications in which data is perturbed by norm-bounded, artificially-generated, imperceptible noise.  Thus while adversarial training algorithms can resolve security threats arising from artificial tampering of the data, these schemes  cannot provide similar levels of robustness to changes that may arise due to other kinds of perturbations or variation \cite{hendrycks2019benchmarking}.  And to this end, numerous recent papers have unanimously shown that deep learning is extremely fragile to unbounded shifts in the data-distribution which commonly occur due to a wide range of \textit{natural phenomena} \cite{djolonga2020robustness,taori2020measuring,hendrycks2020many,hendrycks2019benchmarking} and which cannot be modeled by additive, norm-bounded perturbations.  Such phenomena include unseen distributional shifts such as changes in image lighting, variable weather conditions, or blurring \cite{pei2017deepxplore, chernikova2019self}.   And while such unseen distributional shifts are arguably more common in safety-critical domains than norm-bounded perturbations, there are remarkably few general, principled techniques that provide robustness against these forms of out-of-distribution, naturally-occurring variation \cite{hendrycks2019augmix}.  Thus, it is of critical importance for the deep learning community to design novel algorithms that are robust against natural, out-of-distribution shifts in data.

In this paper, we formulate the first general-purpose algorithms that (1) use unlabeled data to learn models that describe arbitrary forms of natural variation and (2) exploit these models to provide significant robustness against natural, out-of-distribution shifts in data.  To this end, we propose a paradigm shift from perturbation-based adversarial robustness to {\em model-based robust deep learning}.  In this paradigm, following the observation that data can vary in highly nonlinear and unbounded ways in real-world, safety-critical environments, we first obtain \textit{models of natural variation} which describe how data varies in natural environments.  Such models of natural variation may be known \textit{a priori}, as is the case for geometric transformations such as rotation or scaling.  Alternatively, in some settings a model of natural variation may not be known beforehand and therefore must be learned from data; for example, there are no analytic models that describe how to change the weather conditions in images.  Once such models of natural variation have been obtained, in this paradigm, we formulate a novel robust optimization problem that exploits models of natural variation to produce neural networks that are robust to the source of natural variation captured by the model.  In this way, the goal of the model-based robust paradigm is to develop general-purpose algorithms that can be used to train neural networks to be robust against natural, out-of-distribution shifts in data.  

Our experiments show that across a variety of naturally-occurring and challenging conditions, such as changes in lighting, background color, haze, decolorization, snow, rain, frost, fog, and contrast, in \textit{twelve distinct datasets} including MNIST, SVHN, GTSRB, CURE-TSR, ImageNet, and ImageNet-c, neural networks trained with our model-based algorithms significantly outperform classifiers trained via empirical risk minimization, norm-bounded robust deep learning algorithms, data augmentation methods, and, when applicable, domain adaptation techniques.  In particular, we show that classifiers trained on ImageNet using our model-based algorithms and tested on various subsets of ImageNet-c improve over state-of-the-art classifiers by up to 30 percentage points.  Furthermore, we show that the model-based robust deep learning paradigm is model-agnostic and adaptable, meaning that it can be used to provide robustness against arbitrary forms of natural variation in data and regardless of whether models of natural variation are known a priori or learned from data.   To demonstrate the broad applicability of our approach, we present apply our paradigm to four novel settings.  (1) First, we show that our algorithms are the first to provide out-of-distribution robustness on a range of challenging settings.  (2) Next, we show that models of natural variation can be composed to provide robustness against multiple simultaneous distribution shifts.  To evaluate this property, we curate several new datasets, each of which has two simultaneous natural shifts.  (3) Thirdly, we show that models of natural variation trained on one dataset can be used to provide robustness on datasets that are entirely unseen while training the model.  (4) Lastly, we show that in the setting of unsupervised domain adaptation, our algorithms outperform traditional domain adaptation techniques by significant margins.

While the experiments in this paper focus on image classification tasks subject to challenging natural conditions, our model-based robust deep learning paradigm is much broader and can, in principle, be applied to many other application domains as long as one can obtain accurate models describing how data naturally varies.  In that sense, we believe that this approach will open up numerous directions for future research.

\vspace{10pt}

\noindent\textbf{Contributions.}  The contributions of our paper can be summarized as follows:

\begin{itemize}
    \item \textbf{Paradigm shift.}  We propose a paradigm shift from perturbation-based adversarial robustness to model-based robust deep learning, in which models of natural variation express changes due to natural conditions that frequently appear in data.
    \item \textbf{Learning models of natural variation.}  For many forms of natural variation that are commonly encountered in safety-critical applications, we use deep generative models to learn models of natural variation from unlabelled data that are consistent with realistic conditions. 
    
    \item \textbf{Robust-optimization-based formulation.}  We formulate a novel training procedure by constructing a general robust optimization problem that searches for challenging out-of-distribution shifts in data to train classifiers to be robust against natural variation.
    
    \item \textbf{Novel model-based training algorithms.} We propose a family of novel training algorithms that exploit models of natural variation to improve the robustness of deep learning against challenging natural conditions captured my models of natural variation.
    
    \item \textbf{ImageNet-c robustness.}  We show that our algorithms improve the robustness of classifiers trained on ImageNet and tested on ImageNet-c by as much as 30 percentage points on a variety of challenging settings, including changes in snow, contrast, brightness, and frost.
    
    \item \textbf{Out-of-distribution robustness.}  We show that our algorithms are the first to consistently provide robustness against natural, out-of-distribution shifts in data, including changes in snow, rain, fog, and brightness on SVHN, GTSRB, CURE-TSR, and ImageNet, that frequently occur in real-world environments.
    
    \item \textbf{Robustness to simultaneous distributional shifts.}  We show that our framework is composable and thus can be used to improve robustness against multiple simultaneous distributional shifts in data.  To evaluate this feature, we curate four new datasets, each of which has two simultaneous distributional shifts.
    
    \item \textbf{Robustness to unseen domains.}  We show that models of natural variation can be reused on datasets that are entirely unseen during training to improve out-of-distribution generalization.  This property demonstrates that model-based robustness is \emph{transferrable} to unseen domains.
    
    \item \textbf{Robustness in the setting of unsupervised domain adaptation.}  We show that in the setting of unsupervised domain adaptation, our algorithms provide higher levels of robustness than traditional domain adaptation techniques.
    
\end{itemize}

\section{Perturbation-based robustness in deep learning}
\label{sect:pert-based-robustness}

In this paper, we consider a standard classification task in which the data is distributed according to a joint distribution $(x,y) \sim\mathcal{D}$ over instances $x\in\R^d$ and corresponding labels $y\in[k] := \{1, \dots, k\}$.  We assume that we are given a suitable loss function $\ell(x, y; w)$; common examples include the cross-entropy or quadratic losses.  In this notation, we let $w\in\R^p$ denote the weights of a neural network.  The goal of the learning task is to find the weights $w$ that minimize the risk over $\mathcal{D}$ with respect to the loss function $\ell$.  That is, we wish to solve
\begin{align}
    w^\star \in \argmin_{w\in\R^p}\E_{(x,y)\sim\mathcal{D}}\left[\ell(x,y;w)\right]. \label{eq:standard-opt}
\end{align}
In a litany of past works, it has been shown empirically that first-order methods (e.g.\ SGD or Adam \cite{kingma2014adam}) can be used to approximately solve \eqref{eq:standard-opt} to obtain weights $w^\star$ that engender neural networks which achieve high classification accuracy on a variety of image classification tasks \cite{lecun2015deep,he2016deep}

As observed in previous works \cite{madry2017towards, wong2017provable}, solving the optimization problem stated in \eqref{eq:standard-opt} does not result in robust neural networks.  More specifically, neural networks trained by solving \eqref{eq:standard-opt} are known to be susceptible to \textit{adversarial attacks}.  This means that given a datum $x$ with a corresponding label $y$, one can find another datum $x^{\text{adv}}$ such that (1) $x$ is close to $x^{\text{adv}}$ with respect to a given Euclidean norm and (2) $x^{\text{adv}}$ is predicted by the learned classifier as belonging to a different class $c\in[k]$ where $c\neq y$.  If such a datum $x^{\text{adv}}$ exists, it is called an \emph{adversarial example}.

To address this striking vulnerability, researchers have sought to improve the robustness of deep learning by developing \textit{adversarial training} algorithms, which inure neural networks against small, norm-bounded perturbations \cite{goodfellow2014explaining}.  The dominant paradigm toward training neural networks to be robust against adversarial examples relies on a robust optimization perspective \cite{ben2009robust}.  Indeed, the approach used in \cite{madry2017towards,wong2017provable} to provide robustness to adversarial examples is formalized by considering a distinct yet related optimization problem to \eqref{eq:standard-opt}.  In particular, the idea is to train neural networks to be robust against a \emph{worst-case} perturbation of  each instance $x$.  This worst-case perspective can be formulated in the following way:
\begin{align}
    w^\star \in \argmin_{w\in\R^p} \E_{(x,y)\sim\mathcal{D}} \left[\max_{\delta\in\Delta} \ell(x + \delta, y; w)\right]. \label{eq:min-max-opt}
\end{align}
Here the set of allowable perturbation $\Delta = \Delta(\epsilon) := \{\delta\in\R^d : \norm{\delta}_p \leq \epsilon\}$ is typically a norm-ball with respect to a suitably chosen Euclidean $p$-norm $\norm{\cdot}_p$.

We can think of the optimization problem in \eqref{eq:min-max-opt} as comprising two coupled optimization problems: an inner maximization problem and an outer minimization problem:
\vspace{-0.8cm}
\begin{multicols}{2}
    \begin{equation}
        \delta^\star \in \argmax_{\delta\in\Delta} \ell(x + \delta, y; w) \label{eq:inner-problem}
    \end{equation} 
    \begin{equation}
        w^\star \in \argmin_{w\in\R^p} \E_{(x,y)\sim\mathcal{D}} \left[ \ell(x + \delta^\star, y; w) \right] \label{eq:outer-problem}
    \end{equation}
\end{multicols}
\noindent First, in the inner maximization problem of \eqref{eq:inner-problem}, we seek a perturbation $\delta\in\Delta$ that results in large loss values when we perturb $x$ by the amount $\delta$. When $\Delta$ is a norm-ball, any solution $\delta^*$ to the inner maximization problem of~\eqref{eq:inner-problem} is a worst-case, norm-bounded perturbation in so much as the datum $x+\delta^*$ is most likely to be classified as any label $c\in[k]$ other than the true label $y$.  If indeed the trained classifier predicts any class $c$ other than $y$ for the datum $x^{\text{adv}} := x+\delta^*$, then $x^{\text{adv}}$ is a \textit{bona fide} adversarial example.  After solving this inner maximization problem, we can rewrite the outer minimization problem via the expression shown in \eqref{eq:outer-problem}.
From this point of view, the goal of the outer problem is to find the weights $w\in\R^p$ that ensure that the worst-case datum $x + \delta^*$ is classified by our model as having label $y$.  To connect robust training to the standard  training paradigm for neural networks given in~\eqref{eq:standard-opt}, note that if $\delta^* = 0$ or if $\Delta=\{0\}$ is trivial, then the outer minimization problem~\eqref{eq:outer-problem} reduces to the empirical risk minimization problem of \eqref{eq:standard-opt}.

\vspace{10pt}

\noindent\textbf{Limitations of perturbation-based robustness.}  While there has been significant progress toward developing algorithms that train neural networks to be robust against norm-bounded perturbations \cite{madry2017towards,wong2017provable,madaan2019adversarial,prakash2018deflecting,zhang2019theoretically,kurakin2016adversarial,moosavi2016deepfool}, there are significant limitations to adversarial training.  Notably, it has been unanimously shown in a spate of recent work that deep learning is also fragile to various forms of \emph{natural variation} \cite{djolonga2020robustness,taori2020measuring,hendrycks2020many,hendrycks2019benchmarking}. In the context of image classification, such natural variation includes  changes in lighting, weather, or background color \cite{eykholt2018robust,hendrycks2019natural,hosseini2018semantic}, spatial transformations such as rotation or scaling \cite{xiao2018spatially,karianakis2016empirical}, and sensor-based attacks \cite{kurakin2016adversarial}.  These realistic forms of variation in data, which are known as \emph{nuisances} in the computer vision community, cannot be modeled by the norm-bounded perturbations $x\mapsto x + \delta$ used in the standard adversarial training paradigm of \eqref{eq:min-max-opt} \cite{sharif2018suitability}.  And while these natural distributional shifts are ubiquitous in numerous application domains, there are remarkably few general, principled techniques that provide robustness against these forms of out-of-distribution, naturally-occurring variation \cite{hendrycks2019augmix}. Therefore, an important open problem in the deep learning community is to develop algorithms that can train neural networks to be robust against natural and realistic forms of out-of-distribution data that are often inherent in safety-critical applications.

\vspace{10pt}

\noindent\textbf{Challenges in designing a more general robustness paradigm.}  Given the efficacy of works that seek to improve the robustness of deep learning against adversarial perturbations, it is of fundamental interest to determine whether the adversarial robustness literature can be leveraged toward developing more general notions of robustness.  To this end, in this paper we identify two fundamental challenges toward achieving this objective.  

Firstly, unlike in the setting of perturbation-based robustness, in real-world environments, data can vary in unknown and highly nonlinear ways.  Thus, the first step toward building a more general robust training procedure must be to design mechanisms that accurately describe how data varies in such environments.  Indeed, in many scenarios, known geometric or physical structure can be used to describe how data naturally varies, as is the case for spatial transformations such as rotations or scalings.  On the other hand, many transformations, such as changes in weather conditions in images, cannot be described by analytical mathematical expressions.  For this reason, it is essential that a more general robustness paradigm be able to leverage known structure and to learn this structure from data when no analytic expression describing how data varies is available.  

The second challenge underlying the task of developing a more general robustness paradigm is to formulate a principled training procedure that leverages suitable models that describe how data varies toward generalizing to out-of-distribution shifts in the data distribution.  Indeed, assuming one has access a suitable model of natural variation, such a procedure should be agnostic to the specific parameterization of the model and adaptable to both models that are known \emph{a priori} as well as models that are learned from data.  

\vspace{10pt}

\noindent\textbf{A unifying solution:\ Model-Based Robust Deep Learning.}  In this paper, we present a new training paradigm for deep learning that improves robustness against natural, out-of-distribution shifts in data by addressing both of these unique challenges.  Rather than perturbing data in a norm-bounded manner, our robust training approach exploits {\em models of natural variation} that describe how data changes with respect to particular shifts in the data distribution.  However, we emphasize that our approach is {\em model-agnostic} in the sense that it provides a paradigm that is applicable across arbitrary classes of naturally-occurring variation. Indeed, in this paper we will show that even if a model of natural variation is not explicitly known \emph{a priori}, one can train neural networks to be robust against natural variation by learning a model of this variation in an offline and data-driven fashion. More broadly, we claim that the framework described in this paper represents a new and more general paradigm for robust deep learning as it provides a methodology for improving the robustness of deep learning against arbitrary sources of natural variation.

\section{Model-based robust deep learning}

In the following section, we introduce the model-based robust deep learning paradigm.  Motivated by past work concerning robustness against adversarially-chosen, norm-bounded perturbations, we formulate a robust optimization problem that characterizes a new notion of robustness with respect to natural variation.  To concertize this formulation, we also offer a geometric interpretation of this novel notion of robustness.

\begin{figure}
    \centering
    \begin{subfigure}[t!]{0.48\textwidth}
        \centering
        \includegraphics[width=0.3\textwidth]{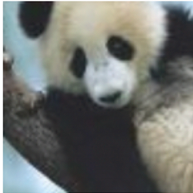}\qquad\qquad
        \includegraphics[width=0.3\textwidth]{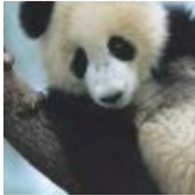}
        \caption{\textbf{Perturbation-based adversarial example.}    In the perturbation-based robustness setting, an input datum such as the image of the panda on the left is perceptually indistinguishable from the adversarial example shown on the right.}
        \label{fig:pandas}
    \end{subfigure} \quad 
    \begin{subfigure}[t!]{0.48\textwidth}
        \centering
        \includegraphics[width=0.48\textwidth]{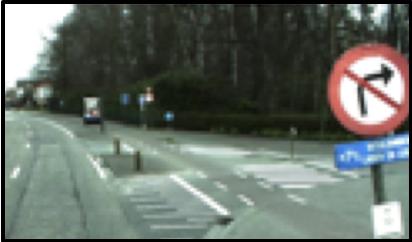}
        \includegraphics[width=0.47\textwidth]{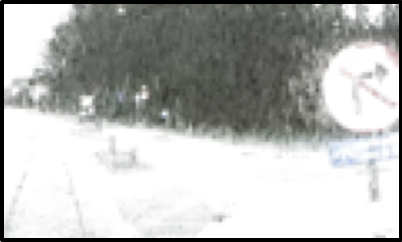}
        \caption{\textbf{Natural variation.}  In this paper, we study robustness with respect to natural variation. In this example, the image of the street in snowy weather on the right vis-a-vis the image on the left illustrates one form of natural variation.}
        \label{fig:street-signs-nat-variation}
    \end{subfigure}
    \caption[A new notion of robustness.]{\textbf{A new notion of robustness.}  The adversarial robustness community has predominantly focused on \textit{norm-bounded} adversaries.  Such adversaries add artificial noise to an input image to produce an \textit{adversarial example} that looks perceptually similar to the input, but fools a deep neural network.  In this paper, we focus on adversaries which change an input datum by subjecting it to \textit{natural variation}.  Such variation often does not obey norm-bounded constraints and renders transformed data perceptually quite different from the original image.}
    \label{fig:compare-adv-robustness}
\end{figure}

\subsection{Adversarial examples versus natural variation}

As we showed in Section \ref{sect:pert-based-robustness}, the problem of defending neural networks against adversaries that can perturb data by a small amount $\delta$ in some Euclidean $p$-norm can be formulated as a robust optimization problem, as described by equation \eqref{eq:min-max-opt}.  In this way, solving \eqref{eq:min-max-opt} engenders neural networks that are robust to imperceptible noise.  This notion of robustness is illustrated in the canonical example shown in Figure~\ref{fig:pandas}, in which the adversary can arbitrarily perturb any pixel values in the image of the panda bear on the left-hand-side to create a new image as long as the perturbation is bounded, meaning that $\delta\in\Delta := \{\delta\in\R^d : \norm{\delta}_\infty \leq \epsilon\}$.
When $\epsilon > 0$ is small, the two panda bears in Figure~\ref{fig:pandas} are seemingly identical and yet the small perturbation $\delta$ can lead to misclassification.

While adversarial training provides robustness against the imperceptible perturbations described in Figure~\ref{fig:pandas}, in natural environments data varies in ways that cannot be captured by additive, norm-bounded perturbations. For example, consider the two traffic signs shown in Figure~\ref{fig:street-signs-nat-variation}.  Note that the images on the left and on the right show the same traffic sign; however, the image on the left shows the sign on a sunny day, whereas the image on the right shows the sign in the middle of a snow storm.  This example prompts several relevant questions.  How do we ensure that neural networks are robust to such natural variation?  How can we rethink adversarial training algorithms to provide robustness against natural-varying and challenging data?

\vspace{10pt}

\begin{figure}
    \centering
    \includegraphics[width=0.8\textwidth]{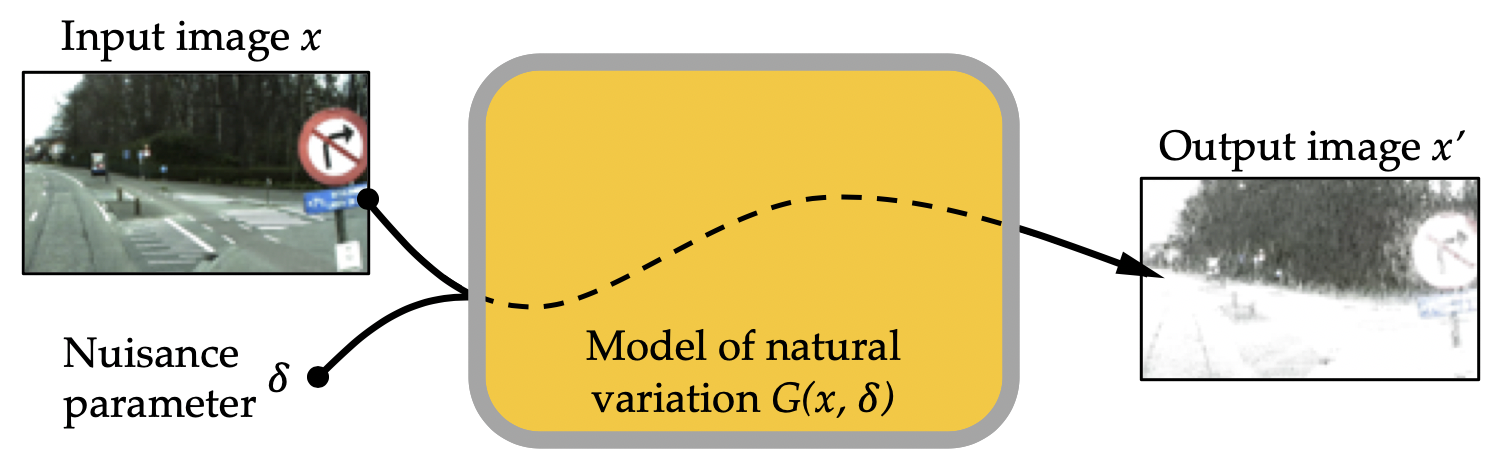}
    \caption[Models of natural variation]{\textbf{Models of natural variation.}  Throughout this paper, we will use \textit{models of natural variation} to describe a wide variety of natural transformations that data are often subjected to in natural, real-world environments.  In our formulation, models of natural variation take the form $G(x, \delta)$, where $x$ is an input datum such as an image and $\delta$ is a \textit{nuisance parameter} that characterizes the extent to which the output datum $x' := G(x, \delta)$ is varied.}
    \label{fig:gen-model-forward-pass}
\end{figure}

\noindent\textbf{Formalizing a more general perspective on robustness.}  In this paper we advocate for a new and more general notion of robustness in deep learning with respect to natural, out-of-distribution shifts in the data.  Critical to our approach is the existence of a {\em model of natural variation} $G(x,\delta)$.  Concretely, a model of natural variation $G:\R^d \times \Delta \rightarrow \R^d$ is a mapping that describes how an input datum $x$ can be naturally varied by a \emph{nuisance parameter} $\delta$ resulting in a new image $x':= G(x, \delta)$. An illustrative example of such a model is shown in Figure~\ref{fig:gen-model-forward-pass}, where the input image $x$ on the left (in this case, in sunny weather) is transformed into a semantically similar image $x'$ on the right (in snowy weather) by varying the nuisance parameter $\delta$.  Ideally, for a fixed input image $x$, the impact of varying the nuisance parameter $\delta$ should be to induce different levels of natural variation on the output image $x'$.  For example, in Figure \ref{fig:gen-model-forward-pass}, the model of natural variation should be able to produce images with a dusting of snow as well as images with an all-out blizzard simply by varying the nuisance parameter $\delta$.

For the time being, we assume the existence of a suitable model of natural variation $G(x,\delta)$; later, in Section~\ref{sect:models-of-natural-var}, we will detail our approach for obtaining models of natural variation that correspond to a wide variety of natural shifts in the data distribution.  In this way, given a model of natural variation $G(x,\delta)$, our immediate goal to develop novel {\em model-based robust training algorithms} that train neural networks to be robust against natural variation by exploiting the model $G(x,\delta)$.   For instance, if $G(x,\delta)$ models variation in the lighting conditions in an image, our model-based training algorithm will provide robustness against lighting discrepancies.  On the other hand, if $G(x,\delta)$ models changes in weather conditions such as in Figure \ref{fig:street-signs-nat-variation}, then our model-based algorithms will improve the robustness of trained classifiers against varying weather conditions.  More generally, our model-based robust training formulation is agnostic to the source of natural variation, meaning that our paradigm is broadly applicable to any source of natural variation that a model of natural variation $G(x,\delta)$ can capture.

\subsection{Formulating the model-based robust optimization problem}

In what follows, we provide a mathematical formulation for the model-based robust deep learning paradigm.  This formulation retains the fundamental elements of the adversarial training paradigm described in Section~\ref{sect:pert-based-robustness}. In this sense, we again consider a classification task in which the goal is to train a neural network with weights $w\in\R^p$ to correctly predict the label $y\in[k]$ of a corresponding input instance $x\in\R^d$, where $(x,y)\sim\mathcal{D}$.  This setting is identical to the setting described in the preamble to equation \eqref{eq:standard-opt}.

Our point of departure from the classical adversarial training formulation of \eqref{eq:min-max-opt} is in the choice of the so-called adversarial perturbation. In this paper, we assume that the adversary has access to a model of natural variation $G(x,\delta)$, which allows it to transform $x$ into a distinct yet related instance $x' := G(x, \delta)$ by choosing different values of $\delta$ from a given \emph{nuisance space} $\Delta$. The goal in this setting is to train a classifier that achieves high accuracy both on a test set drawn i.i.d.\ from $\mathcal{D}$ and on \emph{more-challenging} test data that has been subjected to the source of natural variation that $G$ models.  In this sense, we are proposing a new training paradigm for deep learning that provides robustness against models of natural variation $G(x, \delta)$.   
 
In order to defend a neural network against such an adversary, we propose the following \emph{model-based robust optimization problem}, which will be the central object of study in this paper:
\begin{align}
    \min_{w}\E_{(x,y)\sim\mathcal{D}} \left[\max_{\delta\in\Delta}\ell(G(x, \delta), y; w)\right]. \label{eq:min-max-general}
\end{align}
Here the nuisance space $\Delta$ may be problem-dependent; indeed, different parameterizations of the model of natural variation $G(x,\delta)$ may influence the choice of $\Delta$.  We defer a discussion of the choice of the nuisance space $\Delta$ until Section \ref{sect:alg-selection-criteria}. 

Conceptually, the intuition for this formulation is  similar to the intuition for \eqref{eq:min-max-opt} given in Section \ref{sect:pert-based-robustness}.  Indeed, the optimization problem in \eqref{eq:min-max-general} also comprises an inner maximization problem and an outer minimization problem:
\vspace{-0.8cm}
\begin{multicols}{2}
    \begin{equation}
        \delta^\star \in \argmax_{\delta\in\Delta} \ell(G(x,\delta), y; w) \label{eq:inner-problem-general}
    \end{equation} 
    \begin{equation}
        w^\star \in \argmin_{w\in\R^p} \E_{(x,y)\sim\mathcal{D}} \left[ \ell(G(x,\delta^\star), y; w) \right] \label{eq:outer-problem-general}
    \end{equation}
\end{multicols}
\noindent In the inner maximization problem of \eqref{eq:inner-problem-general}, given an instance-label pair $(x, y)$ and a fixed weight $w\in\R^p$, the adversary seeks a nuisance parameter $\delta^*\in\Delta$ that produces a corresponding instance $x' := G(x, \delta^*)$ which gives rise to high loss values $\ell(G(x, \delta^*), y; w)$ under the current weight $w$.  One can think of this vector $\delta^*$ as characterizing the \emph{most-challenging} nuisance that can be captured by the model $G(x, \delta^*)$ for the original instance $x$.  After solving this inner problem, we can rewrite the outer minimization problem via the expression shown in \eqref{eq:outer-problem-general}.  In this outer problem, we seek the weight $w\in\R^p$ that minimizes the risk against challenging instances of the form $G(x, \delta^*)$.  By training the network to correctly classify these worst-case data, ideally the classifier should become invariant to the model $G(x, \delta)$ for any $\delta\in\Delta$ and consequently to the original source of natural variation.

\subsection{Geometry of model-based robust training}\label{sect:geometry}

\begin{figure}[t]
    \centering
    \begin{subfigure}[t]{0.48\textwidth}
        \centering
        \includegraphics[width=0.85\textwidth]{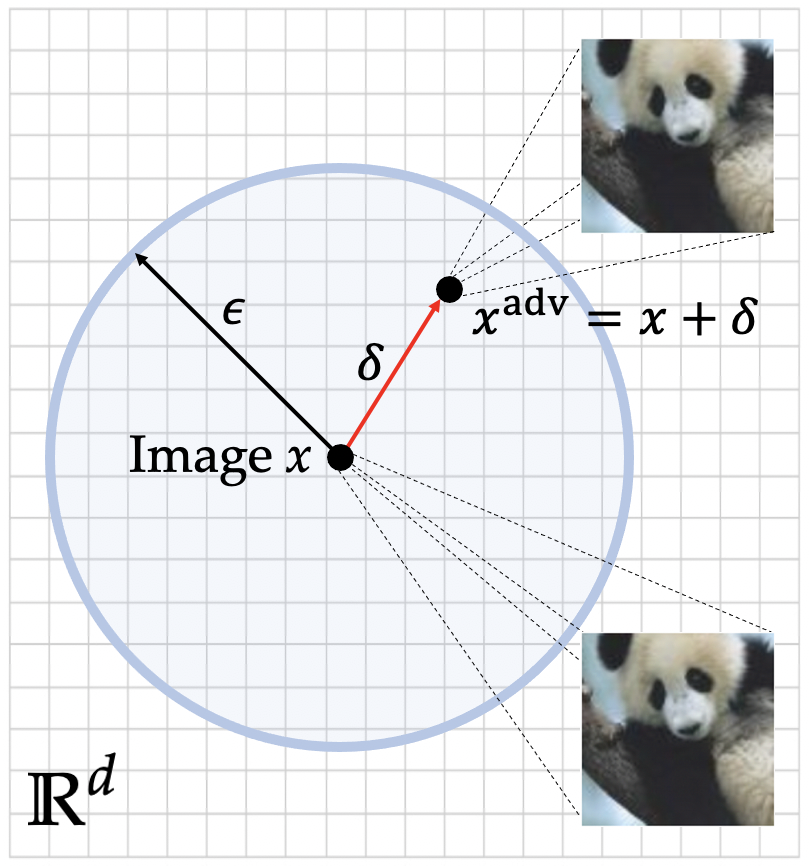}
        \caption{\textbf{Perturbation-based robustness.}  In perturbation-based adversarial robustness, an adversary can perturb a datum $x$ into a perceptually similar datum $x^{\text{adv}} := x + \delta$.  When $\delta$ is constrained to lie in a set $\Delta := \{\delta \in\R^d : \norm{\delta}_p \leq \epsilon\}$, the underlying geometry of the problem can be used to find worst-case additive perturbations.}
        \label{fig:perturb-based-robustness}
    \end{subfigure}\quad
    \begin{subfigure}[t]{0.48\textwidth}
        \centering
        \includegraphics[width=0.90\textwidth]{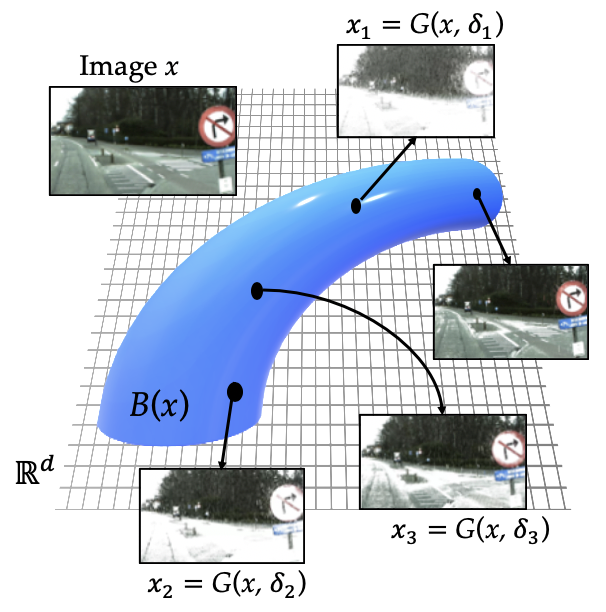}        
        \caption{\textbf{Model-based robustness.}  In our paradigm, models of natural variation can be though of characterizing a class of \emph{learned image manifolds} $B(x)$.  By searching over these manifolds, in the model-based robust deep learning paradigm, we seek images $x'\in B(x)$ that have been subjected to high levels of natural variation.}
        \label{fig:general-robustness}
    \end{subfigure}
    \caption[Geometry of perturbation-based and model-based robustness]{\textbf{Geometry of perturbation-based and model-based robustness.}  Both the perturbation-based and model-based training paradigms have useful geometric interpretations.  Indeed, whereas the perturbation-based paradigm considers a local Euclidean notion of robustness, the model-based paradigm considers much larger changes in data induced by models of natural variation.}
    \label{fig:adversary-difference}
\end{figure}

To provide further intuition for \eqref{eq:min-max-general}, in Figure \ref{fig:adversary-difference}, we consider the underlying geometry of the perturbation-based and model-based robust training paradigms. The geometry of perturbation-based adversarial robustness is shown in Figure~\ref{fig:perturb-based-robustness}, wherein each datum $x$ can be perturbed to any other datum $x^{\text{adv}}$ contained in a small $\epsilon$-neighborhood around $x$.  That is, the data can be additively perturbed via $x\mapsto x^{\text{adv}} := x + \delta$ where $\delta$ is constrained to lie in a set $\Delta := \{\delta\in\R^d  : \norm{\delta}_p \leq \epsilon\}$.  As $\epsilon$ is generally chosen to be quite small, perturbations generated in the perturbation-based paradigm characterize a local notion of robustness, meaning that adversarial examples are constrained to be close to the original image $x$ with respect to a Euclidean norm over the data space $\R^d$. 

Figure~\ref{fig:general-robustness} shows the geometry of the model-based robust training paradigm.  Let us consider a task in which our goal is to correctly classify images of street signs in varying weather conditions.  In the model-based robust training paradigm, we assume that we are equipped with a model of natural variation $G(x, \delta)$ which, by varying the nuisance parameter $\delta\in\Delta$, changes the output image $x' := G(x,\delta)$ according to the natural phenomena captured by the model.  For example, if our data contains images $x$ in sunny weather, the model $G(x, \delta)$ may be designed to continuously vary the weather conditions in these images without changing the scene, other vehicles on the road, or the size and shape of the street signs in these images.  

More generally, such model-based variations around $x$ have a manifold-like structure in the data space $\R^d$.  More specifically, the set of output images corresponding to a fixed input image $x$ that can be obtained by varying the nuisance parameter $\delta$ can be captured by the \emph{learned image manifold} $B(x)$, which we define as follows:
\begin{align}
    B(x) = B_G(x) := \{x' \in \R^d : x'=G(x,\delta) \text{ for some }\delta\in \Delta \} \label{eq:learned-image-manifold}
\end{align}
Note that the learned image manifold is defined implicitly in terms of a model of natural variation $G(x,\delta)$.  Formally, for a fixed image $x$, the set $B(x)$ is a parameterized $\text{dim}(\Delta)$-manifold lying in $\R^d$, where $\text{dim}(\Delta)$ denotes the dimension of the nuisance space $\Delta$.  As we will show, for many models of natural variation, $\text{dim}(\Delta)$ and therefore the dimension of the learned image manifold $B(x)$ will be significantly smaller than the dimension $d$ of the data space $\R^d$.  Furthermore, given the definition of the learned image manifold in \eqref{eq:learned-image-manifold}, the outer maximization problem \eqref{eq:outer-problem-general} can be rewritten in the following way:
\begin{align}
    \delta^\star \in \argmax_{x'\in B(x)} \ell(x', y; w)  \label{eq:inner-problem-manifold}
\end{align}
This shows that solving the inner maximization problem of \eqref{eq:inner-problem-general} corresponds to finding a point $x'$ on the learned image manifold $B(x)$ which causes high loss under the current weight $w\in\R^p$.  The goal of the model-based training algorithms we provide in Section \ref{sect:algorithms} will be to search over these manifolds to find images with challenging levels of natural variation.

\section{Models of natural variation}
\label{sect:models-of-natural-var}

Our model-based robustness paradigm of \eqref{eq:min-max-general} critically relies on the existence of a model of natural variation $G(x,\delta)$ that maps $x\mapsto G(x,\delta) := x'$ and consequently describes how a datum $x$ can be deformed into $x'$ via the choice of a \emph{nuisance parameter} $\delta\in\Delta$.  In this section, we consider cases in which (1) a model $G(x,\delta)$ is known \emph{a priori}, and (2) a model $G(x,\delta)$ is unknown and therefore must be learned offline from data.  In this second case in which models of natural variation must be learned from data, we propose a formulation for obtaining such models.

\subsection{Known models \texorpdfstring{$G(x,\delta)$}{} of natural variation}

\begin{figure}
    \begin{minipage}{0.44\textwidth}
        \centering
        \includegraphics[width=0.9\textwidth]{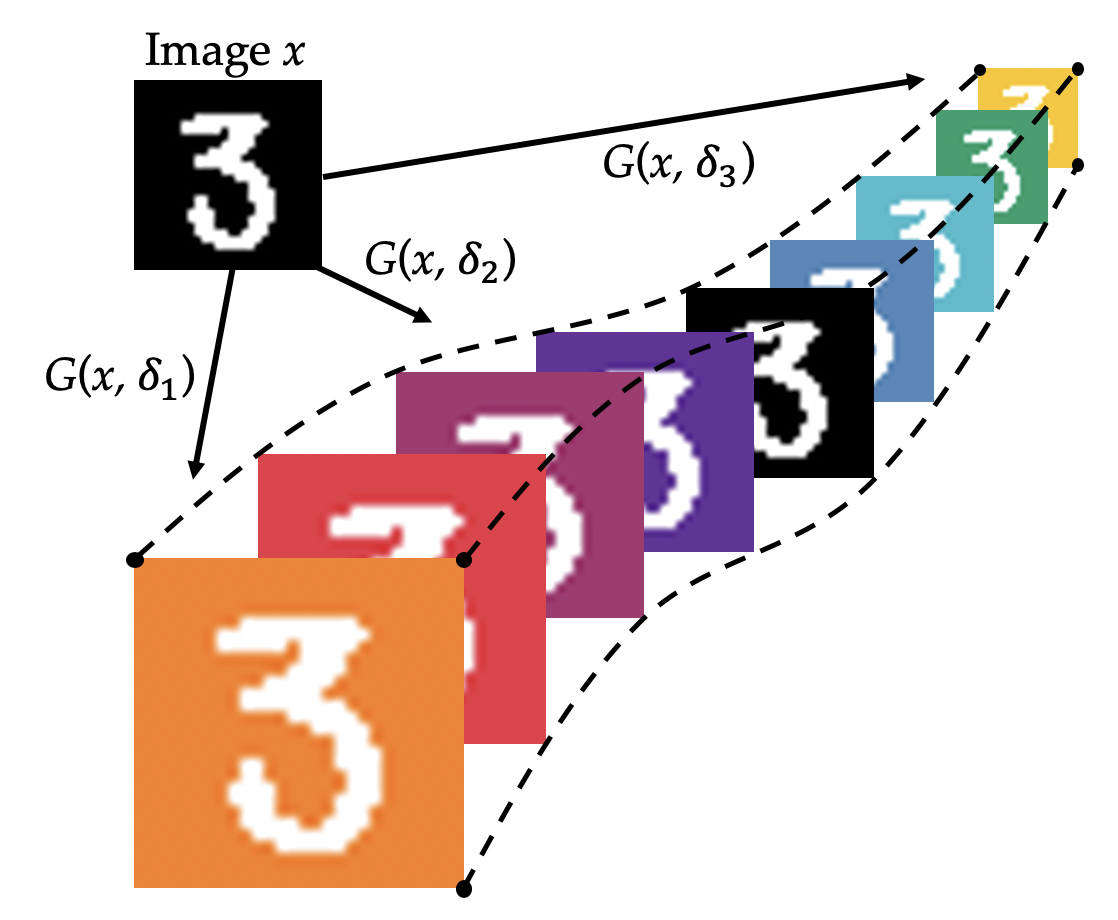}
        
    \end{minipage} \quad
    \begin{minipage}{0.52\textwidth}
        \begin{algorithm}[H]
        \centering
        \begin{algorithmic}[1]
            \Statex \textbf{Inputs:} $x\in\R^{C\times H\times W}$, $\delta := (r, g, b) \in [0, 255]^3$
            \Statex \textbf{Output:} new image $x$
            \Statex
            \State $bgd\_image \gets 0_{C\times H\times W}$ 
            \State $bgd\_image[0,\enskip :,\enskip :] \gets r$
            \State $bgd\_image[1,\enskip :,\enskip :] \gets g$
            \State $bgd\_image[2,\enskip :,\enskip :] \gets b$
            \State $x \gets \texttt{where}(x \leq 12, bgd\_image, x)$
        \end{algorithmic}
        \caption{Known model for background color}
    \label{alg:known-background-color-model}
\end{algorithm}
    \end{minipage}
    \caption[Known models of natural variation]{\textbf{Known models of natural variation.}  In a variety of cases, a model of how data varies in a robustness problem is known \emph{a priori}.  In these cases, the model can immediately be exploited in our model-based training paradigm.  For example, a known model of how background colors change for the MNIST digits can be directly leveraged for model-based training.}
    \label{fig:known-models-of-nat-var}
\end{figure}

For many problems, a model $G(x,\delta)$ is known \emph{a priori} due to underlying physical or geometric laws and can be immediately exploited in our model-based robust training formulation. One straightforward example in which a model of natural variation $G(x,\delta)$ is known is the classical adversarial training paradigm described by equation~\eqref{eq:min-max-opt}.   Indeed, by inspecting equations~\eqref{eq:min-max-opt} and~\eqref{eq:min-max-general}, we can immediately extract the  well-known norm-bounded adversarial model:
\begin{align}
    G(x,\delta)  = x+\delta \quad\text{for } \delta\in\Delta := \{\delta\in\R^d : \norm{\delta}_p \leq \epsilon\}. \label{eq:known-adversarial-model}
\end{align}
The above example of a known model shows that in some sense the perturbation-based adversarial training paradigm of equation~\eqref{eq:min-max-opt} is a special case of the model-based robust deep learning paradigm ~\eqref{eq:min-max-general} when $G(x,\delta)=x+\delta$.  Of course, for this choice of adversarial perturbations there is a plethora of robust training algorithms
\cite{madry2017towards,wong2017provable,madaan2019adversarial,prakash2018deflecting,zhang2019theoretically,kurakin2016adversarial,moosavi2016deepfool}.

Another example of a known model of natural variation is shown in Figure~\ref{fig:known-models-of-nat-var}.  Consider a scenario in which we would like to be invariant to changes in the background color for the MNIST dataset \cite{lecun2010mnist}.  This would require having a model $G(x,\delta)$ that takes an MNIST digit $x$ as input and reproduces the same digit but with various colorized RGB backgrounds which correspond to different values of $\delta\in\Delta$.  This model is relatively simple to describe; pseudocode is provided in Algorithm \ref{alg:known-background-color-model}.

More broadly, there are many settings in which naturally-occuring variation in data has geometric structure that is known \emph{a priori}. For example, in image classification tasks, there are usually intrinsic geometric structures that identify how data can be rotated, translated, or scaled.  Indeed, geometric models for rotating an image along a particular axis can be characterized by a one-dimensional angular parameter $\delta$.  In this case, a known model of natural variation for rotation can be described by 
\begin{align}
    G(x,\delta)  = R(\delta) x
    \quad\text{for }\delta\in\Delta := [0, 2\pi). \label{eq:invariance-to-rotation}
\end{align}
where $R(\delta)$ is a rotation matrix.  Such geometric models can facilitate adversarial distortions of images using a low-dimensional parameter $\delta$.  In prior work, this idea has been exploited to train  neural networks to be robust against rotations of the data around a given axis \cite{engstrom2017exploring,balunovic2019certifying,kamath2020invariance}.    

Altogether, these examples show that for a variety of problems, \emph{known models} can be used to analytically describe how data changes.  In the context of known models, our model-based approach offers a more general framework that is {\em model-agnostic} in the sense that it is applicable to all such models of how data varies.  Before describing our approach for learning models of natural variation from data when a known model is not available, we briefly explore the connection between known models of natural variation and equivariant neural networks.

\vspace{10pt}

\noindent\textbf{Connections to equivariant neural networks.}  Recently, geometric and spatial transformations have been considered in the development of \emph{equivariant} neural network architectures.  In many of these studies, one considers a transformation $T:\R^d \times \Delta \rightarrow \R^d$ where $\Delta$ has some algebraic structure \cite{esteves2017polar}.  By definition, we say that a function $f$ is equivariant with respect to $T$ if $f(T(x,\delta)) = T(f(x),\delta)$ for all $\delta \in \Delta$.  That is, applying $T$ to an input $x$ and then applying $f$ to the result is equivalent to applying $T$ to $f(x)$.   To this end, there has been a great deal of recent work that involves designing architectures that are equivariant to various transformations of data \cite{jaderberg2015spatial,esteves2017polar,worrall2017harmonic,esteves2018learning,cohen2019gauge}. Recently, this has been extended to leveraging group convolutions, which can be used to provide equivariance with respect to certain symmetric groups \cite{cohen2016group} and to permutations of data \cite{guttenberg2016permutation}.  Interestingly, it has been shown that rotationally equivariant neural networks are significantly less vulnerable to geometric invariance-based adversarial attacks \cite{dumont2018robustness}.  In the context of this paper, these structured transformations of data $T:\R^d \times \Delta \rightarrow \R^d$ can be viewed as models of natural variation by directly setting $G(x,\delta)=T(x,\delta)$, where $\Delta$ may have additional group structure.

In contrast to the literature that concerns equivariance, much of the adversarial robustness community has focused on what is often called \emph{invariance}.  A function $f$ is said to be invariant to $T$ if $f(T(x,\delta)) = f(x)$ for any  $\delta \in \Delta$, meaning that transforming an input $x$ by $T$ has no impact on the output.    While  previous approaches exploit such transformations for designing architectures that respect this structure, our goal is to exploit this structure toward developing robust training algorithms.

\subsection{Learning unknown models of natural variation \texorpdfstring{$G(x,\delta)$}{} from data}

\begin{figure}
    \centering
    \includegraphics[width=0.7\textwidth]{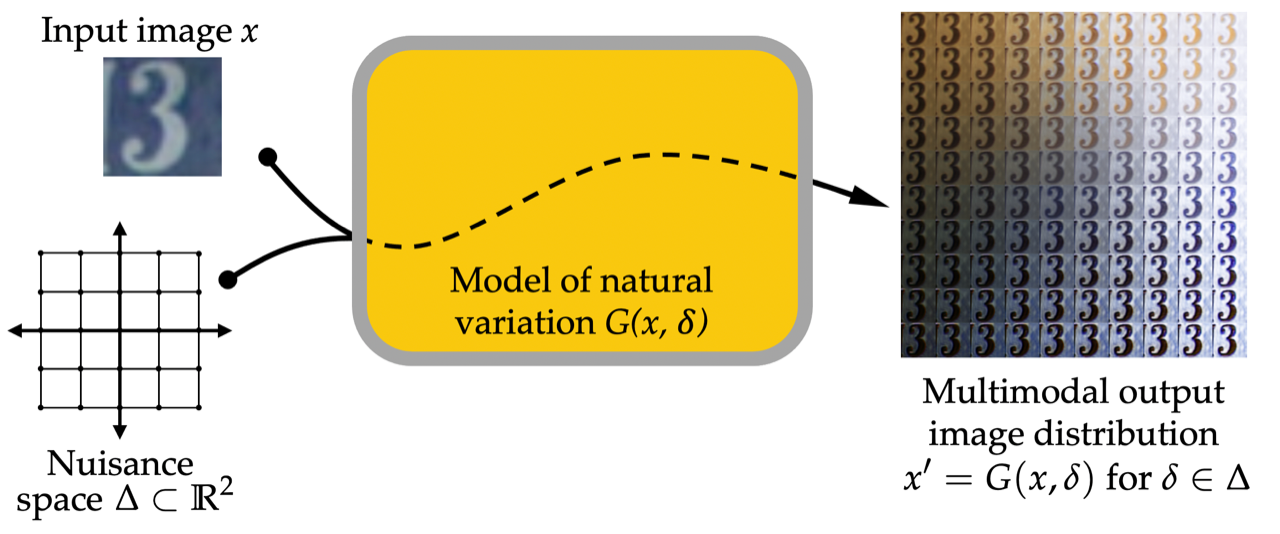}
    \caption[Learned models of natural variation]{\textbf{Learned models of natural variation.}  When a known model of natural variation is not available, we advocate for learning models of natural variation $G(x,\delta)$ offline from data.  To this end, we formulate a statistical procedure that characterizes the problem of learning a suitable model of natural variation $G(x,\delta)$ from unlabelled and unpaired data.  The model in this figure was trained on SVHN using the MUNIT framework to vary the brightness in a given input image; in this case, the nuisance space $\Delta$ was the cube $[-1, 1] \times [-1, 1] \subset\R^2$.}
    \label{fig:learned-model-svhn}
\end{figure}

While geometry and physics may provide analytical models of natural variation $G(x,\delta)$ that can be exploited in our model-based robust training procedure, in many situations such models are not known or are too costly to obtain.  For example, consider Figure~\ref{fig:general-robustness} in which a model of natural variation $G(x,\delta)$ describes the impact of adding snowy weather to an image $x$. In this case, the transformation $G(x, \delta)$ takes an image $x$ of a street sign in sunny weather and maps it to an image $x' := G(x, \delta)$ in snowy weather. Even though there is a relationship between the snowy and the sunny images, obtaining  a model $G$ relating the two images is extremely challenging if we resort to physics or geometric structure.  For such problems, we advocate for \emph{learning} the model $G(x,\delta)$ from data \emph{prior to} model-based robust training.  An example of a learned model of natural variation is shown in Figure \ref{fig:learned-model-svhn}.

In what follows, we introduce a statistical framework for learning models of natural variation from data.  In particular, we advocate for a procedure in which a model of natural variation $G(x, \delta)$ is learned \emph{offline} using \emph{unlabeled and unpaired} data prior to performing model-based robust training on a new and possibly different dataset. We note that while the procedure we describe is quite general, there are likely other formulations that may also result in suitable models of natural variation.  Indeed, one interesting future direction is to explore approaches in which one learns a model of natural variation $G(x,\delta)$ and trains a classifier via the model-based robust training paradigm simultaneously.

\vspace{10pt}

\noindent\textbf{A statistical framework for learning models of natural variation.}  In order to learn a model of natural variation $G(x,\delta)$, we assume that we have access to two \emph{unpaired image domains} $A$ and $B$ that are drawn from a common dataset or distribution.  Generally speaking, in our setting domain $A$ will contain the original data without any natural variation, and domain $B$ will contain data that has been transformed by an underlying natural phenomenon.  Thus, in the example of Figure \ref{fig:general-robustness}, domain $A$ would contain images of traffic signs in sunny weather, and domain $B$ would contain images of street signs in snowy weather.  We emphasize that the domains $A$ and $B$ are unpaired, meaning that it may not be possible to select an image of a traffic sign in sunny weather from domain $A$ and find a corresponding image of that same street sign in the same scene with snowy weather in domain $B$.

Our approach toward formalizing the idea of learning a model of natural variation $G(x,\delta)$ from data is to view $G$ as a mechanism that transforms the distribution of data in domain $A$ so that it resembles the distribution of data in domain $B$.  More formally, let $\Prob_A$ and $\Prob_B$ be the data distributions corresponding to domains $A$ and $B$ respectively.  Our objective is to learn a mapping $(x,\delta)\mapsto G(x,\delta)$ that takes as input a datum $x \sim \Prob_A$ and a nuisance parameter $\delta \in \Delta$ and then produces a new datum $x' \sim \Prob_B$.  Statistically speaking, the nuisance parameter $\delta$ represents the extra randomness or variation required to engender a multimodal distribution over output images distributed according to $\Prob_B$ with different levels of natural variation that correspond semantically to a given input image $x$.  For example, when considering images with varying weather conditions, the randomness in the nuisance parameter $\delta$ might control whether an image of a sunny scene is mapped to a corresponding image with a dusting of snow or to an image in an all-out blizzard.  In this way, we without loss of generality we assume that the nuisance parameter is independently generated from a simple distribution $\Prob_\Delta$ (e.g. uniform or Gaussian) to represent the extra randomness required to generate $x'$ from $x$.  In this sense, the role of the nuisance parameter is somewhat similar to the role of the noise variable in generative adversarial networks \cite{goodfellow2014generative}.

Using this formalism, we can view $G(\cdot,\cdot)$ as a mapping that transforms the distribution $\Prob_A \times \Prob_\Delta$ into the distribution $\Prob_B$. More specifically, $G$ pushes forward the measure $\Prob_A \times \Prob_\Delta$, which is defined over $A \times \Delta$, to $\Prob_B$, which is defined over $B$.  That is, ideally a model of natural variation should approximately satisfy the following expression:
\begin{align}
    \Prob_B = G \, \# \, (\Prob_A \times \Prob_\Delta)
\end{align}
where $\#$ denote the push-forward measure.  Now in order to make this framework for learning a model of natural variation $G$ concrete, we consider a parametric family of models of natural variation $\mathcal{G} := \{G_\theta$ : $\theta \in \Theta\}$ defined over a parameter space $\Theta\subset\R^m$.  We can express the problem of learning a model of natural variation $G_{\theta^*}$ parameterized by the $\theta^*\in\Theta$ that best fits the above formalism in the following way:
 \begin{equation}
     \theta^* \in \argmin_{\theta \in \Theta} \, d\, ( \Prob_B , G_\theta \, \# \, (\Prob_A \times \Prob_\Delta) ). \label{eq:stat-learn-G}
 \end{equation}
Here $d(\cdot,\cdot)$ is an appropriately-chosen distance metric that measures the distance between two probability distributions (e.g. the KL-divergence or Wasserstein distance).  

This problem has received broad interest in the machine learning community thanks to the recent advances in generative modeling.  In particular, in the fields of image-to-image translation and style-transfer, learning mappings between unpaired image domains is a well-studied problem \cite{huang2018multimodal,zhu2017toward,yi2017dualgan}.  In the next subsection, we will show how the breakthroughs in these fields can be used to learn a model of natural variation $G(x,\delta)$ that closely approximates underlying natural phenomena. 

\subsection{Using deep generative models to learn models of natural variation}

\begin{figure}
    \centering
    \includegraphics[width=\textwidth]{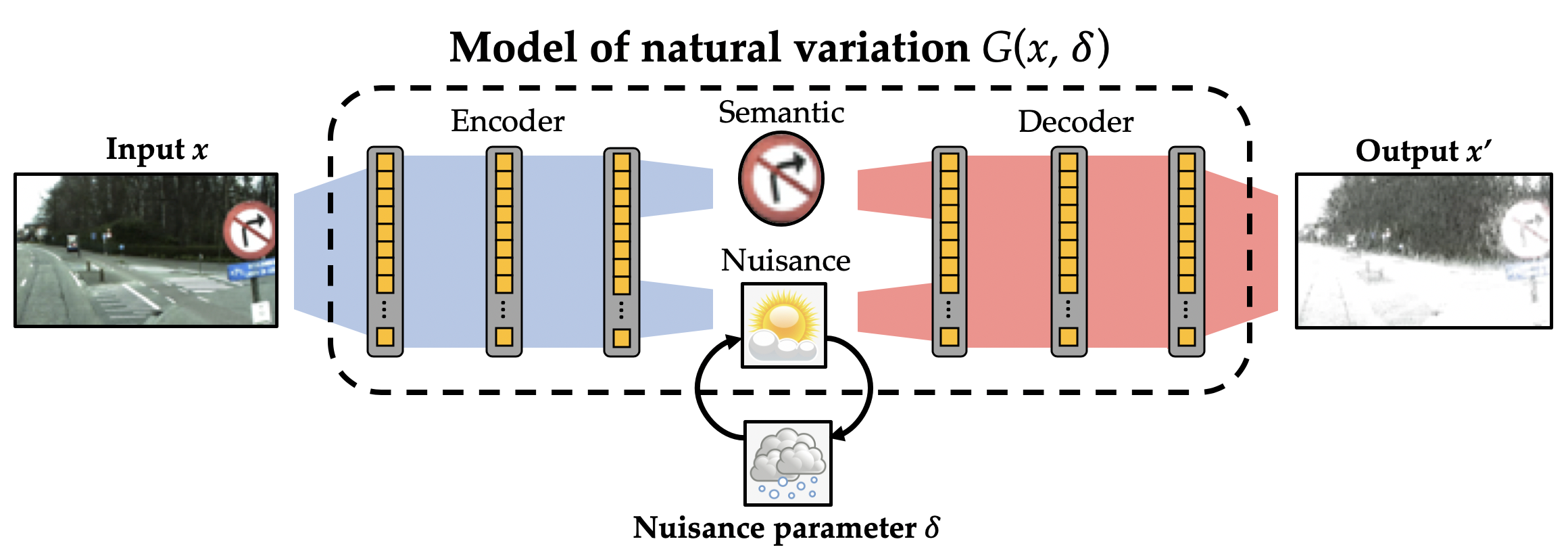}
    \caption[Learning models of natural variation via disentangled representations]{\textbf{Learning models of natural variation via disentangled representations.}  In this paper, we exploit recent progress in generative modeling toward learning models of natural variation $G(x, \delta)$ from data.  Such architectures generally use an encoder-decoder structure, in which an encoding network learns to separate \emph{semantic} from \emph{nuisance} content in two latent spaces, and the decoder learns to reconstruct an image from the representations in these latent spaces \cite{huang2018multimodal, klys2018learning}.  Thus, by varying the nuisance content (which we refer to as the \emph{nuisance parameter}), one can produce a multimodal distribution over output images.}
    \label{fig:learning-models-of-nat-var}
\end{figure}

Recall that in order to learn a model of natural variation from data, we aim to solve the optimization problem in \eqref{eq:stat-learn-G} and to consequently obtain a model $G_{\theta^*}$ that transforms $x\sim\Prob_A$ into corresponding samples $x'\sim\Prob_B$.  In the literature concerning image-to-image translation networks, a variety of works have sought to solve this problem.  Indeed, numerous methods have leveraged \emph{disentangled representations} and \emph{cycle-consistency} toward achieving this goal \cite{zhu2017unpaired,huang2018multimodal,yi2017dualgan,zhu2017toward,almahairi2018augmented,klys2018learning,liu2017unsupervised}.  Furthermore, a closely related line of work has relied on class-conditioning in an image-to-image translation framework to generate realistic images \cite{wang2019stochastic,choi2018stargan,choi2020stargan,mao2019mode}.  We note that while class-conditional image-to-image translation networks have been shown to be successful at generating realistic samples, in our framework we do not assume that the data used for training such networks from either domain is labeled; an exploration of how conditioning could be used toward learning models of natural variation is a fruitful direction for future work.

Among the methods mentioned in the previous paragraph, \cite{klys2018learning,almahairi2018augmented,huang2018multimodal} all seek to learn \emph{multimodal} mappings without relying on class-conditioning, meaning that they seek to disentangle the \textit{semantic content} of a datum (i.e.\ its label or the characterizing component of an input image) from the \emph{nuisance content} (e.g.\ background color, weather conditions, etc.) to produce a multimodal distribution over varying output images.  We highlight these methods because learning a multimodal mapping is a concomitant property toward learning models of natural variation that can produce images subject to a range of natural conditions.  Indeed, any of these architectures are suitable for solving \eqref{eq:stat-learn-G}.  However, throughout the experiments that are presented in Section \ref{sect:experiments}, for simplicity we adhere to a particular choice for the architecture for $G$.

\vspace{10pt}

\noindent\textbf{An architecture for models of natural variation.}  In this paper we will predominantly use the Multimodal Unsupervised Image-to-Image Translation (MUNIT) framework \cite{huang2018multimodal} to learn models of natural variation.  At its core, MUNIT combines two autoencoding networks \cite{kingma2013auto} and two generative adversarial networks (GANs) \cite{goodfellow2014generative} to learn two mappings: one that maps images from domain $A$ to corresponding images in $B$ and one that maps in the other direction from $B$ to $A$.  For completeness, we provide a complete characterization of the MUNIT framework and the hyperparamters we used to train models of natural variation using MUNIT in Appendix \ref{app:training-details}.  For the purposes of this paper, we will only exploit the mapping from $A$ to $B$, although one direction for future work is to incorporate both mappings.  Therefore, the map $G:A\times\Delta\rightarrow B$ learned in the MUNIT framework can be thought of as taking as input an image $x\in A$ and a nuisance parameter $\delta\in\Delta$ and outputting an image $x'\in B$ that has the same semantic content as the input image $x$ but that has a different level of natural variation.  This architecture is illustrated in Figure \ref{fig:learning-models-of-nat-var}.  Notice that the encoding network encodes the input image $x$ into a semantic component and a nuisance parameter; by varying this nuisance parameter and then decoding, the MUNIT framework can be used to vary the natural conditions in the input image.

We emphasize that our model-based robust deep learning paradigm is not reliant on any particular feature of the MUNIT framework.  Indeed, while our results exhaustively show that MUNIT is a suitable method for a variety of datasets and sources of natural variation, an interesting future direction is to explore the impact of different image-to-image translation architectures toward learning suitable models of natural variation.  In Table \ref{tab:models-nat-variation}, we show images from several datasets and corresponding images generated by models of natural variation learned using the MUNIT framework. Each of these learned models of natural variation corresponds to a different source of natural variation.  For each of these models, we used an eight-dimensional nuisance space $\Delta := [0,1]^8\subset\R^8$; the output images are generated by sampling different values from $\Delta$ uniformly at random.

\subsection{A gallery of learned models of natural variation}

To demonstrate the efficacy of the MUNIT framework toward learning models of natural variation $G(x,\delta)$, in Table \ref{tab:models-nat-variation} we show a gallery of learned models of natural variation learned using MUNIT for various datatsets and sources of natural variation.  Importantly, this table shows that our framework in conjunction with the MUNIT architecture can be used to learn perceptually realistic models of natural variation for both low-dimension data (e.g.\ SVHN) and high-dimensional data (e.g.\ ImageNet).  To this end, in Section \ref{sect:alg-selection-criteria}, we will more quantitatively evaluate the ability of learned models of natural variation to produce realistic output image distributions.

\begin{table}
    \centering
    \begin{tabular}{|C{2cm}|c|c|c|} \hline
        \multirow{2}{*}{\textbf{Dataset}} & \multirow{2}{*}{\textbf{\makecell{Natural \\ Variation}}} & \multicolumn{2}{c|}{\textbf{Images}} \\ \cline{3-4}
         
        & & \thead{Original} & \thead{Generated} \\ \hline
      
        \multirow{11}{*}{SVHN} & Brightness &
        \begin{minipage}{2.1cm}
            \centering 
            \vspace{5pt}
            \includegraphics[width=2cm]{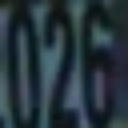}
            \vspace{5pt}
        \end{minipage}
         & 
         \begin{minipage}{6.1cm}
            \centering 
            \vspace{5pt}
            \includegraphics[width=6cm]{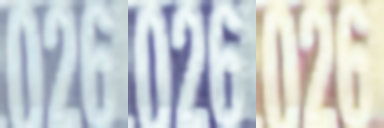}
            \vspace{5pt}
        \end{minipage} \\ \cline{2-4}
        
        & Contrast &
        \begin{minipage}{2.1cm}
            \centering 
            \vspace{5pt}
            \includegraphics[width=2cm]{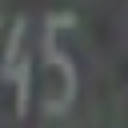}
            \vspace{5pt}
        \end{minipage}
         & 
         \begin{minipage}{6.1cm}
            \centering 
            \vspace{5pt}
            \includegraphics[width=6cm]{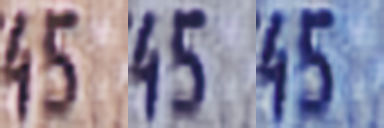}
            \vspace{5pt}
        \end{minipage} \\ \cline{2-4}
        
        & Hue &
        \begin{minipage}{2.1cm}
            \centering 
            \vspace{5pt}
            \includegraphics[width=2cm]{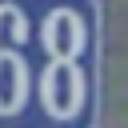}
            \vspace{5pt}
        \end{minipage}
         & 
         \begin{minipage}{6.1cm}
            \centering 
            \vspace{5pt}
            \includegraphics[width=6cm]{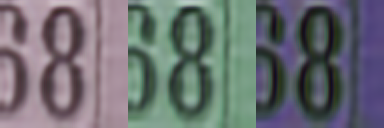}
            \vspace{5pt}
        \end{minipage} \\ \hline

         GTSRB & Brightness &
         \begin{minipage}{2.1cm}
            \centering 
            \vspace{5pt}
            \includegraphics[width=2cm]{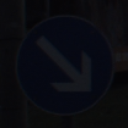}
            \vspace{5pt}
        \end{minipage}
         & 
         \begin{minipage}{6.1cm}
            \centering 
            \vspace{5pt}
            \includegraphics[width=6cm]{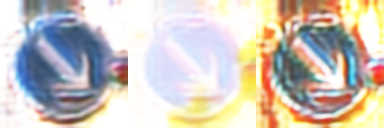}
            \vspace{5pt}
        \end{minipage} \\ \hline
        
       CURE-TSR & Snow & 
       \begin{minipage}{2.1cm}
            \centering 
            \vspace{5pt}
            \includegraphics[width=2cm]{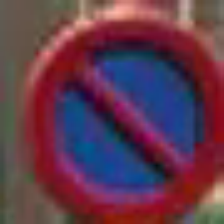}
            \vspace{5pt}
        \end{minipage}
         & 
         \begin{minipage}{6.1cm}
            \centering 
            \vspace{5pt}
            \includegraphics[width=6cm]{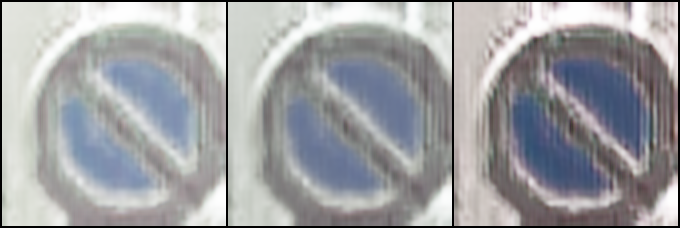}
            \vspace{5pt}
        \end{minipage} \\ \hline
        
        \multirow{11}{*}{ImageNet} & Snow & 
        \begin{minipage}{2.1cm}
            \centering
            \vspace{5pt}
            \includegraphics[width=2cm]{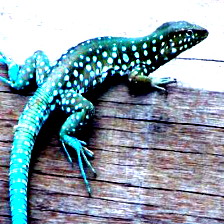}
            \vspace{5pt}
        \end{minipage} &
        \begin{minipage}{6.1cm}
            \centering
            \vspace{5pt}
            \includegraphics[width=6cm]{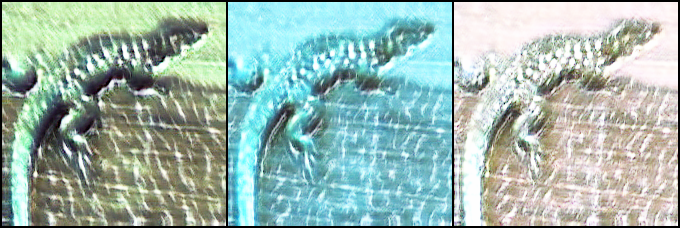}
            \vspace{5pt}
        \end{minipage} \\ \cline{2-4}
        
        & Brightness & \begin{minipage}{2.1cm}
            \centering
            \vspace{5pt}
            \includegraphics[width=2cm]{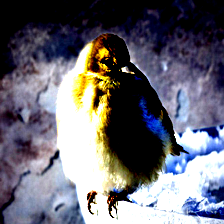}
            \vspace{5pt}
        \end{minipage} &
        \begin{minipage}{6.1cm}
            \centering
            \vspace{5pt}
            \includegraphics[width=6cm]{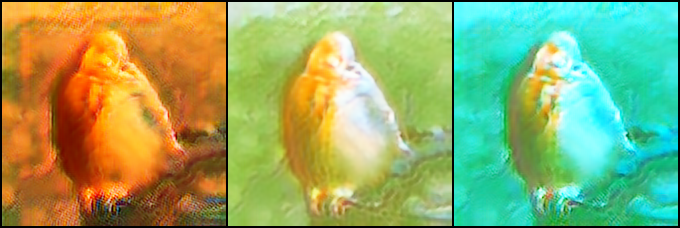}
            \vspace{5pt}
        \end{minipage} \\ \cline{2-4}
        
        & Fog & \begin{minipage}{2.1cm}
            \centering
            \vspace{5pt}
            \includegraphics[width=2cm]{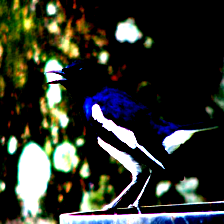}
            \vspace{5pt}
        \end{minipage} &
        \begin{minipage}{6.1cm}
            \centering
            \vspace{5pt}
            \includegraphics[width=6cm]{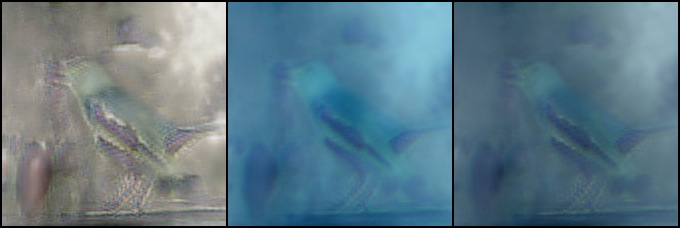}
            \vspace{5pt}
        \end{minipage} \\ \hline
        
    \end{tabular}
    \caption[A gallery of learned models of natural variation]{\textbf{A gallery of learned models of natural variation.}  For a range of datasets, we show images generated by passing data through learned models of natural variation.}
    \label{tab:models-nat-variation}
\end{table}

\section{Model-based robust training algorithms}
\label{sect:algorithms}

In the previous section, we described a procedure that can be used to train models of natural variation $G(x, \delta)$.  In 
some cases, such models may be \textit{known a priori} while in other cases such models may be \textit{learned} offline from data.  Regardless of their origin, we will now assume that we have access to a suitable model $G(x, \delta)$ and shift our attention toward exploiting $G$ in the development of novel robust training algorithms.  

To begin, recall the optimization-based formulation of \eqref{eq:min-max-general}.  Given a model of natural variation $G(x,\delta)$, \eqref{eq:min-max-general} is a nonconvex-nonconcave min-max problem, and is therefore difficult to solve exactly.  We will therefore resort to approximate methods for solving this challenging optimization problem.  To elucidate our approach for solving \eqref{eq:min-max-general}, we first characterize the problem in the finite-sample setting.  That is, rather than assuming access to the full joint distribution $(x, y)\sim\mathcal{D}$, we assume that we are given given a finite number of samples $\mathcal{D}_n := \{(x^{(j)}, y^{(j)})\}_{j=1}^n$ distributed i.i.d.\ according to the true data distribution $\mathcal{D}$.  The empirical version of \eqref{eq:min-max-general} in the finite-sample setting can be expressed in the following way:
\begin{align}
    w^\star \in \argmin_{w\in\R^p} \frac{1}{n} \sum_{j=1}^n \left[ \max_{\delta\in\Delta} \ell\left(G\left(x^{(j)}, \delta\right), y^{(j)}; w\right)\right]. \label{eq:min-max-empirical}
\end{align}
Concretely, we search for the parameter $w\in\R^p$ that induces the smallest empirical error while each sample $(x^{(j)}, y^{(j)})$ is varied according to the model of natural variation $G(x^{(j)},\delta)$.  In particular, while subjecting each instance-label pair $(x^{(j)}, y^{(j)})$ to the source of natural variation modeled by $G$, we search for nuisance parameters $\delta\in\Delta$ so as to train the classifier on the most challenging natural conditions.

When the learnable weights $w\in\R^p$ parameterize a neural network $f_{w}$, the outer minimization problem and the inner maximization problem are inherently nonconvex and nonconcave respectively.  Therefore, we will rely on zeroth- and first-order optimization techniques for solving this problem to a locally optimal solution. We will propose three algorithmic variants, each of which takes a integer parameter $k > 0$: (1) \emph{Model-based Robust Training} (MRT-$k$), (2) \emph{Model-based Adversarial Training} (MAT-$k$), and (3) \emph{Model-based Data Augmentation} (MDA-$k$).  

\vspace{10pt}

\noindent\textbf{An overview of the model-based robust training algorithms.}  Each of the three algorithms we propose in this paper -- MRT-$k$, MAT-$k$, and MDA-$k$ -- seeks a solution to \eqref{eq:min-max-empirical} by alternating between solving the outer minimization problem and solving the inner maximization problem.  Indeed, one similarity amongst these three algorithms is that each procedure seeks a solution to the outer problem by using a standard first-order optimization technique (e.g.\ SGD or Adam).  However, the algorithms differ in how they search for a solution to the inner problem; at a high level, each of these methods seeks such a solution by augmenting the original training dataset $\mathcal{D}_n$ with new data generated by a given model of natural variation $G(x,\delta)$.  In particular, MRT randomly queries $G$ to generate several new data points and then selects those generated data that induce the highest loss in the inner maximization problem.  On the other hand, MAT employs a gradient-based search in the nuisance space $\Delta$ to find loss-maximizing generated data. Finally, MDA augments the training dataset $\mathcal{D}_n$ with generated data by sampling randomly in $\Delta$ to produce a wide range of natural conditions.  We note that past approaches have used similar adversarial \cite{madry2017towards} and statistical \cite{volpi2018generalizing} augmentation techniques.  However, the main difference between these past works and our algorithms are that our algorithms exploit models of natural variation $G(x,\delta)$ to generate new data.  

In the remainder of this section, we will describe each algorithm in detail and provide psuedocode for each algorithm.  Python implementations of each algorithm are available at the following link: \url{https://github.com/arobey1/mbrdl}.

\subsection{Model-based Robust Training (MRT)}

\begin{algorithm}[t!]
    \centering
    \begin{algorithmic}[1]
        \Statex \textbf{Input: } weight initialization $w$,  trade-off parameter $\lambda\in[0, 1]$, number of steps $k$
        \Statex \textbf{Output: } learned weight $w$
        \Statex 
        \Repeat 
        \For{\text{ minibatch} $B_m := \{(x^{(1)}, y^{(1)}), (x^{(2)}, y^{(2)}), \dots, (x^{(m)}, y^{(m)}) \} \subset \mathcal{D}_n$}
        \State Initialize $max\_loss \gets 0$ and $\delta_{\text{adv}} := ( \delta_{\text{adv}}^{(1)}, \delta_{\text{adv}}^{(2)}, \dots, \delta_{\text{adv}}^{(m)} ) \gets (0_q, 0_q, \dots, 0_q)$
        \For{$k$ steps}
            \State Sample $\delta^{(j)}$ uniformly at random from $\Delta$ for $j=1,\dots,m$
            \State $current\_loss \gets \sum_{j=1}^m \ell(G(x^{(j)}, \delta^{(j)}), y^{(j)}; w)$
            \If{$current\_loss > max\_loss$}
                \State $max\_loss \gets current\_loss$
                \State $\delta_{\text{adv}}^{(j)} \gets \delta^{(j)}$ for $j = 1, \dots, m$
            \EndIf
        \EndFor
        \State $g \gets \nabla_{w}  \sum_{j=1}^m [ \ell(G(x^{(j)}, \delta_{\text{adv}}^{(j)}), y^{(j)}; w) + \lambda \cdot \ell(x^{(j)}, y^{(j)}; w) ]$
        \State $w \gets \text{Update}(g, w)$
        \EndFor
        \Until {convergence}
    \end{algorithmic}
    \caption{Model-based Robust Training (MRT)}
    \label{alg:MRT}
\end{algorithm}

In general, solving the inner maximization problem in \eqref{eq:min-max-empirical} is difficult and motivates the need for methods that yield approximate solutions.  In this vein, one simple scheme is to sample different nuisance parameters $\delta\in\Delta$ for each instance-label pair $(x^{(j)},y^{(j)})$ and among those sampled values, find the nuisance parameter $\delta^{\text{adv}}$ that gives the highest empirical loss under $G$. Indeed, this approach is not designed to find an exact solution to the inner maximization problem; rather it aims to find a difficult example by sampling in the nuisance space of the model of natural variation.

Once we obtain this difficult example by sampling in $\Delta$, the next objective is to solve the outer minimization problem.  The procedure we propose in this paper for solving this problem amounts to using the worst-case nuisance parameter $\delta^{\text{adv}}$ obtained via the inner maximization problem to perform data-augmentation.  That is, for each instance-label pair $(x^{(j)}, y^{(j)})$, we treat $(G(x^{(j)}, \delta^{\text{adv}}), y^{(j)})$ as a new instance-label pair that can be used to supplement the original dataset $\mathcal{D}_n$.  These training data can be used together with first-order optimization methods to solve the outer minimization problem to a locally optimal solution $w^*$.

Algorithm \ref{alg:MRT} contains the pseudocode for the MRT algorithm.  In particular, in lines 4-11, we search for a difficult example by sampling in $\Delta$ and picking the parameter $\delta_{\text{adv}} \in \Delta$ that induces the highest empirical loss.  Then in lines 12-13, we calculate a stochastic gradient of the loss with respect to the weights $w\in\R^p$ of the classifier; we then use this gradient to update $w$ using a first-order method.  There are a number of potential algorithms for this \texttt{Update} function in line 13, including stochastic gradient descent (SGD), Adam \cite{kingma2014adam}, and Adadelta \cite{zeiler2012adadelta}.

Throughout the experiments in the forthcoming sections, we will train classifiers via MRT with different values of $k$.  In this algorithm, $k$ controls the number of data points we consider when searching for a loss-maximing datum.  To make clear the role of $k$ in this algorithm, we will refer to Algorithm \ref{alg:MRT} as MRT-$k$ when appropriate.

\subsection{Model-based Adversarial Training (MAT)}

\begin{algorithm}[t]
    \centering
    \begin{algorithmic}[1]
        \Statex \textbf{Input: } weight initialization $w$, trade-off parameter $\lambda\in[0,1]$, number of steps $k$
        \Statex \textbf{Output: } learned weight $w$
        \Statex
        \Repeat 
            \For{\text{ minibatch} $B_m := \{(x^{(1)}, y^{(1)}), (x^{(2)}, y^{(2)}), \dots, (x^{(m)}, y^{(m)}) \} \subset \mathcal{D}_n$}
                \State Initialize $\delta_{\text{adv}} := ( \delta_{\text{adv}}^{(1)}, \delta_{\text{adv}}^{(2)}, \dots, \delta_{\text{adv}}^{(m)} ) \gets (0_q, 0_q, \dots, 0_q)$
                \For{$k$ steps}
                    \State $g_{\text{adv}} \gets \nabla_{\delta_{\text{adv}}} \sum_{j=1}^m \ell(G(x^{(j)}, \delta_{\text{adv}}^{(j)}), y^{(j)}; w)$
                    \State $\delta_{\text{adv}} \gets \Pi_{\Delta} [\delta_{\text{adv}} + \alpha g_{\text{adv}}]$ 
                \EndFor
                \State $g \gets \nabla_{w}  \sum_{j=1}^m [ \ell(G(x^{(j)}, \delta_{\text{adv}}^{(j)}), y^{(j)}; w) + \lambda \cdot \ell(x^{(j)}, y^{(j)}; w) ]$
                \State $w \gets \text{Update}(g, w)$\qquad 
            \EndFor
        \Until{convergence}
    \end{algorithmic}
    \caption{Model-based Adversarial Training (MAT)}
      \label{alg:MAT}
\end{algorithm}

At first look, the sampling-based approach used by MRT may not seem as powerful as a first-order (i.e. gradient-based) adversary that has been shown to be effective at improving the robustness of trained classifiers against norm-bounded, perturbation-based attacks \cite{athalye2018obfuscated}. Indeed, it is natural to extend the ideas encapsulated in this previous work that advocate for first-order adversaries to the model-based setting.  That is, under the assumption that our model of natural variation $G(x,\delta)$ is differentiable, in principle we can use projected gradient ascent (PGA) in the nuisance space $\Delta\subset\R^q$ of a given model to solve the inner maximization problem.  This idea motivates the formulation of our second algorithm, which we call Model-based Adversarial Training (MAT).

In Algorithm \ref{alg:MAT}, we present pseudocode for MAT.  Notably, by ascending the stochastic gradient with respect to $\delta_{\text{adv}}$ in lines 4-7, we seek a nuisance parameter $\delta_{\text{adv}}^*$ that maximizes the empirical loss.  In particular, in line 6 we perform the update step of PGA to obtain $\delta_{\text{adv}}\in\Delta$; in this notation, $\Pi_{\Delta}$ denotes the projection onto the set $\Delta$.  However, performing PGA until convergence at each iteration leads to a very high computational complexity.  Thus, at each training step, we perform $k$ steps of projected gradient ascent.  Following this procedure, we use the loss-maximization nuisance parameter $\delta_{\text{adv}}^*$ to augment $\mathcal{D}_n$ with data $G(x^{(j)}, \delta_{\text{adv}}^*)$ that has been subjected to worst-case nuisance variability.  The update step is then carried out by computing the stochastic gradient of the loss over the augmented training sample with respect to the learnable weights $w\in\R^p$ in line 8.  Finally, we update $w$ in line 9 in a similar fashion as was done in the description of the MRT algorithm.

An empirical analysis of the performance of MAT will be given in Section \ref{sect:experiments}.  To emphasize the role of the number of gradient steps $k$ used to find a loss maximizing nuisance parameter $\delta_{\text{adv}}^*\in\Delta$, we will often refer to Algorithm $\ref{alg:MAT}$ as MAT-$k$.

\subsection{Model-based Data Augmentation (MDA)}

\begin{algorithm}[t!]
    \centering
    \begin{algorithmic}[1]
        \Statex \textbf{Input: } weight initialization $w$, trade-off parameter $\lambda\in[0, 1]$, number of steps $k$
        \Statex \textbf{Output: } learned weight $w$
        \Statex
        \Repeat 
        \For{\text{ minibatch} $B_m := \{(x^{(1)}, y^{(1)}), (x^{(2)}, y^{(2)}), \dots, (x^{(m)}, y^{(m)}) \} \subset \mathcal{D}_n$}
        \State Initialize $x_{i}^{(j)} \gets 0_d$ for $i = 1, \dots, k$ and for $j = 1, \dots, m$  
        \For{$k$ steps}
            \State Sample $\delta^{(j)}$ randomly from $\Delta$ for $j=1,\dots,m$
            \State $x_i^{(j)} \gets G(x^{(j)}, \delta^{(j)})$ for $j=1,\dots, m$ 
        \EndFor
        \State $g \gets \nabla_{w}  \sum_{j=1}^m [ \sum_{i=1}^k \ell(x_i^{(j)}, y^{(j)}; w) + \lambda \cdot \ell(x^{(j)}, y^{(j)}; w) ]$
        \State $w \gets \text{Update}(g, w)$
        \EndFor
        \Until {convergence}
    \end{algorithmic}
    \caption{Model-Based Data Augmentation (MDA)}
    \label{alg:MDA}
\end{algorithm}

Both MRT and MAT adhere to the common philosophy of selecting loss-maximizing data generated by a model of natural variation $G(x,\delta)$ to augment the original training dataset $\mathcal{D}_n$.  That is, in keeping with the min-max formulation of \eqref{eq:min-max-general}, both of these methods search \textit{adversarially} over $\Delta$ to find challenging natural variation.  More specifically, for each data point $(x^{(j)}, y^{(j)})$, these algorithms select $\delta \in \Delta$ such that $G(x^{(j)}, \delta) =: x_{\text{adv}}^{(j)}$ maximizes the loss term $\ell(x_{\text{adv}}^{(j)}, y^{(j)}; w)$.  The guiding principle behind these methods is that by showing the neural network these challenging, model-generated data during training, the trained classifier will be able to robustly classify data over a wide spectrum of natural conditions.

Another interpretation of \eqref{eq:min-max-general} is as follows.  Rather than taking an adversarial point of view in which we expose neural networks to the most challenging model-generated examples, an alternative is to expose these networks to a {\em diversity} of model-generated data during training.   In this approach, by augmenting $\mathcal{D}_n$ with model-generated data corresponding to a wide range of natural variations $\delta\in\Delta$, one might hope to achieve higher levels of robustness with respect to a given model of natural variation $G(x,\delta)$.

This idea motivates the third and final algorithm, which we call Model-based Data Augmentation (MDA).  The psuedocode for this algorithm is given in Algorithm~\ref{alg:MDA}.  Notably, rather than searching adversarially over $\Delta$ to find model-generated data subject to worst-case (i.e. loss-maximizing) natural variation, in lines 4-7 of MDA we randomly sample in $\Delta$ to obtain a diverse array of nuisance parameters.  For each such nuisance parameter, we augment $\mathcal{D}_n$ with a new datum and calculate the stochastic gradient with respect to the weights $w\in\R^p$ in line 8 using both the original dataset $\mathcal{D}_n$ and these diverse augmented data.

In MDA, the parameter $k$ controls the number of model-based data points per data point in $\mathcal{D}_n$ that we append to the training set.  To make this explicit, we will frequently refer to Algorithm \ref{alg:MDA} as MDA-$k$.

\section{Experiments} \label{sect:experiments}

We present experiments in \emph{five different and challenging settings} over \emph{twelve distinct datasets} to demonstrate the broad applicability of the model-based robust deep learning paradigm.  

\vspace{10pt}

\noindent\textbf{Experimental overview.}  First, in Sections \ref{sect:out-of-dist-experiments}-\ref{sect:imagenet-experiments}, we show that our algorithms are the first to consistently provide out-of-distribution robustness across a range of challenging corruptions, including shifts in brightness, contrast, snow, fog, frost, and haze on CURE-TSR, ImageNet, and ImageNet-c.  Next, in Section \ref{sect:composition-experiments}, we show that models of natural variation can be composed to provide robustness against simultaneous shifts.  To evaluate this feature, we curate several new datasets containing simultaneous sources of natural variation.  Following this, in Section \ref{sect:transfer-experiments}, we show that models of natural variation $G(x,\delta)$ trained on a fixed dataset can be reused to provide robustness on datasets entirely unseen while training $G$.  This demonstrates that model-based robustness is highly \emph{transferable}, meaning that once a model of natural variation has been learned for a particular source of natural variation, it can be reused to provide robustness against the same source of natural variation in future applications.  Finally, in Section \ref{sect:unsup-dom-adapt-experiments}, we show that in the setting of unsupervised domain adaptation, our algorithm outperforms a well-known domain adaptation baseline.  

\vspace{10pt}

\begin{figure}[t]
    \centering
        \includegraphics[width=\textwidth]{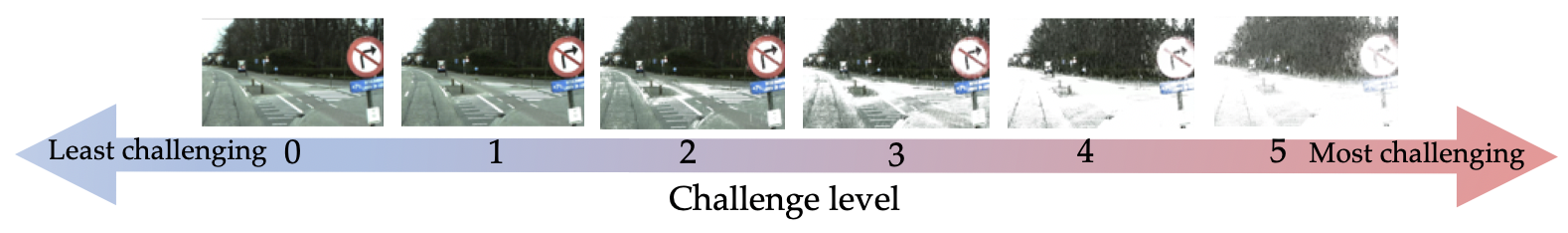}
        \caption[CURE-TSR snow challenge levels]{\textbf{CURE-TSR snow challenge levels.}  From left to right we show the same image with different levels of snow natural variation.  Challenge-level 0 corresponds to no natural variation, whereas challenge-level 5 corresponds to the highest level of natural variation. }
        \label{fig:cure-snow-challenges}
\end{figure}

\noindent\textbf{Notation for image domains.}   Throughout this section, we consider a wide range of datasets, including MNIST \cite{lecun2010mnist}, SVHN \cite{netzer2011reading}, GTSRB \cite{Stallkamp-IJCNN-2011}, CURE-TSR \cite{temel2019traffic}, MNIST-m \cite{ganin2016domain}, Fashion-MNIST \cite{xiao2017fashion}, EMNIST \cite{cohen2017emnist}, KMNIST \cite{clanuwat2018deep}, QMNIST \cite{yadav2019cold}, USPS \cite{hull1994database}, ImageNet \cite{deng2009imagenet}, and ImageNet-c \cite{hendrycks2019benchmarking}.  For many of these datasets, we extract subsets corresponding to different sources of natural variation; henceforth, we will call these subsets {\em domains}.  To indicate the source of natural variation under consideration in a given experiment, we use the notation ``\texttt{source} (\textit{A}$\rightarrow$\textit{B})'' to denote a distributional shift from domain $A$ to domain $B$.  For example, ``contrast (low$\rightarrow$high)'' will denote a shift from low-contrast to high-contrast within a particular dataset.  Images from domains $A$ and $B$ for each of the shifts used in this paper are available in Appendix \ref{app:gallery-of-learned-models}.  We note that our experiments contain domains with both \emph{natural} and \emph{artificially-generated} variation; details concerning how we extracted non-artificial variation can be found in Appendix \ref{app:datasets}.

We emphasize that each image domain used in this paper contains a training set and a test set.  While both the training and test set in each domain come from the same distribution, neither the models of natural variation nor the classifiers have access to test data from any domain during the training phase.  More explicitly, when learning models of natural variation $G(x,\delta)$ and training classifiers, we use data from the \textit{training set} of the relevant domains.  Conversely, when testing the classifiers, we use data from the \textit{test set} of the relevant domains.  

\vspace{10pt}

\noindent\textbf{Baseline algorithms and evaluation metrics.}  In the experiments, we consider a variety of baseline algorithms.  In particular, where approapriate, we compare our model-based algorithms to empirical risk mimization (ERM), the adversarial training algorithm PGD \cite{madry2017towards}, a recently-proposed data-augmentation technique called AugMix \cite{hendrycks2019augmix}, and the domain adaptation technique known as Adversarial Discriminative Domain Adaptation (ADDA) \cite{tzeng2017adversarial}.  To evaluate the performance of each of these algorithms, we will report the top-1 accuracy (i.e.\ the standard classification accuracy) and, where appropriate, the top-5 accuracy (i.e.\ the frequency with which the ground-truth label is one of the top five predicted classes).  Further details concerning architecture selection and the hyperparameters used are given in Appendix \ref{app:training-details}.

\subsection{Out-of-distribution robustness}
\label{sect:out-of-dist-experiments}

\begin{table}[]
    \centering
    \begin{tabular}{|c|c|c|c|c|c|c|c|c|c|} \hline
         \multirow{4}{*}{\thead{\makecell{CURE-TSR \\ subset}}} & \multicolumn{9}{c|}{\thead{Test accuracy (top-1) on challenge-levels 3, 4, and 5}} \\ \cline{2-10}
         & \multicolumn{3}{c|}{\thead{ERM + Aug}} & \multicolumn{3}{c|}{\thead{PGD + Aug}} & \multicolumn{3}{c|}{\thead{MRT}} \\ \cline{2-10}
         & \thead{3} & \thead{4} & \thead{5} & \thead{3} & \thead{4} & \thead{5} & \thead{3} & \thead{4} & \thead{5}  \\ \hline
         Snow & 86.5 & 74.8 & 60.9 & 82.9 & 77.3 & 61.8 & \textbf{88.0} & 
        \textbf{77.8} & \textbf{70.7} \\ \hline
         
         Haze & 55.2 & 54.0 & 47.5 & 83.8 & 63.1 & 53.4 & \textbf{83.9} & \textbf{79.1} & \textbf{70.1} \\ \hline
         
         Decolorization & 87.9 & 85.1 & 78.8 & 84.7 & 75.2 & 64.9 & \textbf{90.5} & \textbf{89.6} & \textbf{89.4} \\ \hline
         
         Rain & 72.7 & 71.7 & 66.9 & 68.9 & 66.4 & 60.5 & \textbf{80.7} & \textbf{78.7} & \textbf{74.8} \\ \hline
    \end{tabular}
    \caption[Out-of-distribution robustness]{\textbf{Out-of-distribution robustness.}  In each experiment, we train a model of natural variation $G(x,\delta)$ to map from challenge-level 0 to challenge-level 2 data from various subsets of CURE-TSR.  We then perform model-based training using challenge-level 0 data and test on challenge-levels 3-5.  We allow all baseline classifiers access to labeled data from challenge-levels 0 and 2.  As our method does not use labeled data from challenge-level 2, this is in some sense an unfair comparison to our algorithms; despite this, model-based algorithms outperform the baselines across the board.}  
    \label{tab:ood-experiments}
\end{table}

In many applications, one might have data corresponding to low levels of natural variation, such as a dusting of snow in images of street signs.  However, it is often difficult to collect data corresponding to high levels of natural variation, such as images taken during a blizzard.  In such cases, we show that our algorithms can be used to provide significant out-of-distribution robustness against data with high levels of natural variation by training on data with relatively low levels of the same source of natural variation.  To do so, we use data from the CURE-TSR dataset \cite{temel2019traffic}, which contains images of street signs divided into subsets according to various sources of natural variation and corresponding severity levels.  For example, for images in the ``snow'' subset, challenge-level 0 corresponds to no snow, whereas challenge-level 5 corresponds to a full blizzard.  

For each row of Table \ref{tab:ood-experiments}, we use unlabeled data from challenge-levels 0 and 2 to learn a model of natural variation corresponding to a given subset of CURE-TSR.  We then train classifiers using MRT-10 with labeled challenge-level 0 data.  We also train classifiers using ERM and PGD using the labeled data from challenge-levels 0 and 2.  To denote the fact that ERM and PGD are trained using an \emph{augmented} dataset containing labeled challenge-level 2 data in addition to challenge-level 0 data, we denote these algorithms in Table \ref{tab:ood-experiments} by \texttt{ERM+Aug} and \texttt{PGD+Aug}.  We then test all classifiers on data from challenge-levels 3, 4, and 5.  Note that in some sense this is an unfair comparison for our methods, given that the model-based algorithms are not given access to labeled challenge-level 2 data. In spite of this unfair comparison, our algorithms still outperform the baseline algorithms.

By considering each row in \ref{tab:ood-experiments}, a general pattern emerges.  When tested on challenge-level 3 and 4 data, MRT improves over ERM+Aug and PGD+Aug, although in some cases the improvements are somewhat modest.  These modest improvements are due in part to the fact that challenge-level 3 data has only slightly more natural variation than challenge-level 2 data, which all of the classifiers have seen in either labeled form (in the case of the baselines) or unlabeled form (in the case of MRT).  However, by comparing the performance of the trained classifiers on challenge-level 5 data, it is clear that MRT improves significantly over the baselines by as much as 20 percentage points.  This shows that as the challenge-level becomes more severe, our model-based algorithms outperform the baselines by relatively larger margins.

\subsection{Model-based robustness on the shift from ImageNet to ImageNet-c}
\label{sect:imagenet-experiments}

\begin{table}[]
    \centering
    \begin{tabular}{|c|c|c|c|c|c|c|c|c|} \hline
         \multirow{2}{*}{\thead{\makecell{Model dataset \\ (classes 0-9)}}} & \multirow{2}{*}{\thead{\makecell{Training dataset \\ (classes 10-59) }}} & \multirow{2}{*}{\makecell{\thead{Test dataset \\ (classes 10-59) }}} & \multicolumn{6}{c|}{\thead{Test accuracy (top-1/top-5)}} \\ \cline{4-9}
         
         & & & \multicolumn{2}{c|}{\thead{ERM}} & \multicolumn{2}{c|}{\thead{AugMix}} &  \multicolumn{2}{c|}{\thead{MDA}} \\ \hline
         
         IN-c Snow & \multirow{4}{*}{ImageNet} & IN-c Snow & 20.9 & 49.9 & 1.10 & 8.3 & \textbf{31.1} & \textbf{61.2} \\ \cline{1-1} \cline{3-9} 
         
         IN-c Contrast & & IN-c Contrast & 41.1 & 73.4 & 0.72 & 6.76 & \textbf{50.0} & \textbf{79.5} \\ \cline{1-1} \cline{3-9} 
         
         IN-c Brightness & & IN-c Brightness & 26.9 & 59.2 & 0.56 & 5.20 & \textbf{53.0} & \textbf{81.7} \\ \cline{1-1} \cline{3-9} 
         
         IN-c Fog & & IN-c Fog & 8.7 & 29.5 & \textbf{29.1} & \textbf{57.8} & 24.6 & 50.8 \\ \cline{1-1} \cline{3-9} 
         
         IN-c Frost & & IN-c Frost & 16.3 & 39.0 & 29.5 & 58.4 & \textbf{36.0} & \textbf{67.2} \\ \hline
    \end{tabular}
    \caption[ImageNet to ImageNet-c robustness]{\textbf{ImageNet to ImageNet-c robustness.}  In each experiment, we train a model of natural variation to map from classes 0-9 of ImageNet to the same classes from a subset of ImageNet-c.  Next, we use this model to perform model-based training on classes 10-59 of ImageNet, and we test each network on classes 10-59 from the same subset ImageNet-c on which the model was trained.  Therefore, despite the fact that each model of natural variation $G(x,\delta)$ is trained on classes that are entirely distinct from the classes considered when training and testing the classifier, the model $G$ still provides significant levels of robustness against the shift from ImageNet to ImageNet-c.}
    \label{tab:imagenet-experiments}
\end{table}

To demonstrate the scalability of our approach, we perform experiments on ImageNet \cite{deng2009imagenet} and the recently-curated ImageNet-c dataset \cite{hendrycks2019benchmarking}.  ImageNet-c contains images from the ImageNet test set that are corrupted according to artificial transformations, such as snow, rain, and fog, and are labeled from 1-5 depending on the severity of the corruption.  For brevity, in Table \ref{tab:imagenet-experiments}, we abbreviate ImageNet-c to ``IN-c.''  

For numerous challenging corruptions, we train models to map from the classes 0-9 of ImageNet to the corresponding classes of ImageNet-c.  We then train all classifiers, each of which uses the ResNet50 architecture \cite{he2016deep}, on classes 10-59 of ImageNet, and test on the corresponding classes for various subsets of ImageNet-c.  Note that in this setting, the ImageNet classes used to train the model of natural variation are disjoint from those that are used to train the classifier, so many techniques, including most domain adaptation methods, do not apply.  To offer a point of comparison, we include the accuracies of classifiers trained using AugMix, which is a recently proposed method that adds known transformations to the data \cite{hendrycks2019augmix}.  

By considering Table \ref{tab:imagenet-experiments}, the first notable observation is that the classifiers trained using ERM perform very poorly on the shift from ImageNet to ImageNet-c.  That is, when evaluating ResNet50 classifiers trained on ImageNet on the original ImageNet test set, one would expect to attain a peak top-1 accuracy of more than 80 percentage points \cite{yalniz2019billion}; therefore, the drop in classification accuracy for the classifiers trained using ERM is remarkable given that the corruptions included in this table are quite common.  This demonstrates the degree to which classifiers trained on ImageNet lack robustness to common corruptions and transformations of data \cite{hendrycks2019benchmarking}.  To this end, the authors of \cite{hendrycks2019augmix} sought to address this fragility by introducing the AugMix algorithm, which adds randomly corrupted data to augment the training dataset.  As shown in Table \ref{tab:imagenet-experiments}, while this method does improve accuracy on two challenges, for other corruptions, the top-1 and top-5 accuracies plummet.  On the other hand, in almost all cases that we considered, MDA-3 significantly outperformed both methods, in some cases improving by nearly 30 percentage points in top-1 accuracy.

\subsection{Robustness to simultaneous distributional shifts}
\label{sect:composition-experiments}

\begin{table}[]
    \centering
    \begin{tabular}{|c|c|c|c|c|} \hline
         \multirow{2}{*}{\thead{\makecell{Dataset}}} & \multirow{2}{*}{\thead{\makecell{Challenge 1 \\ (dom.\ $A_1$$\rightarrow$dom.\ $B_1$)}}} & \multirow{2}{*}{\thead{\makecell{Challenge 2 \\ (dom.\ $A_2$$\rightarrow$dom.\ $B_2$)}}} & \multicolumn{2}{c|}{\thead{Test acc.\  (top-1)}} \\ \cline{4-5}
         & & & \thead{ERM} & \thead{MDA} \\ \hline
         SVHN & Brightness (low$\rightarrow$high) & Contrast (low$\rightarrow$high) & 54.9 & \textbf{67.2} \\ \hline

         \multirow{4}{*}{\makecell{ImageNet \& \\ ImageNet-c}} & Brightness (low$\rightarrow$high) & Contrast (high$\rightarrow$low) & 13.6 & \textbf{49.9} \\ \cline{2-5}
         
         & Brightness (low$\rightarrow$high) & Snow (no$\rightarrow$yes) & 53.3 & \textbf{58.3} \\ \cline{2-5}
         
         & Brightness (low$\rightarrow$high) & Fog (no$\rightarrow$yes) & 50.3 & \textbf{58.8} \\ \cline{2-5}
         
         & Contrast (high$\rightarrow$low) & Fog (no$\rightarrow$yes) & 8.40 & \textbf{23.2} \\ \hline
    \end{tabular}
    \caption[Composing models of natural variation]{\textbf{Composing models of natural variation.}  We consider distributional shifts corresponding to two distinct and simultaneous sources of natural variation.  To perform model-based training, we first train two models of natural variation $G_1(x,\delta)$ and $G_2(x,\delta)$ separately using unlabeled data to describe two distinct sources of natural variation.  We then compose these models to form a new model $G(x,\delta) := G_1(G_2(x, \delta), \delta)$, which can vary the natural conditions in a given input image $x$ according to both of the sources of natural variation modeled by $G_1$ and $G_2$.}
    \label{tab:composition}
\end{table}

In practice, it is common to encounter multiple simultaneous distributional shifts.  For example, in image classification, there may be shifts in both brightness and contrast; yet while there may be examples corresponding to shifts in either brightness or contrast in the training data, there may not be any examples of both shifts occurring simultaneously.  To address this robustness challenge, for each row of Table \ref{tab:composition}, we learn two models of natural variation $G_1(x,\delta)$ and $G_2(x,\delta)$ using unlabeled training data corresponding to two separate shifts, which map domains $A_1$$\rightarrow$$B_1$ (e.g.\ low- to high-brightness) and $A_2$$\rightarrow$$B_2$ (e.g.\ low- to high-contrast).  We then compose these models to form a new model
\begin{align}
    G(x,\delta) = G_1(G_2(x, \delta), \delta)
\end{align}
which can be used to vary the natural conditions in a given input image $x$ according to both sources of natural variation modeled by $G_1$ and $G_2$.  We then train classifiers on labeled data from $A_1\cup A_2$ (e.g.\ data with \emph{either} low-brightness \emph{or} low-contrast) and test on data from $B_1\cap B_2$ (e.g. data with \emph{both} high-brightness \emph{and} high-contrast).  

To gather data from $B_1 \cap B_2$ containing images with high-brightness and high-contrast from SVHN, we threshold the SVHN training set according to both of these sources of natural variation.  On the other hand, to create the data from $B_1\cap B_2$ for the ImageNet experiments, \emph{we curate four new datasets} by applying pairs of transformations that were originally used to create the ImageNet-c datasets.  Images corresponding to these dataset, which contain shifts in brightness and contrast, brightness and snow, brightness and fog, and contrast and fog, as well as more details concerning how they were curated are available in Appendix \ref{app:datasets}.  

The results in Table \ref{tab:composition} reveal that across each of the settings we considered, MDA-3 significantly outperformed ERM with respect to top-1 classification accuracy.  Despite the fact that neither algorithm has access to data from domain $B_1\cap B_2$ during training, MDA is able to improve by between 5 and 35 percentage points over ERM across these five settings.  The composable nature of our methods is unique to our framework; we plan to explore this feature further in future work.

\subsection{Transferability of model-based robustness}
\label{sect:transfer-experiments}

\begin{table}[]
    \centering
    \begin{tabular}{|c|c|c|c|c|c|c|c|} \hline
         \multirow{2}{*}{\thead{\makecell{Training \\ dataset $\mathcal{D}_1$}}} &  \multirow{2}{*}{\thead{\makecell{Test \\ dataset $\mathcal{D}_2$}}} & \multirow{2}{*}{\thead{\makecell{Challenge \\ (dom.\ A$\rightarrow$dom.\ B)}}} & \multicolumn{5}{c|}{\thead{Test accuracy (top-1)}} \\ \cline{4-8} 
         & & & \thead{ERM} & \thead{PGD} & \thead{MRT} & \thead{MDA} & \thead{MAT} \\ \hline
         
         \multirow{5}{*}{MNIST} & \makecell{Fashion- \\ MNIST} & \multirow{5}{*}{\makecell{Background \\ color \\ (blue$\rightarrow$red)}} & 69.3 & 67.7 & $\mathbf{81.4}$ & 80.1 & 76.1 \\ \cline{2-2} \cline{4-8}
         
         & Q-MNIST & & 87.0 & 79.9 & $\mathbf{98.0}$ & $\mathbf{98.0}$ & $\mathbf{98.0}$ \\ \cline{2-2} \cline{4-8}
         
         & E-MNIST & & 63.5 & 49.3 & \textbf{86.1} & 85.9 & 84.1 \\ \cline{2-2} \cline{4-8}
         
         & K-MNIST & & 47.9 & 47.7 & 89.1 & \textbf{89.3} & 86.8 \\ \cline{2-2} \cline{4-8}
         
         & USPS & & 89.9 & 87.4 & 93.3 & \textbf{93.4} & 91.9 \\ \hline
         
         SVHN & MNIST-m & Decolorization (no$\rightarrow$yes) & 76.1 & 75.3 & 77.1 & 78.3 & \textbf{79.2} \\ \hline
         
         GTSRB & CURE-TSR & Brightness (high$\rightarrow$low) & 47.6 & 43.6 & \textbf{73.0} & 72.4 & 67.8 \\ \hline
         
         ImageNet \& & \multirow{2}{*}{CURE-TSR} & Snow (no$\rightarrow$yes) & 52.0 & 53.0 & 59.4 & \textbf{62.2} & 59.4 \\ \cline{3-8}
         
         ImageNet-c & & Brightness (low$\rightarrow$high) & 41.5 & 40.2 & 46.6 & 46.7 & \textbf{47.5} \\ \hline
    \end{tabular}
    \caption[Transferability of model-based robustness]{\textbf{Transferability of model-based robustness.}  In each experiment, we train a model of natural variation on a given training dataset $\mathcal{D}_1$.  Then, we use this model to perform model-based training on a new dataset $\mathcal{D}_2$ entirely unseen during the training of the model.  This shows that once they have been trained, models of natural variation can be used to improve robustness in new applications without the need for retraining.}
    \label{tab:transfer}
\end{table}

Because we learn models of natural variation $G(x,\delta)$ offline before training a classifier, our paradigm can be applied to domains that are \emph{entirely unseen} while training the model $G$.  In particular, we show that models can be reused on similar yet unseen datasets to provide robustness against a common source of natural variation. For example, one might have access to two domains corresponding to the shift from images of European street signs taken during the day to images taken at night.  However, one might wish to provide robustness against the same shift from daytime to nighttime on a new dataset of American street signs without access to any nighttime images in this new dataset.   Whereas many techniques, including most domain adaptation methods, do not apply in this scenario, in the model-based robust deep learning paradigm, we can simply learn a model of natural variation $G(x,\delta)$ corresponding to the changes in lighting for the European street signs and then apply this model to the dataset of the American signs.

Table \ref{tab:transfer} shows several experiments of this stripe in which a model of natural variation $G(x,\delta)$ is learned on one dataset $\mathcal{D}_1$ and then applied on another dataset $\mathcal{D}_2$.  Notably, classifiers trained using our model-based algorithms significantly outperform the ERM and PGD baselines in each case, despite the fact that all classifiers have access to the same training data from dataset $\mathcal{D}_2$.  In particular, by using models of natural variation trained on $\mathcal{D}_1$, and then training in the model-based paradigm on $\mathcal{D}_2$, we are able to improve the test accuracy on the shift from domain $A$ to domain $B$ on $\mathcal{D}_2$ by as much as 40 percentage points.  This demonstrates that models of natural variation can be reused to provide robustness against domains that are entirely unseen during the training of the model of natural variation.

 \subsection{Model-based robust deep learning for unsupervised domain adaptation}
\label{sect:unsup-dom-adapt-experiments}

The results of Sections \ref{sect:out-of-dist-experiments}, \ref{sect:imagenet-experiments}, \ref{sect:composition-experiments}, and \ref{sect:transfer-experiments} show that our approach does not require labelled or unlabelled data from domain $B$ to improve robustness on the shift from domain $A$ to domain $B$.  In particular, in Section \ref{sect:out-of-dist-experiments}, we trained classifiers on low levels of natural variation and then evaluated these classifiers on higher levels of natural variation.  Next, in Section \ref{sect:imagenet-experiments} we used a different distribution of classes to train a model of natural variation $G(x,\delta)$ than we used to train and evaluate classifiers.  Further, in Section \ref{sect:composition-experiments}, we assumed access to data corresponding to two fixed distributional shifts, but not to the data corresponding to both shifts occurring simultaneously.  Finally, in  Section \ref{sect:transfer-experiments}, we trained models of natural variation $G(x,\delta)$ on a different dataset $\mathcal{D}_1$ from the dataset $\mathcal{D}_2$ used to train and evaluate classifiers.  

However, when unlabelled data corresponding to a fixed domain shift from domain $A$ to domain $B$ is available, it is of interest to evaluate how our approach compares to relevant methods such as domain adaptation.  We emphasize that while this is one of the most commonly studied settings in domain adaptation, it represents only one particular setting to which the model-based robust deep learning paradigm can be applied.

In Table \ref{tab:uda-experiments}, for each shift from domain $A$ to $B$, we assume access to labeled data from domain $A$ and unlabeled data from domain $B$.  In each row, we use unlabeled data from both domains to train a model of natural variation $G(x,\delta)$.  We then train classifiers using our algorithms, ERM, and PGD using data from domain $A$; we then evaluate these classifiers on data from the test set of domain $B$.  Furthermore, we compare to ADDA, which is a well-known domain adaptation method \cite{tzeng2017adversarial}.  In every scenario, our model-based algorithms significantly outperform the baselines, often by 10-20 percentage points.  Notably, while ADDA offers strong performance on SVHN, it fairs significantly worse on GTSRB and CURE-TSR.  

\begin{table}[]
    \centering
    \begin{tabular}{|c|c|c|c|c|c|c|c|} \hline
         \multirow{2}{*}{\thead{Dataset}} & \multirow{2}{*}{\thead{\makecell{Challenge \\ (dom.\ A$\rightarrow$dom.\ B)}}} & \multicolumn{6}{c|}{\thead{Test accuracy  (top-1)}} \\ \cline{3-8}
         & & \thead{ERM} & \thead{PGD} & \thead{ADDA} & \thead{MRT} & \thead{MDA} & \thead{MAT} \\ \hline
         
         \multirow{2}{*}{SVHN} & Brightness (low$\rightarrow$high) & 30.5 & 36.2 & 60.1 & \textbf{70.9} & 69.5 & 52.2  \\ \cline{2-8}
         
         & Contrast (low$\rightarrow$high) & 55.9 & 57.9 & 54.6 & \textbf{74.3} & 74.1 & 55.2 \\ \hline
         
         \multirow{2}{*}{GTSRB} & Brightness (low$\rightarrow$high) & 40.3 & 34.7 & 27.6 & 50.4 & 48.3 & \textbf{64.8} \\ \cline{2-8}
         
         & Contrast (low$\rightarrow$high) & 44.5 & 41.9 & 14.7 & 68.4 & \textbf{69.4} & 55.1 \\ \hline
         
         \multirow{3}{*}{CURE-TSR} & Snow (no$\rightarrow$yes) & 52.0 & 53.0 & 16.1 & 74.0 & \textbf{74.5} & 72.3 \\ \cline{2-8} 
         
         & Haze (no$\rightarrow$yes) & 57.2 & 50.9 & 49.2 & 72.5 & 70.0 & \textbf{74.6} \\ \cline{2-8}
         
         & Rain (no$\rightarrow$yes) & 62.6 & 62.3 & 16.5 & 75.2 & 73.7 & \textbf{75.3} \\ \hline
    \end{tabular}
    \caption[Model-based training in the setting of unsupervised domain adaptation]{\textbf{Model-based training in the setting of unsupervised domain adaptation.}  In each experiment, we assume access to labeled data from domain $A$ as well as unlabeled data from domain $B$; we use (unlabelled) data from both domains to train a model of natural variation $G(x,\delta)$ for each row of this table.  We then train classifiers on labeled data from domain $A$, and test classifiers on test data from domain $B$.  We compare to suitable baselines, including the domain adaptation method ADDA, which uses the labelled domain $A$ data as well as the unlabelled domain $B$ data.}
    \label{tab:uda-experiments}
\end{table}
\section{Discussion} \label{sect:discussion}

Following the experiments of Section \ref{sect:experiments}, we now discuss further aspects of model-based training.  In particular, we provide an additional experiment which shows that more accurate models of natural variation engender classifiers that are more robust against natural variation in our paradigm.  We also discuss choices for the integer parameter $k$ in each of the model-based training algorithms.  Furthermore, we provide details concerning how we chose to define the nuisance spaces $\Delta$ for learned models of natural variation used in the experiments of Section \ref{sect:experiments}.

\subsection{Impact of model quality }
\label{sect:better-models}

An essential yet so far undiscussed piece of the efficacy of our model-based paradigm is the impact of the quality of learned models of natural variation on the robustness we are ultimately able to provide.  In scenarios where we do not have access to a known model of natural variation, the ability to provide any sort of meaningful robustness relies on learned models that can accurately render realistic looking data consistent with various sources of natural variation.  To this end, it is reasonable to expect that models that can more effectively render realistic yet challenging data should result in classifiers that are more robust to shifts in natural variation. 

\begin{figure}[t]
    \centering
    \begin{subfigure}[b]{0.4\textwidth}
        \centering
        \begin{subfigure}{0.3\textwidth}
            \centering
            \includegraphics[width=\textwidth]{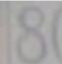}
            \caption{\textbf{Original.}}
            \label{fig:orig-better-models}
        \end{subfigure}
        
        \begin{subfigure}{0.23\textwidth}
            \centering
            \includegraphics[width=\textwidth]{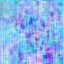}
            \caption{\textbf{10.}}
            \label{fig:10-better-models}
        \end{subfigure}
        \begin{subfigure}{0.23\textwidth}
            \centering
            \includegraphics[width=\textwidth]{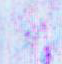}
            \caption{\textbf{100.}}
        \end{subfigure}
        \begin{subfigure}{0.23\textwidth}
            \centering
            \includegraphics[width=\textwidth]{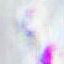}
            \caption{\textbf{250.}}
        \end{subfigure}
        \begin{subfigure}{0.23\textwidth}
            \centering
            \includegraphics[width=\textwidth]{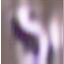}
            \caption{\textbf{500}}
        \end{subfigure}
        
        \begin{subfigure}{0.23\textwidth}
            \centering
            \includegraphics[width=\textwidth]{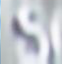}
            \caption{\textbf{1000.}}
        \end{subfigure}
        \begin{subfigure}{0.23\textwidth}
            \centering
            \includegraphics[width=\textwidth]{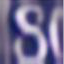}
            \caption{\textbf{2000.}}
        \end{subfigure}
        \begin{subfigure}{0.23\textwidth}
            \centering
            \includegraphics[width=\textwidth]{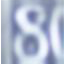}
            \caption{\textbf{3000.}}
        \end{subfigure}
        \begin{subfigure}{0.23\textwidth}
            \centering
            \includegraphics[width=\textwidth]{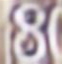}
            \caption{\textbf{4000.}}
            \label{fig:4k-better-models}
        \end{subfigure}
        \caption*{\textbf{Output images from models in $\mathcal{G}$.}  We show an example image from domain $A$ in (a), and subsequently show the corresponding output images for each $G\in\mathcal{G}$ for a randomly chosen $\delta\in\Delta$ in (b)-(i).}
    \end{subfigure} \quad
    \begin{subfigure}[b]{0.56\textwidth}
        \centering
        \includegraphics[width=\textwidth]{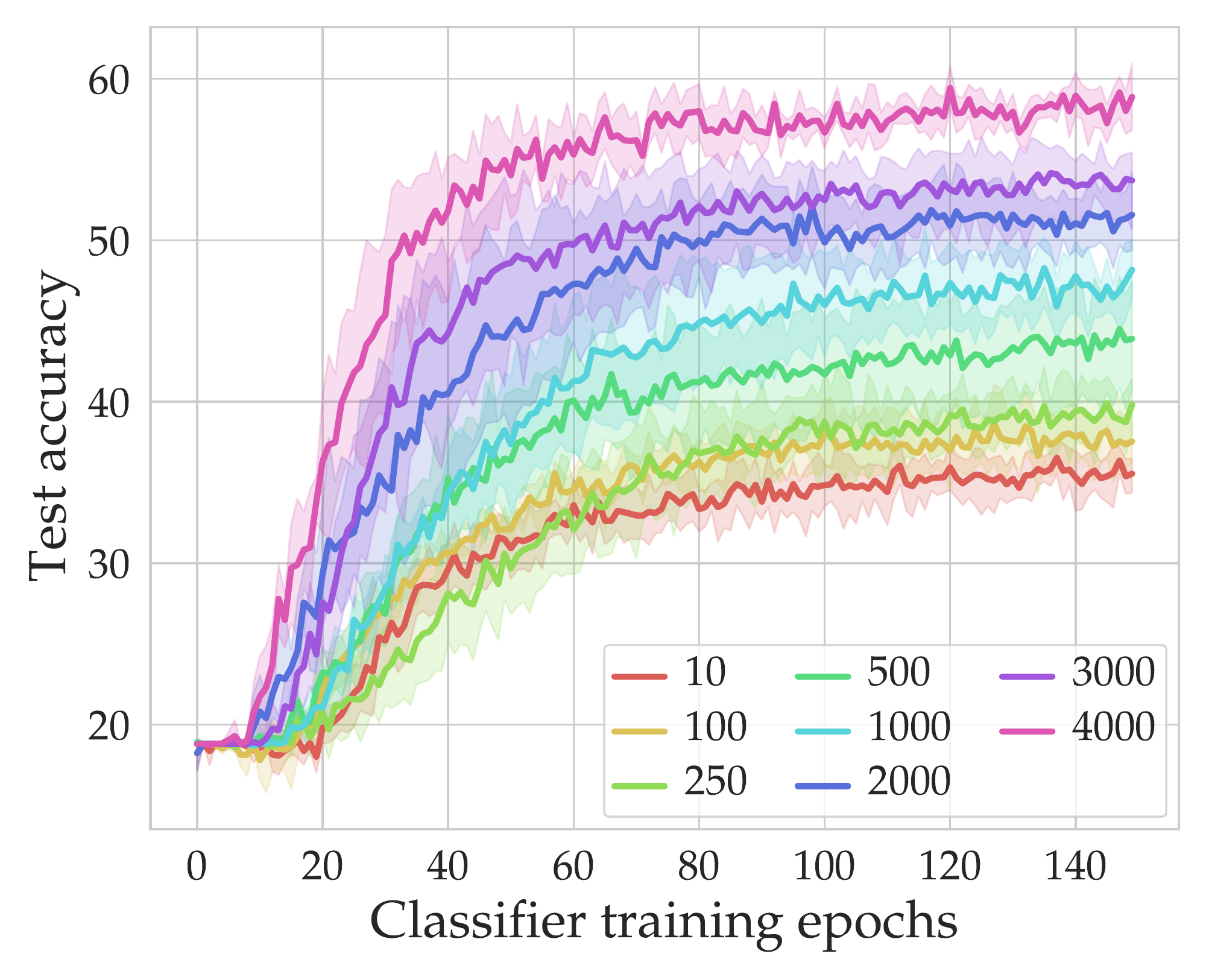}
        \caption{\textbf{MRT using models from $\mathcal{G}$.}  For each model in $\mathcal{G}$, we run MRT-10 for five trials and show the resulting test accuracy on samples from the test set from domain $B$.  Note that the robustness of the trained classifier increases as the number of training steps used to train the model increases.}
        \label{fig:better-models-plot}
    \end{subfigure}
    
    \caption[A better models implies more robustness]{\textbf{A better model implies more robustness.}  By learning a family of models $\mathcal{G}$, where each model of natural variation $G(x,\delta)$ in $\mathcal{G}$ has been trained for a different number of iterations, we show empirically that models that can more accurately reconstruct input data subject to varying natural conditions engender classifiers with higher levels of robustness.}
    \label{fig:better-models}
\end{figure}

To examine the impact of the quality of learned models of natural variation in our paradigm, we consider a task performed in Section \ref{sect:unsup-dom-adapt-experiments}, wherein we learned a model of natural variation $G(x,\delta)$ that mapped low-contrast samples from SVHN, which comprised domain $A$, to high-contrast samples from SVHN, which comprised domain $B$.  While learning this model, we saved snapshots at various points during the training procedure.  In particular, we collected a family of intermediate models $\mathcal{G}$, where the index refers to the number of MUNIT training iterations that were completed before saving the snapshot:
\begin{align*}
    \mathcal{G} = \Big\{G_{10}, G_{100}, G_{250}, G_{500}, G_{1000}, G_{2000}, G_{3000}, G_{4000}\Big\}
\end{align*}
In Figure \ref{fig:better-models-plot}, we show the result of training classifiers with MRT-10 using each model of natural variation $G\in\mathcal{G}$.  Note that the models that are trained for more training steps engender classifiers that provide higher levels of robustness against the shift in natural variation.  Indeed, as the model $G_{10}$ produces random noise, the performance of this classifier performs at effectively the level as the baseline classifier (shown in Table \ref{tab:uda-experiments}).  On the other hand, the model $G_{4000}$ is able to accurately preserve the semantic content of the input data while varying the nuisance content, and is therefore able to provide higher levels of robustness.   In other words, better models induce improved robustness for classifiers trained in the model-based paradigm.

While this experiment demonstrates that models of natural variation $G(x,\delta)$ which generate more realistic data ultimately lead to classifiers that are more robust against out-of-distribution shifts, further questions remain.  In particular, in our setting, as learned models of natural variation critically rely on the ability of deep generative models to render realistic images, there is no \emph{guarantee} that learned models will always accurately reflect natural conditions.  Indeed, it remains an open problem as to how to ensure that deep generative models generalize effectively \cite{arora2017generalization}.  To this end, one promising direction for future work is to study the conditions under which reliable models of natural variation can be learned.

\subsection{Algorithm and hyperparameter selection criteria}\label{sect:alg-selection-criteria}

\noindent\textbf{Sampling versus adversarial perspective.}  From an optimization perspective, we can group our model-based algorithms into two categories: sampling (zeroth-order) methods and adversarial (first-order) methods.  Sampling-based methods refer to those that seek to solve the inner maximization problem in equation \eqref{eq:inner-problem-general} by querying the model of natural variation $G(x,\delta)$.  This is particularly important for models that are not differentiable.  Both MRT and MDA are sampling-based (zeroth-order) methods in that they vary data by randomly sampling different nuisance parameters  $\delta\in\Delta$ for each batch in the training set.  On the other hand, MAT varies data via an adversarial (first-order) method, wherein we statistically approximate the gradient $\nabla_{\delta} \E_{(x,y)\sim\mathcal{D}}[\ell(G(x, \delta), y; w)]$ to perform the optimization; that is, we search for the worst-case nuisance parameter $\delta\in\Delta$.  Importantly, we note that while differentiability is not required in our paradigm, differentiability is required for training classifiers via MAT.

Throughout the experiments, in general we see that the sampling-based algorithms presented in this paper achieve higher levels of robustness against almost all sources of natural variation.  This finding stands in contrast to field of perturbation-based robustness, in which adversarial methods have been shown to be the most effective in improving the robustness against small, norm-bounded perturbations \cite{athalye2018obfuscated}.  Going forward,  an interesting research direction is not only to consider new algorithms but also to understand whether sampling-based or adversarial techniques provide more robustness with respect to a given model of natural variation.  

\vspace{10pt}

\noindent\textbf{The impact of $k$ in the model-based training algorithms.}  The discussion surrounding the difference between the sampling-based and adversarial mindsets that characterize our model-based algorithms is intimately related to the parameter $k$ in each of the model-based training algorithms.  Notably, in each of these algorithms, the parameter $k$ serves distinct yet related purposes.  In MRT, $k$ controls the number of nuisance parameters $\delta$ that are sampled at each iteration.  Similarly, in MAT, $k$ determines the number of steps of gradient ascent performed at each iteration.  On the other hand, in MDA, $k$ controls the number of data points that are added to the training set at each iteration.  

\begin{table}[]
    \centering
    \begin{tabular}{|c|c|c|c|} \hline
         \multirow{2}{*}{\thead{$k$}} & \multicolumn{3}{c|}{Test accuracy (top-1)} \\ \cline{2-4}
          & \thead{MAT-$k$} & \thead{MRT-$k$} & \thead{MDA-$k$} \\ \hline
          1 & 81.9 & 81.2 & 81.7 \\ \hline
          5 & 75.6 & 81.7 & 81.1 \\ \hline
          10 & 75.7 & 82.6 & 81.1 \\ \hline
          20 & 79.9 & 83.2 & 80.3 \\ \hline
          50 & 76.3 & 82.5 & 79.7 \\ \hline
    \end{tabular}
    \caption[The impact of varying $k$ in the model-based algorithms]{\textbf{The impact of varying $k$ in the model-based algorithms.}  We study the impact of varying $k$ for each of the model-based algorithms on the brightness (low$\rightarrow$high) shift on SVHN.}
    \label{tab:varying-k}
\end{table}

As we show in our experiments, each of the three model-based algorithms can be used to provide significant out-of-distribution robustness against various sources of natural variation.  In this subsection, we focus on the impact of varying the parameter $k$ in each of these algorithms.  In particular, in Table \ref{tab:varying-k}, we see that varying $k$ has a different impact for each of the three algorithms.  For MAT, we see that increasing $k$ decreases the accuracy of the trained classifier; one interpretation of this phenomenon is that larger values of $k$ allow MAT to find more challenging forms of natural variation.  On the other hand, the test accuracy of MRT improves slightly as $k$ increases.  Recall that while both MRT and MAT seek to find ``worst-case'' natural variation, MRT employs a sampling-based approach to solving the inner maximization problem as opposed to the more precise, gradient-based procedure used by MAT.  Thus the differences in the impact of varying $k$ between MAT and MRT may be due to the fact that MRT only approximately solves the inner problem at each iteration.  Finally, we see that increasing $k$ slightly decreases the test accuracy of classifiers trained with MDA.  

This study can also be used as an algorithm selection criteria.  Indeed, when data presents many modes corresponding to different levels of natural variation, it may be more efficacious to use MRT or MDA, which will observe a more diverse set of natural conditions due to their sampling-based approaches.  On the other hand, when facing a single challenging source of natural variation, it may be more useful to use MAT, which seeks to find ``worst-case,'' natural, out-of-distribution data.

\vspace{10pt}

\noindent\textbf{The dimension and radius of the nuisance space $\Delta$.}  Given a fixed instance $x\in \R^d$, $\Delta$ characterizes the set of images that can be obtained under the mapping of a model of natural variation $G(x,\delta)$.  In Section \ref{sect:geometry}, we described a geometric perspective that allowed us to rewrite the inner maximization in \eqref{eq:inner-problem-general} implicitly in terms of the learned image manifold. This representation of the inner maximization problem elucidated the fact that $\Delta$ must be rich enough to be able to produce representative images on the learned image manifold parameterized by a model of natural variation $G(x,\delta)$.  However, we note that in the extreme case when $\dim(\Delta) = d$, as is the case in much of the adversarial robustness literature, it is well known that $\delta$ is difficult to efficiently optimize over \cite{madry2017towards}.  Therefore, the dimension of $\Delta$, henceforth denoted as $\dim(\Delta)$, should be small enough so that $\Delta$ can be efficiently optimized over and large enough so that it can accurately capture the underlying source of natural variation that the model $G(x,\delta)$ describes.  In this sense, $\dim(\Delta)$ should reflect the complexity of both the source of natural variation and indeed of the data itself.

To this end, throughout our experiments, we generally scale $\dim(\Delta)$ with $d$.  For low-dimensional data (e.g. MNIST, SVHN, etc.), we found that $\dim(\Delta) = 2$ sufficed toward capturing the underlying source of natural variation effectively.  However, on datasets such as GTSRB, for which we rescaled instances into $64\times 64\times 3$ arrays, we found that $\dim(\Delta) = 8$ was more appropriate for capturing the full range of natural variation.  Indeed, on ImageNet, which contains instances of size $224\times224\times 3$, we found that $\dim(\Delta) = 8$ still produced images that captured the essence of the underlying source of natural variation.

Having studied different choices for the dimension of $\Delta$, a concomitant question is how to pick the radius of $\Delta$.  In every experiment described in this paper, we let $\Delta := \{x\in\R^q | -1 \preceq x \preceq 1\}$ where $q = \dim(\Delta)$.  This choice is not fundamental, and indeed we plan to explore varying this radius in future work.  In particular, rather than restricting $\Delta$ to be compact, one could imagine placing a distribution over the space of nuisance parameters.

\vspace{10pt}

\section{Related works}

In this section, we attempt to characterize works from a variety of fields, including adversarial robustness, generative modeling, equivariant neural networks, and domain adaptation, which have been influential in the development of the model-based robust deep learning paradigm.

\subsection{Perturbation-based adversarial robustness}

A rapidly growing body of work has addressed adversarial robustness of neural networks with respect to small  norm-bounded perturbations.  This problem has motivated an arms-race-like amalgamation of adversarial attacks and defenses within the scope of norm-bounded adversaries \cite{athalye2018obfuscated,tramer2020adaptive} and has prompted researchers to closely study the theoretical properties of adversarial robustness \cite{dobriban2020provable,schmidt2018adversarially,carmon2019unlabeled,tsipras2018robustness,javanmard2020precise}.  And while some defenses have withstood a variety of strong adversaries \cite{madry2017towards}, it remains to be seen as to whether such progress will ultimately lead to deep learning models that are reliably robust against adversarial, perturbation-based attacks.

Several notable works that propose methods for defending against adversarial attacks formulate so-called adversarial training algorithms, the goal of which is to defend neural networks against worst-case perturbations \cite{madry2017towards,wong2017provable,madaan2019adversarial,prakash2018deflecting,zhang2019theoretically,kurakin2016adversarial,moosavi2016deepfool}.  Some of the most successful works take a robust optimization perspective, in which the goal is to find the worst-case adversarial perturbation of data by solving a min-max problem \cite{madry2017towards,wong2017provable}.   However, it has also been shown that randomized smoothing based defenses are able to withstand strong attacks \cite{cohen2019certified,salman2019provably}.  In a different yet related line of work, optimization-based methods have been proposed to provide certifiable guarantees on the robustness of neural networks against small perturbations \cite{raghunathan2018certified,fazlyab2019safety,katz2019marabou,fazlyab2019efficient}.  Others have also studied how different architectural choices can result in classifiers that are more robust against adversarial examples \cite{cisse2017parseval,meng2017magnet}.  

As adversarial training methods have become more sophisticated, a range of adaptive adversarial attacks, or attacks specifically targeting a particular defense, have been proposed \cite{tramer2020adaptive}.  Prominent among the attacks on robustly-trained classifiers have been algorithms that circumvent so-called obfuscated gradients \cite{athalye2018obfuscated,carlini2017adversarial}.  Such attacks generally focus on generating adversarial examples that are perceptually similar to a given input image \cite{su2019one,dong2018boosting}.   Very recently, the authors of \cite{laidlaw2020perceptual} developed the notion of a ``perceptual adversarial attack,'' in which an adversary can corrupt a given input image subject to a learned ``neural perceptual distance.''  As far as the authors are aware, this is the strongest imperceptible adversarial attack for trained classifiers.

In summary, the commonality among all the approaches mentioned above is that they generally consider norm-bounded adversarial perturbations which are perceptually indistinguishable from true examples. Contrary to these approaches, in this work we  propose a paradigm shift from norm-bounded perturbation-based robustness to model-based robustness. To this end, we have provided training algorithms that improve robustness against natural shifts in the data distribution; such changes are often perceptually distinct from input data, as is the case for varying lighting or weather conditions.

\subsection{Generative models in the context of robustness}

Another line of work has sought to leverage deep generative models in the loop of training to facilitate strategies for generating and defending against adversarial examples.  In \cite{xiao2018generating}, \cite{lee2017generative}, and \cite{wang2019direct}, the authors propose attack strategies that use the generator from a generative adversarial network (GAN) to generate additive perturbations that can be used to attack a classifier.  On the other hand, a framework called DefenseGAN, which uses a Wasserstein GAN to ``denoise'' adversarial examples \cite{samangouei2018defense}, has been proposed to defend against perturbation-based attacks.  This defense method was later broken by the Robust Manifold Defense \cite{jalal2017robust}, which searches over the parameterized manifold induced by a generative model to find worst-case perturbations of data.  The min-max formulation used in this work is analogous to the PGD defense~\cite{madry2017towards}, and was foundational in our development of the model-based robust deep learning paradigm.

Closer to the approach we describe in this paper are works that use deep generative models to generate adversarial inputs themselves, rather than generating small perturbations.  The authors of \cite{schott2018towards} and \cite{zhao2017generating} use the generator from a GAN to generate adversarial examples that obey Euclidean norm-based constraints.  Alternatively, the authors of \cite{naseer2019cross} use GANs to generate adversarial patterns that can be used to construct adversarial examples in multiple domains.  Similarly, \cite{dunn2019generating} and \cite{wang2019gan} generate unrestricted adversarial examples, or instances that are not subjected to a norm-based constraint \cite{song2018constructing}, via a deep generative model.  Finally, two recent papers \cite{vandenhende2019three,arruda2019cross} use a GAN to perform data-augmentation by generating perceptually realistic samples.

In this work, we use generative networks to learn and model the natural variability within data. This is indeed different than generating norm-based adversarial perturbations or perceptually realistic adversarial examples, as have already been considered in the literature. Our generative models are designed to capture natural shifts in data, whereas the relevant literature has sought to create synthetic adversarial perturbations that fool neural networks.     

\subsection{A broader view of robustness in deep learning}

Aside from the algorithms we introduced in Section \ref{sect:algorithms}, we are not aware of any other algorithms that can be used to address out-of-distribution robustness across the diverse array of tasks presented in Section \ref{sect:experiments}.  However, several lines of research have sought to address this problem in constrained settings or under highly restrictive assumptions.  Of note are works that have sought to provide robustness against specific transformations of data that are more likely to be encountered in applications than norm-bounded perturbations.  Transformations that have recently received attention from the adversarial robustness community include adversarial quilting \cite{guo2017countering}, adversarial patches and clothing \cite{wu2019making}, geometric transformations \cite{engstrom2017exploring,balunovic2019certifying,kamath2020invariance,kanbak2018geometric}, distortions \cite{athalye2017synthesizing}, deformations and occlusions \cite{wang2017fast}, and nuisances encountered by unmanned aerial vehicles \cite{wu2019delving}.
In response to these works and motivated by myriad safety-critical applications, first steps toward robust defenses against specific distributional shifts have recently been proposed \cite{dunn2019generating,song2018constructing,vandenhende2019three,arruda2019cross}.  The resulting methodologies generally leverage properties specific to the transformation of interest.  

Another line of work has sought to use various mechanisms to create semantically-realistic adversarial examples via more general frameworks.  For example, in \cite{hosseini2018semantic}, the authors seek to create ``semantic adversarial examples'' via color-based shifts, which relates to the perspective advocated for in \cite{dreossi2018semantic}, in which the authors argue that system semantics and specifications should be considered when generating meaningful disturbances in the data.  Similarly, in \cite{jain2019generating}, the authors use differentiable renderers to generate semantically meaningful changes.  In the same spirit, the idea in \cite{jacobsen2018excessive} is to leverage an information theoretic approach to edit the nuisance content of images to create perceptually realistic data that causes misclassification.  

While this progress has helped to motivate new notions of robustness, defenses against specific threat models are limited  in the sense that they often cannot be generalized to develop a learning paradigm that is broadly applicable across different forms of natural variation.  This contrasts with the motivation behind this paper, which is to provide general robust training algorithms that can improve the robustness of trained neural networks across a variety of scenarios and applications.

More related to the current work are two concurrent papers that formulate robust training procedures under the assumption that data is corrupted according to a fixed generative architecture.  The authors of \cite{gowal2020achieving} exploit properties specific to the StyleGAN \cite{karras2019style} architecture to formulate a training algorithm that provides robustness against color-based shifts on MNIST and CelebA \cite{liu2015faceattributes}.  In our work, we propose a more general framework and three novel robust training algorithms that can exploit any suitable generative model, and we show improvements on more challenging, naturally-occurring shifts across twelve distinct datasets. The authors of \cite{wong2020learning} use conditional VAEs to learn perturbation sets corresponding to simple corruptions from pairs of images.  In our framework we improve robustness against more challenging, natural shifts by learning from \emph{unpaired} datasets and we do not rely on class-conditioning to generate realistic images.

\subsection{Domain adaptation and domain generalization}

In the domain adaptation literature, various methods have been proposed which rely on the restrictive assumption that unlabeled data corresponding to a fixed distributional shift is available during training \cite{ajakan2014domain,ganin2015unsupervised,saito2017maximum,daume2009frustratingly}.  Several works in this vain use an adversarial min-max formulation to adapt a classifier trained on a source domain to perform well on a related target domain, for which labels are unavailable at training time \cite{tzeng2017adversarial,hoffman2018cycada,long2018conditional,pei2018multi}.  We note that these ``adversarial'' methods differ from the model-based paradigm we introduce in that rather than formulating the problem of adapting a classifier to perform well on the target domain, we employ a min-max procedure to search for worst-case shifts in data for a \emph{fixed} generative model.  Indeed, the main difference between domain adaptation techniques and our paradigm is that our solution does not assume access to unlabeled data from a fixed shift and can be applied to datasets that are entirely unseen during training.  

Also related is the field of domain generalization \cite{blanchard2011generalizing,muandet2013domain,li2018deep}, in which one assumes access to a variety of training domains, all of which are related to an unseen target domain on which the trained classifier is ultimately evaluated.  Such works often rely on transfer learning \cite{blanchard2017domain} and distribution matching \cite{albuquerque2019adversarial} to improve classification accuracy on the unseen domain.  While the experiments in Sections \ref{sect:transfer-experiments} tackle a similar problem, in which knowledge from one domain is used to learn a classifier that performs well on an unseen test domain, the experiments in the other subsections of Section \ref{sect:experiments} show that our model-based paradigm is much more broadly applicable that techniques domain generalization techniques.

\section{Conclusion and future directions} \label{sec:conclusions}

In this paper, we formulated a novel problem that concerns the robustness of deep learning with respect to natural, out-of-distribution shifts in data.  Motivated by perceptible nuisances in computer vision, such as lighting or weather changes, we propose the novel {\em model-based robust deep learning paradigm}, the goal of which is to improve the robustness of deep learning systems with respect to naturally occurring conditions.  This notion of robustness offers a departure from that of norm-bound, perturbation-based adversarial robustness, and indeed our optimization-based formulation results in a new family of training algorithms that can be used to train neural networks to be robust against arbitrary forms of natural variation.  Across a range of diverse experiments and across twelve distinct datasets, we empirically show that the model-based paradigm is broadly applicable to a variety of challenging domains.

Our model-based robust deep learning paradigm open numerous directions for future work.  In what follows, we briefly highlight several of these broad directions.

\vspace{10pt}

\noindent\textbf{Learning a library of nuisance models.}  One natural future direction is to determine superior ways of learning models of natural variations to perform model-based training.  In this paper, we used the MUNIT framework \cite{huang2018multimodal}, but other existing architectures may be better suited for specific forms of natural variation or for different datasets.  Indeed, a more rigorous statistical analysis of problem \eqref{eq:stat-learn-G} may lead to the discovery of new architectures designed specifically for model-based training.  To this end, recent work that concerns learning invariances in neural networks may provide insight into learning physically meaningful models \cite{van2018learning,benton2020learning}.  Beyond computer vision, learning such models in other domains such as robotics would enable new applications.

\vspace{10pt}

\noindent\textbf{Model-based algorithms and architectures.} Another important direction involves the development of new algorithms for solving the min-max formulation of \eqref{eq:min-max-general}.  In this paper, we presented three algorithms -- MRT, MDA, and MAT -- that can be used to approximately solve this problem, but other algorithms are possible and may result in higher levels of robustness.  In particular, adapting first-order methods to search globally over the learned image manifold may provide more efficient, scalable, or robust results.   Indeed, one open question is whether it is necessary to decouple the procedures used to learn classifiers and models of natural variation.  Another interesting direction is to follow the line of work that has developed equivariant neural network architectures toward designing new architectures that are invariant to models of natural variation.

\vspace{10pt}

\noindent\textbf{Applications beyond image classification.} Throughout the paper, we have empirically demonstrated the utility of our approach across  many image classification tasks.  However, our model-based paradigm could also be broadly applied to further applications, both in computer vision as well as outside computer vision.  Within computer vision, one can consider other tasks such as semantic segmentation in the presence of challenging natural conditions.   Outside computer vision, one exciting area is to exploit physical models of robot dynamics with deep reinforcement learning for applications such as walking in unknown terrains.  In any domain where one has access to suitable models of natural variation, our approach allows domain experts to leverage these models in order to make deep learning far more robust.

\vspace{10pt}

\noindent\textbf{Theoretical foundations.} Finally, we believe that there are many exciting open questions with respect to the theoretical aspects of model-based robust training.  What types of models provide significant robustness gains in our paradigm?  How accurate does a model need to be to engender neural networks that are robust against natural variation and out-of-distribution shifts?  We would like to address such theoretical questions from geometric, physical, and statistical perspectives with an eye toward developing faster algorithms that are both more sample-efficient and that provide higher levels of robustness.  A deeper theoretical understanding of our model-based robust deep learning paradigm could result in new approaches that blend model-based and perturbation-based methods and algorithms. 
\section*{Acknowledgements}  

This work has been partially supported by the Defense Advanced Research Projects Agency (DARPA) Assured Autonomy under Contract No. FA8750-18-C-0090, AFOSR under grant FA9550-19-1-0265 (Assured Autonomy in Contested Environments), NSF CPS 1837210, and ARL CRA DCIST W911NF-17-2-0181 program.

\section*{Authors}

All authors are with the Department of Electrical and Systems Engineering, University of Pennsylvania, Philadelphia, PA 19104. The authors can be reached at the following email addresses: 
\begin{align*}
    \texttt{\{arobey1, hassani, pappasg\}@seas.upenn.edu}.
\end{align*}

\newpage

\bibliographystyle{unsrt}
\bibliography{bibliography}

\newpage

\begin{appendix}
\section{Training details} \label{app:training-details}

In this appendix, we provide details concerning our implementation of both the model-based robust deep learning algorithms as well as models of natural variation.  We note that all experiments described in Sections \ref{sect:experiments} and \ref{sect:discussion} were run on four NVIDIA RTX 5000 GPUs.  

\subsection{Baseline and model-based implementation details}

\noindent\textbf{PGD.}  When training classifiers with PGD \cite{madry2017towards}, we use a perturbation budget of $\epsilon = 8/255$, a step size of $\alpha=0.01$, and twenty iterations of gradient ascent per batch.  Our implementation of PGD is included along with our implementation of the model-based robust deep learning algorithms; this implementation can be found at \url{https://github.com/arobey1/mbrdl}.

\vspace{10pt}

\noindent\textbf{ADDA.}  We used the following implementation of ADDA \cite{tzeng2017adversarial}: \url{https://github.com/Carl0520/ADDA-pytorch}.  We trained each classifier for 100 epochs on the source domain, and we then trained for a further 100 epochs when adaptating the weights of the target encoder.  We used the LeNet \cite{lecun1998gradient} for the encoder networks; to ensure a fair comparison, we used two convolutional layers and two feed-forward layers, which is the same as the classification networks used for model-based training.  

\vspace{10pt}

\noindent\textbf{AugMix.}  We used the following implementation of AugMix \cite{hendrycks2019augmix}: \url{https://github.com/google-research/augmix}.  We used the default parameters for training described in this repository.  

\vspace{10pt}

\noindent\textbf{Model-based algorithms.}  All three model-based algorithms (MRT, MAT, and MDA) are implemented in our repository: \url{https://github.com/arobey1/mbrdl}.  Throughout the experiments, unless stated otherwise, we ran MRT with $k=10$, MAT with $k=10$, and MDA with $k=3$.  We also used a trade-off parameter of $\lambda=1$.

\subsection{Classifier training details}

Here we use the following conventions for describing architectures.  \texttt{c32-3} refers to a $2D$ convolutional operator with 32 kernels, each of which has shape $3\times 3$.  \texttt{p2} refers to a max-pooling layer with kernel size 2.  \texttt{d0.25} refers to a dropout layer, which drops an activation with probability $0.25$.  \texttt{flat} refers to a flattening layer.  \texttt{fc-128} refers to a fully-connected layer mapping into $\R^{128}$.

When training classifiers for MNIST, Q-MNIST, E-MNIST, K-MNIST, Fashion-MNIST, USPS, SVHN, GTSRB, and CURE-TSR we use a simple CNN architecture with two convolutional layers and two feed forward layers.  More specifically, the architecture we used the following architecture:
\begin{align*}
     \texttt{c32-3, c64-3, p2, c128-3, p2, d0.25, flat, fc128, d0.5, fc10}.
\end{align*}
For each of these experiments, we use the Adadelta \cite{zeiler2012adadelta} optimizer with a learning rate of 1.0.  We also use a batch size of 64.  Images from MNIST, Q-MNIST, E-MNIST, K-MNIST, Fashion-MNIST, USPS, SVHN, and CURE-TSR are resized to $32\times 32\times 3$; for grayscale datasets such as MNIST, we repeat the channels three times.  Images from GTSRB are resized to $64\times 64\times 3$.  We train each classifier for 100 epochs.  

When training on ImageNet, we use the ResNet-50 \cite{he2016deep} architecture.  We note that architectural choices are possible and will be explored in future work.  For each of the experiments on ImageNet, we use SGD with an initial learning rate of 0.05; we decay the learning rate linearly to 0.001 over 100 epochs.  We use a batch size of 64 for ERM and PGD.  For the model-based algorithms, we use a batch size of 32, as these algorithms require the GPU to store multiple copies of each image during each training iteration.

\subsection{MUNIT framework overview}

For completeness, we give a brief overview of the MUNIT framework \cite{huang2018multimodal}.  To begin, let $x_A \in A$ and $x_B \in B$ be images from two unpaired image domains $A$ and $B$; in the notation of the previous section, we assume that these images are drawn from two marginal distributions $\Prob_A$ and $\Prob_B$.  Further, the MUNIT model assumes that each image from either domain can be decomposed into two components: a style code $s$ that contains information about factors of natural or nuisance variation, and a content code $c$ that contains information about higher level features such as the label of the image.  Further, it is assumed that the content codes for images in either domain are drawn from a common set $\mathcal{C}$, but that the style codes are drawn from domain specific sets $\mathcal{S}_A$ and $\mathcal{S}_B$.  In this way, a pair of corresponding images $(x_A, x_B)$ are of the form $x_A = \text{Dec}_A(c, s_A)$ and $x_B = \text{Dec}_B(c, s_B)$, where $c\in \mathcal{C}$, $s_A\in \mathcal{S}_A$, $s_B\in\mathcal{S}_B$, and where $\text{Dec}_A$ and $\text{Dec}_B$ are unknown decoding networks corresponding to domains $A$ and $B$ respectively.  The authors of \cite{huang2018multimodal} call this setting a partially shared latent space assumption.

The MUNIT model consists of an encoder-decoder pair $(\text{Enc}_A, \text{Dec}_A)$ and $(\text{Enc}_B, \text{Dec}_B)$ for each image domain $A$ and $B$.  These encoder-decoder pairs are trained to learn a mapping that reconstructs its input.  That is, $x_A \approx \text{Dec}_A(\text{Enc}_A(x_A))$ and $x_B \approx \text{Dec}_B(\text{Enc}_B(x_B))$.  More specifically, $\text{Enc}_A : \mathcal{A} \rightarrow \mathcal{C}\times \mathcal{S}_A$ is trained to encode $x_A$ into a content code $c\in\mathcal{C}$ and a style code $s_A\in\mathcal{S_A}$.  Similarly, $\text{Enc}_B : \mathcal{B} \rightarrow \mathcal{C} \times \mathcal{S}_B$ is trained to encode $x_B$ into $c\in\mathcal{C}$ and $s_B\in\mathcal{S}_B$.  Then the decoding networks $\text{Dec}_A : \mathcal{C} \times \mathcal{S}_A \rightarrow A$ and $\text{Dec}_B : \mathcal{C} \times \mathcal{S}_B \rightarrow B$ are trained to reconstruct the encoded pairs $(c, s_A)$ and $(c, s_B)$ into the respective images $x_A$ and $x_B$.  

Inter-domain image translation is performed by swapping the decoders.  In this way, to map an image $x_A$ from $A$ to $B$, $x_A$ is first encoded into $\text{Enc}_A(x_A) = (c, s_A)$.  Then, a new style vector $s_B$ is sampled from $\mathcal{S}_B$ from a prior distribution $\pi_B$ on the set $\mathcal{S}_B$ and the translated image $x_{A\rightarrow B}$ is equal to $\text{Dec}_B(c, s_B)$.  The translation of $x_B$ from $B$ to $A$ can be described via a similar procedure with $\text{Enc}_B$, $\text{Dec}_A$, and a prior $\pi_A$ supported on $\mathcal{S}_A$.  In this paper, we follow the convention used in \cite{huang2018multimodal} as use a Gaussian distribution for both $\pi_A$ and $\pi_B$ with zero mean and an identity covariance matrix.

Training an MUNIT model involves considering four loss terms.  First, the encoder-decoder pairs $(\text{Enc}_A, \text{Dec}_A)$ and $(\text{Enc}_B, \text{Dec}_B)$ are trained to reconstruct their inputs my minimizing the following loss:
\begin{align*}
    \ell_{\text{recon}} = \E_{x_A\sim\Prob_A} \norm{\text{Dec}_A(\text{Enc}_A(x_A)) - x_A}_1 + \E_{x_B\sim P_B} \norm{\text{Dec}_B(\text{Enc}_B(x_B)) - x_B}_1
\end{align*}
Further, when translating an image from one domain to another, the authors of \cite{huang2018multimodal} argue that we should be able to reconstruct the style and content codes.  By rewriting the encoding networks as $\text{Enc}_A(x_A) = (\text{Enc}_A^c(x_A), \text{Enc}_A^s(x_A))$ and $\text{Enc}_B(x_B) = (\text{Enc}_B^c(x_B), \text{Enc}_B^s(x_B))$, the constraint on the content codes can be expressed in the following way:
\begin{align*}
    \ell_{\text{recon}}^c = \E_{\substack{c_A\sim \Prob(c_A) \\ s_B\sim \pi_B}}\norm{ \text{Enc}_B^c(\text{Dec}_B(c_A, s_B)) - c_A}_1 + \E_{\substack{c_B\sim \Prob(c_B) \\ s_A\sim \pi_A}}\norm{ \text{Enc}_A^c(\text{Dec}_A(c_B, s_A)) - c_B}_1
\end{align*}
where $\Prob(c_A)$  is the distribution given by $c_A = \text{Enc}_A^c(x_A)$ where $x_A\sim \Prob_A$ and $\Prob(c_B)$ is the distribution given by $c_B = \text{Enc}_B^c(x_B)$ where $x_B\sim\Prob_B$.  Similar, the constraint on the style codes can be written as
\begin{align*}
    \ell_{\text{recon}}^s = \E_{\substack{c_A\sim \Prob(c_A) \\ s_B\sim \pi_B}}\norm{ \text{Enc}_B^s(\text{Dec}_B(c_A, s_B)) - s_B}_1 + \E_{\substack{c_B\sim \Prob(c_B) \\ s_A\sim \pi_A}}\norm{ \text{Enc}_A^s(\text{Dec}_A(c_B, s_A)) - s_A}_1.
\end{align*}
Finally, two GANs corresponding to the two domains $A$ and $B$ are used to form an adversarial loss term.  The GANs use the decoders $\text{Dec}_A$ and $\text{Dec}_B$ as the respective generators for domains $A$ and $B$.  By denoting the discriminators for these domains by $D_A$ and $D_B$, we can write the GANs as $(\text{Dec}_A, D_A)$ and $(\text{Dec}_B, D_B)$.  In this way, the final loss term takes the following form:
\begin{align*}
    \ell_{\text{GAN}} &= \E_{\substack{c_A\sim \Prob(c_A) \\ s_B\sim \pi_B}}  \left[ \log\left( 1 - D_B(\text{Dec}_B(c_A, s_B)) \right)\right] + \E_{x_B\sim \Prob_B} [\log D_B(x_B)] \\
    &\qquad + \E_{\substack{c_B\sim \Prob(c_B) \\ s_A\sim \pi_A}}  \left[ \log\left( 1 - D_A(\text{Dec}_A(c_B, s_A)) \right)\right] + \E_{x_A\sim \Prob_A} [\log D_A(x_A)]
\end{align*}

Using the four loss terms we have described, the MUNIT framework uses first-order methods to solve the following nonconvex optimization problem:
\begin{align*}
    \min_{\substack{\text{Enc}_A, \text{Enc}_B \\ \text{Dec}_A, \text{Dec}_B}} \max_{D_1, D_2} \quad \ell_{\text{GAN}} + \lambda_x \ell_{\text{recon}} + \lambda_c\ell_{\text{recon}}^c + \lambda_s \ell_{\text{recon}}^s
\end{align*}

\subsection{Hyperparameters and implementation of MUNIT}

In Table \ref{tab:munit-params}, we record the hyperparameters we used for training models of natural variation via the MUNIT framework.  The hyperparameters we selected are generally in line with those suggested in \cite{huang2018multimodal}.  We use the same architectures for the encoder, decoder, and discriminative networks as are described in Appendix B.2 of \cite{huang2018multimodal}.

\begin{table}
    \centering
    \begin{tabular}{|c|c|} \hline
        \textbf{Name} & \textbf{Value} \\ \hline
         Batch size & 1 \\ \hline
         Weight decay & 0.0001 \\ \hline
         Learning rate & 0.0001 \\ \hline
         Learning rate policy & Step \\ \hline
         $\gamma$ (learning rate decay amount) & 0.5 \\ \hline
         $\lambda_x$ (image reconstruction coefficient) & 10 \\ \hline
         $\lambda_c$ (content cycle-consistency coefficient) & 1 \\ \hline
         $\lambda_s$ (style cycle-consistency coefficient) & 1 \\ \hline
    \end{tabular}
    \caption[MUNIT hyperparameters]{\textbf{MUNIT hyperparameters.}}
    \label{tab:munit-params}
\end{table}
\newpage
\section{A gallery of learned models of natural variation} \label{app:gallery-of-learned-models}

We conclude this section by showing images corresponding to the many distributional shifts used in the experiments section.  Furthermore, we show images generated by passing domain $A$ images through learned models of natural variation.

\mbimagesfig{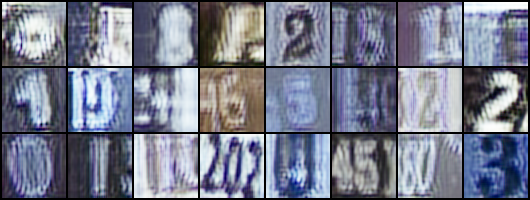}{SVHN contrast (low$\rightarrow$high).}{SVHN contrast (low$\rightarrow$high)}

\mbimagesfig{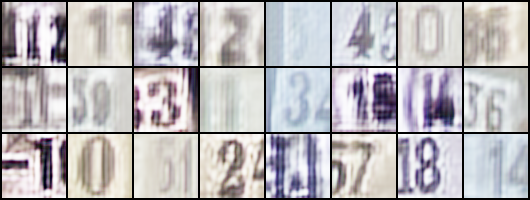}{SVHN brightness (low$\rightarrow$high).}{SVHN brightness (low$\rightarrow$high)}

\mbimagesfig{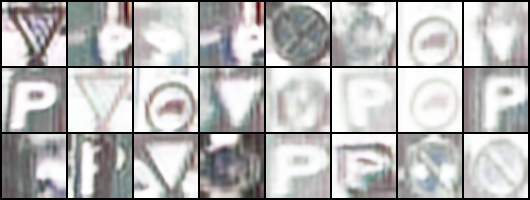}{CURE-TSR snow (no$\rightarrow$yes).}{CURE-TSR snow (no$\rightarrow$yes)}

\mbimagesfig{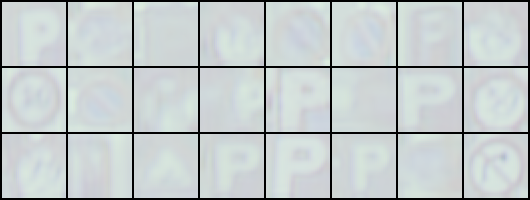}{CURE-TSR haze (no$\rightarrow$yes).}{CURE-TSR haze (no$\rightarrow$yes)}

\mbimagesfig{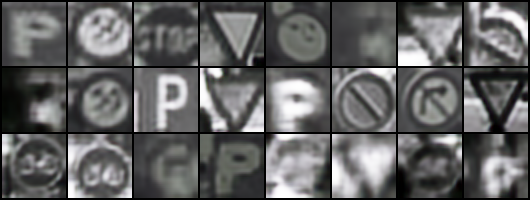}{CURE-TSR decolorization (no$\rightarrow$yes).}{CURE-TSR decolorization (no$\rightarrow$yes)}

\mbimagesfig{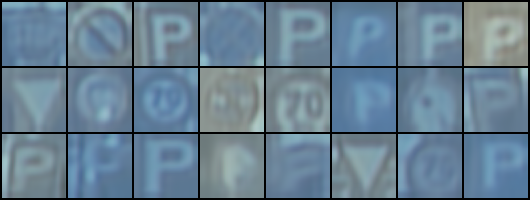}{CURE-TSR rain (no$\rightarrow$yes).}{CURE-TSR rain (no$\rightarrow$yes)}

\mbimagesfig{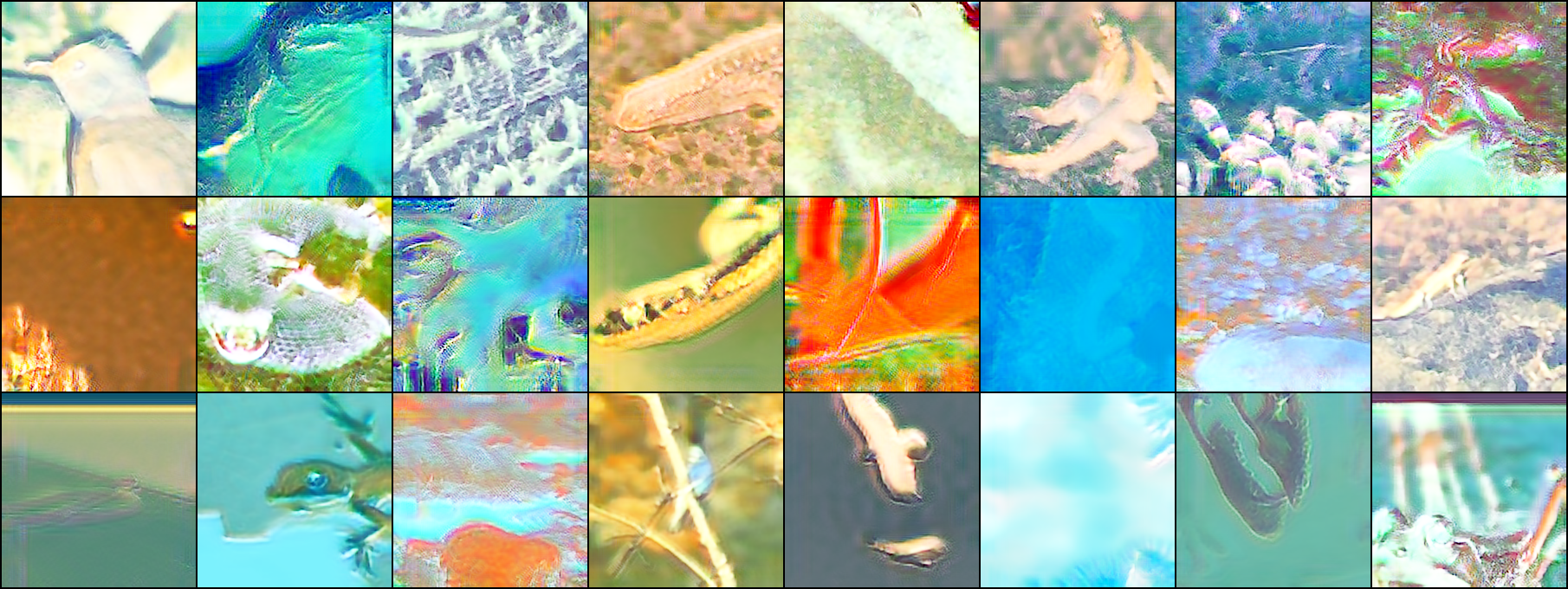}{ImageNet brightness (low$\rightarrow$high).}{ImageNet brightness (low$\rightarrow$high)}

\mbimagesfig{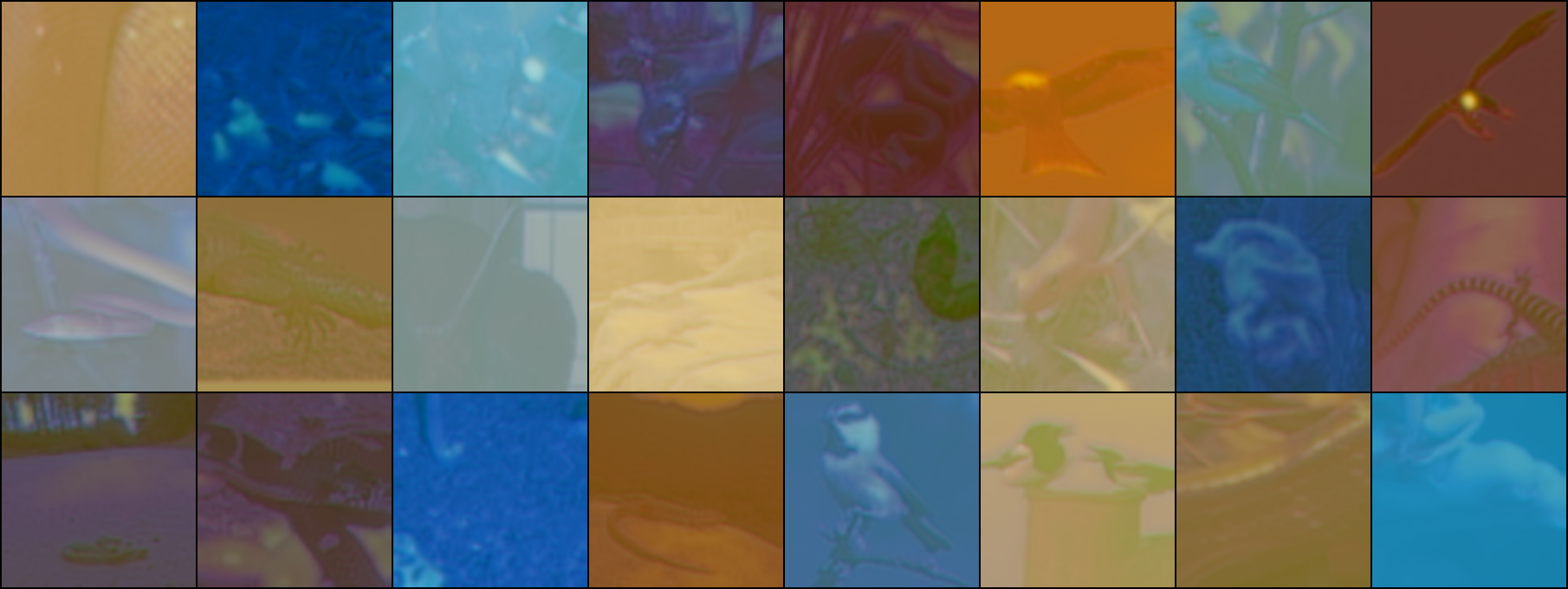}{ImageNet contrast (high$\rightarrow$low).}{ImageNet contrast (high$\rightarrow$low)}

\mbimagesfig{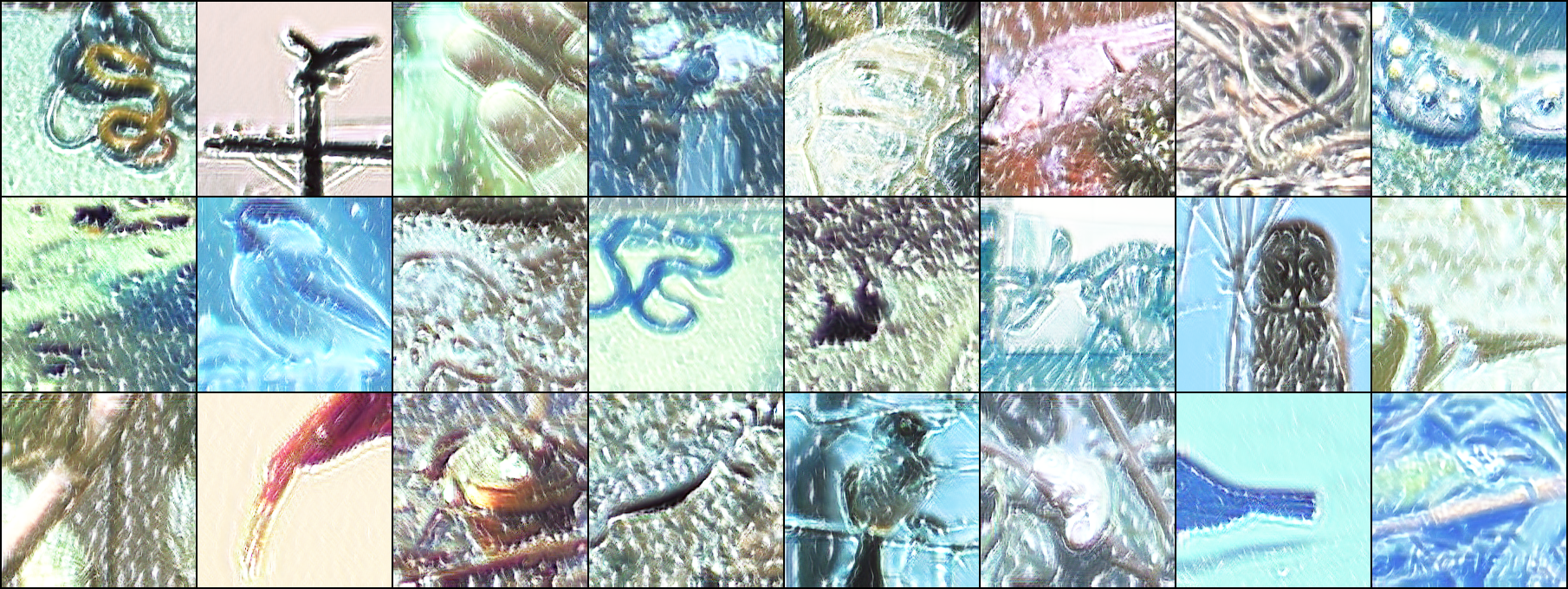}{ImageNet snow (no$\rightarrow$yes).}{ImageNet snow (no$\rightarrow$yes)}

\mbimagesfig{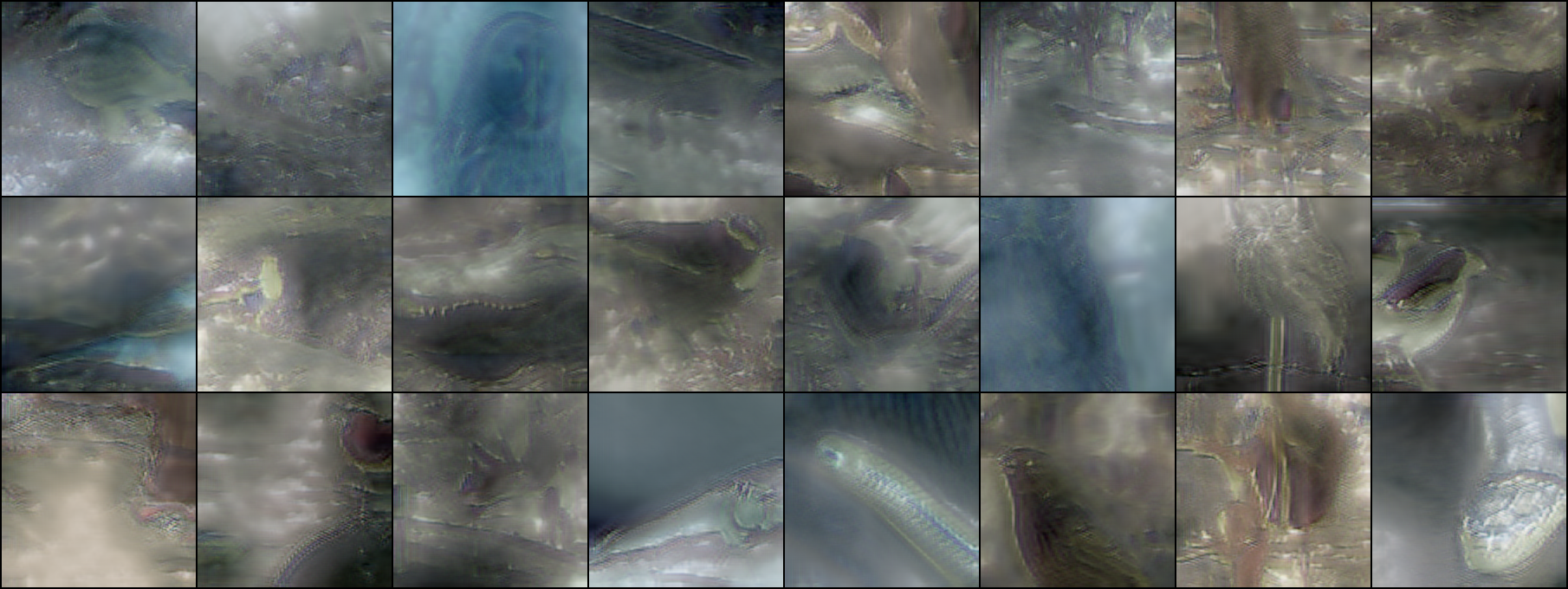}{ImageNet fog (no$\rightarrow$yes).}{ImageNet fog (no$\rightarrow$yes)}

\mbimagesfig{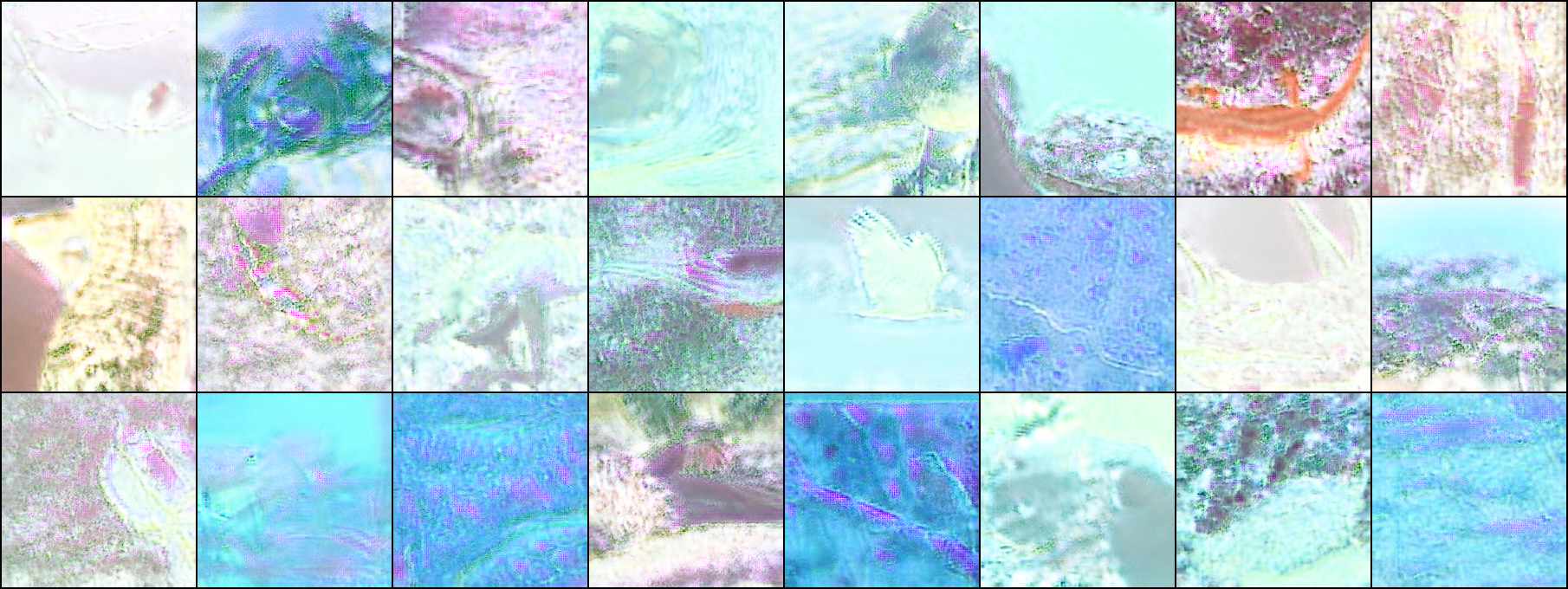}{ImageNet frost (no$\rightarrow$yes).}{ImageNet frost (no$\rightarrow$yes)}

\mbimagesfig{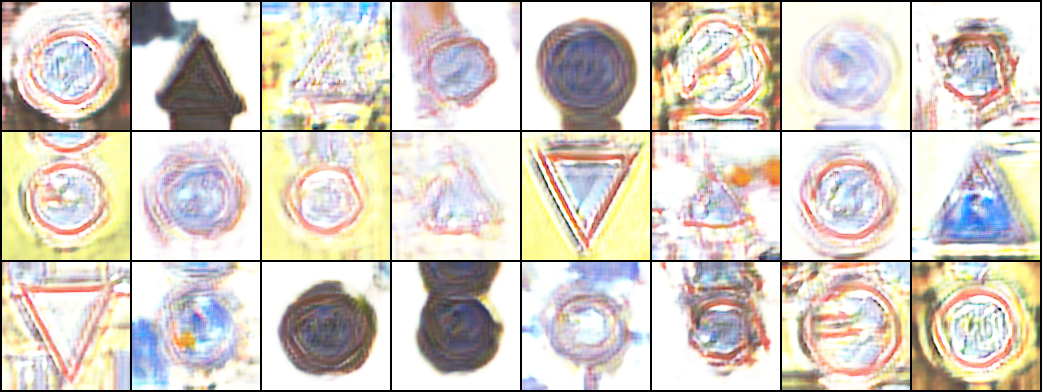}{GTSRB brightness (low$\rightarrow$high).}{GTSRB brightness (low$\rightarrow$high)}

\mbimagesfig{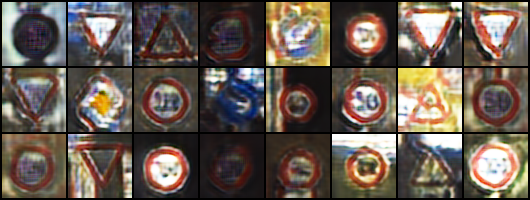}{GTSRB contrast (low$\rightarrow$high).}{GTSRB contrast (low$\rightarrow$high)}

\begin{table}
    \centering
    \begin{tabular}{|C{2cm}|c|c|c|c|c|} \cline{2-6}

         \multicolumn{1}{c|}{} & \multicolumn{5}{|c|}{\makecell{\textbf{Dataset} $\mathcal{D}_2$}} \\ \cline{2-6} 
    
          \multicolumn{1}{c|}{} & \thead{QMNIST} & \thead{EMNIST} & \thead{KMNIST} & \thead{Fashion-MNIST} & \thead{USPS}\\ \hline
         Images from $\mathcal{D}_2$ & 
         \begin{minipage}{2.1cm}
            \centering
            \vspace{5pt}
            \includegraphics[width=2cm]{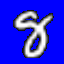} 
            \vspace{5pt}
         \end{minipage}
         & 
         \begin{minipage}{2.1cm}
            \centering
            \vspace{5pt}
            \includegraphics[width=2cm]{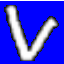} 
            \vspace{5pt}
        \end{minipage}
        & 
        \begin{minipage}{2.1cm}
            \centering
            \vspace{5pt}
            \includegraphics[width=2cm]{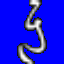} 
            \vspace{5pt}
        \end{minipage}
        & 
        \begin{minipage}{2.1cm}
            \centering
            \vspace{5pt}
            \includegraphics[width=2cm]{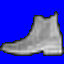}
            \vspace{5pt}
        \end{minipage}
         & 
        \begin{minipage}{2.1cm}
            \centering
            \vspace{5pt}
            \includegraphics[width=2cm]{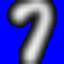}
            \vspace{5pt}
        \end{minipage} \\ \hline
        
         Model-based images &
        \begin{minipage}{2.1cm}
            \centering
            \vspace{5pt}
            \includegraphics[width=2cm]{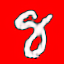}
            \vspace{5pt}
        \end{minipage}
        & 
        \begin{minipage}{2.1cm}
            \centering
            \vspace{5pt}
            \includegraphics[width=2cm]{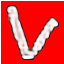}
            \vspace{5pt}
        \end{minipage}
         & 
         \begin{minipage}{2.1cm}
            \centering
            \vspace{5pt}
            \includegraphics[width=2cm]{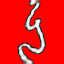}
            \vspace{5pt}
        \end{minipage}
         & 
         \begin{minipage}{2.1cm}
            \centering
            \vspace{5pt}
            \includegraphics[width=2cm]{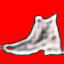}
            \vspace{5pt}
        \end{minipage}
        & 
        \begin{minipage}{2.1cm}
            \centering
            \vspace{5pt}
            \includegraphics[width=2cm]{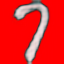}
            \vspace{5pt}
        \end{minipage} \\ \hline
    \end{tabular}
    \caption[Passing images from other datasets through a model learned on MNIST]{\textbf{Passing images from other datasets through a model learned on MNIST.}  The first row of images in this table are samples taken from colorized versions of Q-MNIST, E-MNIST, K-MNIST, Fashion-MNIST, and USPS.  The second row of images shows samples passed through a model trained on the original MNIST dataset to change the background color from blue to red.}
    \label{tab:mnist-model-transfer}
\end{table}
\newpage
\section{Details concerning datasets and domains} \label{app:datasets}

As mentioned in Section \ref{sect:experiments}, we used twelve different datasets in this work to fully evaluate the efficacy of the model-based algorithms we introduced in Section \ref{sect:algorithms}.  For several of these datasets, we curated subsets corresponding to different factors of natural variation, which we refer to as \textit{domains}.

\subsection{Natural vs.\ synthetic variation in data}

Throughout our experiments, we demonstrate that our methods are able to provide robustness against many challenging sources of natural variation. Furthermore, our experiments contain domains with both \emph{naturally-occurring} and \emph{artificially-generated} variation.  Notably, every experiment involving data from SVHN or GTSRB used naturally-occurring variation.  In what follows, we discuss both of these categories.

\subsubsection{Naturally-occurring variation}

Throughout the experiments, we used data from SVHN and GTSRB to train neural networks to be robust against contrast and brightness variation.  To extract naturally-occurring variation from these datasets, we used simple metrics to threshold the data into subsets corresponding to different levels of natural variation.  Specifically, we defined the brightness $\mathcal{B}(x)$ of an RGB image $x$ to be the mean pixel value of $x$, and we define the contrast $\mathcal{C}(x)$ to be the difference between the largest and smallest pixel values.  Table \ref{tab:svhn-gtsrb-nuisances} show the thresholds we chose for contrast and brightness on SVHN and GTSRB.  Note that these thresholds were chosen somewhat subjectively to reflect our perception of low, medium and high values of brightness and contrast.  We intend to experiment with different thresholds in future work.

\begin{figure}
    \centering
    \begin{subfigure}{0.45\textwidth}
        \includegraphics[width=0.9\textwidth]{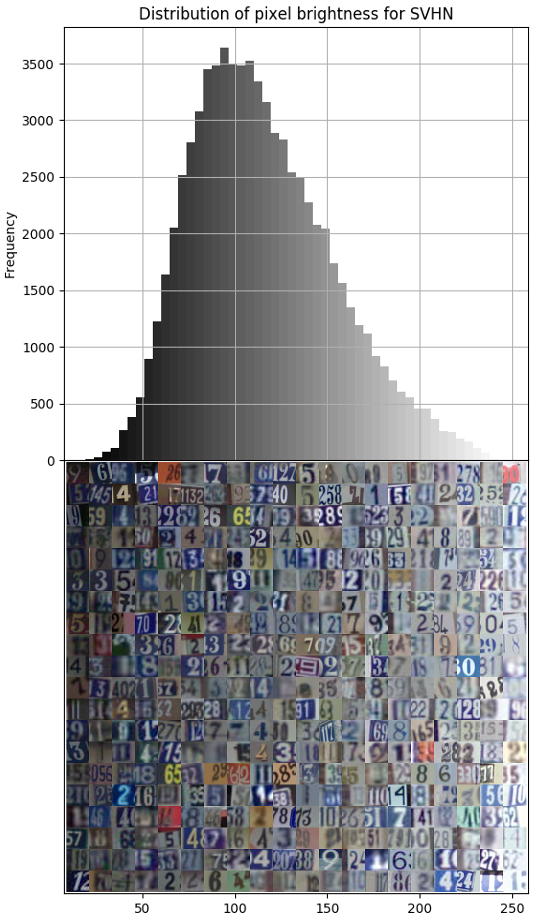}
        \caption{\textbf{SVHN brightness histogram.}  The histogram shows the distribution of pixel brightness for SVHN.  The images below the histogram correspond to the bins of the histogram, meaning samples to the left have low brightness whereas samples further to the right have higher brightness.}
        \label{fig:svhn-brightness-hist}
    \end{subfigure} \quad
    \begin{subfigure}[t!]{0.50\textwidth}
        \begin{subfigure}{\textwidth}
            \centering
            \includegraphics[width=0.8\textwidth]{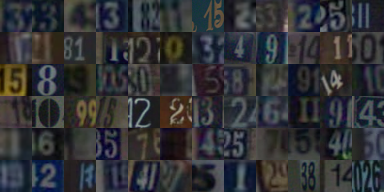}
            \caption{\textbf{Low brightness samples.}}
            \label{fig:svhn-low-brightness}
        \end{subfigure}\vspace{5pt}
        \begin{subfigure}{\textwidth}
            \centering
            \includegraphics[width=0.8\textwidth]{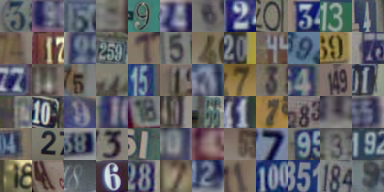}
            \caption{\textbf{Medium brightness samples.}}
            \label{fig:svhn-medium-brightness}
        \end{subfigure}\vspace{5pt}
        \begin{subfigure}{\textwidth}
            \centering
            \includegraphics[width=0.8\textwidth]{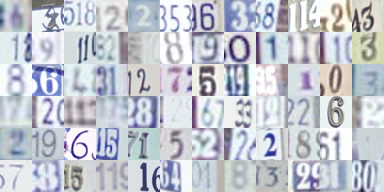}
            \caption{\textbf{High brightness samples.}}
            \label{fig:svhn-high-brightness}
        \end{subfigure}
    \end{subfigure}
    \caption[SVHN brightness thresholding overview]{\textbf{SVHN brightness thresholding overview.}  }
    \label{fig:svhn-brightness-data}
\end{figure}

\begin{figure}
    \centering
    \begin{subfigure}{0.45\textwidth}
        \includegraphics[width=0.9\textwidth]{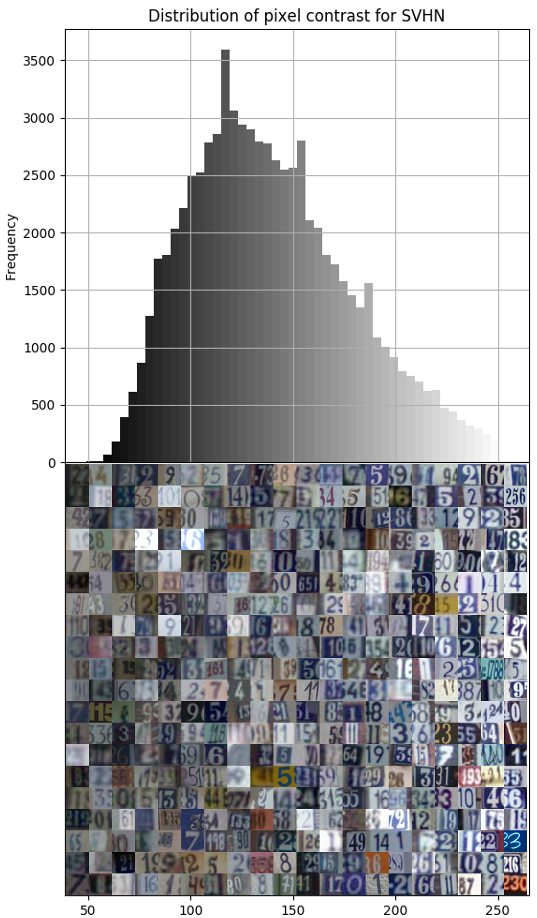}
        \caption{\textbf{SVHN contrast histogram.}  The histogram shows the distribution of pixel contrast for SVHN.  The images below the histogram correspond to the bins of the histogram, meaning samples to the left have low contrast whereas samples further to the right have higher contrast.}
        \label{fig:svhn-contrast-hist}
    \end{subfigure} \quad
    \begin{subfigure}[t!]{0.50\textwidth}
        \begin{subfigure}{\textwidth}
            \centering
            \includegraphics[width=0.8\textwidth]{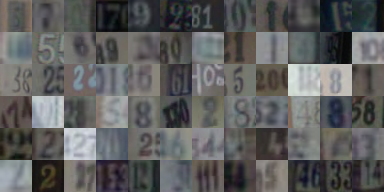}
            \caption{\textbf{Low contrast samples.}}
            \label{fig:svhn-low-contrast}
        \end{subfigure}\vspace{5pt}
        \begin{subfigure}{\textwidth}
            \centering
            \includegraphics[width=0.8\textwidth]{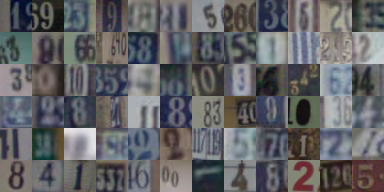}
            \caption{\textbf{Medium contrast samples.}}
            \label{fig:svhn-medium-contrast}
        \end{subfigure}\vspace{5pt}
        \begin{subfigure}{\textwidth}
            \centering
            \includegraphics[width=0.8\textwidth]{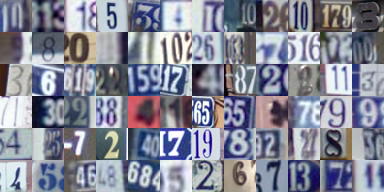}
            \caption{\textbf{High contrast samples.}}
            \label{fig:svhn-high-contrast}
        \end{subfigure}
    \end{subfigure}
    \caption[SVHN contrast thresholding overview]{\textbf{SVHN contrast thresholding overview.}  }
    \label{fig:svhn-contrast-data}
\end{figure}

Figure \ref{fig:svhn-brightness-data} shows a summary of the subsets of SVHN that we compiled corresponding to brightness.  In particular, Figure \ref{fig:svhn-brightness-hist} shows a histogram of the brightnesses of images in SVHN.  We used this histogram to set thresholds for low, medium, and high brightness, which are given in Table \ref{tab:svhn-gtsrb-nuisances}.  The images below the histogram correspond to the bins of the histogram; that is, images further to the left in Figure \ref{fig:svhn-brightness-hist} have lower brightness, whereas images further to the right have high brightness.  In Figures \ref{fig:svhn-low-brightness}, \ref{fig:svhn-medium-brightness}, and \ref{fig:svhn-high-brightness}, we show samples from the subsets of low, medium and high contrast subsets of SVHN that we compiled.  Figure \ref{fig:svhn-contrast-data} tells the same story as \ref{fig:svhn-brightness-data} for the contrast nuisances in SVHN.  Again, Figure \ref{fig:svhn-contrast-hist} shows a histogram and accompanying images corresponding to different values of contrast.  Figures \ref{fig:svhn-low-contrast}, \ref{fig:svhn-medium-contrast}, and \ref{fig:svhn-high-contrast} show samples from the subsets of low, medium, and high contrast images we compiled.

\begin{figure}
    \centering
    \begin{subfigure}{0.45\textwidth}
        \includegraphics[width=0.9\textwidth]{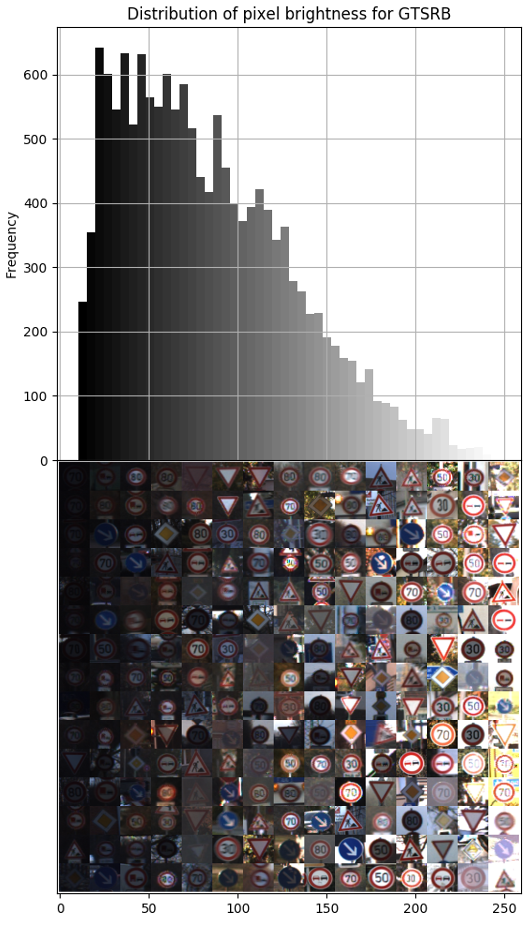}
        \caption{\textbf{GTSRB brightness histogram.}  The histogram shows the distribution of pixel brightness for GTSRB.  The images below the histogram correspond to the bins of the histogram, meaning samples to the left have low brightness whereas samples further to the right have higher brightness.}
        \label{fig:gtsrb-brightness-hist}
    \end{subfigure} \quad
    \begin{subfigure}[t!]{0.50\textwidth}
        \begin{subfigure}{\textwidth}
            \centering
            \includegraphics[width=0.8\textwidth]{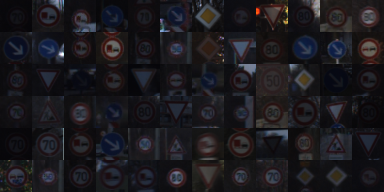}
            \caption{\textbf{Low brightness samples.}}
        \end{subfigure}\vspace{5pt}
        \begin{subfigure}{\textwidth}
            \centering
            \includegraphics[width=0.8\textwidth]{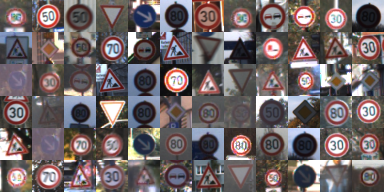}
            \caption{\textbf{Medium brightness samples.}}
            \label{fig:gtsrb-low-brightness}
        \end{subfigure}\vspace{5pt}
        \begin{subfigure}{\textwidth}
            \centering
            \includegraphics[width=0.8\textwidth]{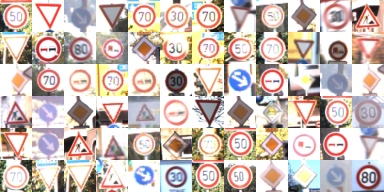}
            \caption{\textbf{High brightness samples.}}
            \label{fig:gtsrb-medium-brightness}
        \end{subfigure}
    \end{subfigure}
    \caption[GTSRB brightness thresholding overview]{\textbf{GTSRB brightness thresholding overview.}  }
    \label{fig:gtsrb-brightness-data}
\end{figure}

\begin{figure}
    \centering
    \begin{subfigure}{0.45\textwidth}
        \includegraphics[width=0.9\textwidth]{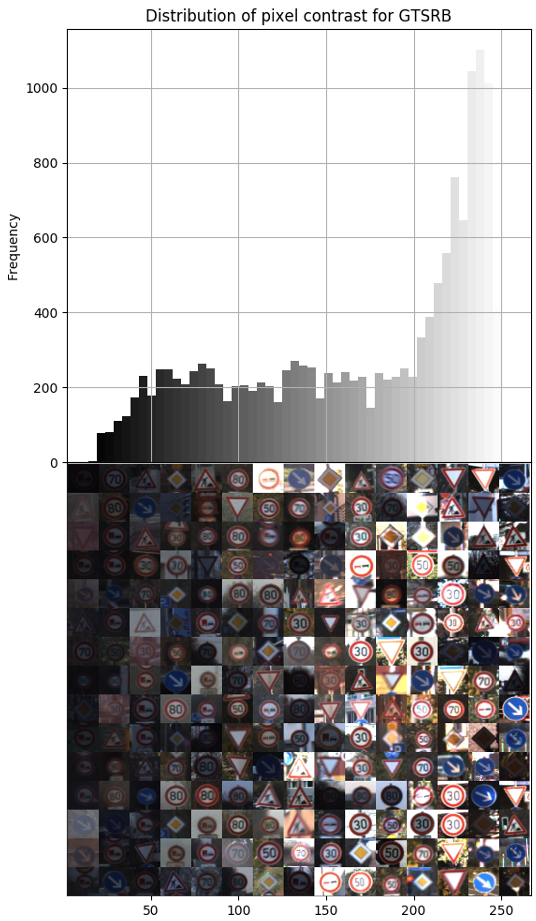}
        \caption{\textbf{SVHN contrast histogram.}  The histogram shows the distribution of pixel contrast for SVHN.  The images below the histogram correspond to the bins of the histogram, mening samples to the left have low contrast whereas samples further to the right have higher contrast.}
        \label{fig:gtsrb-contrast-hist}
    \end{subfigure} \quad
    \begin{subfigure}[t!]{0.50\textwidth}
        \begin{subfigure}{\textwidth}
            \centering
            \includegraphics[width=0.8\textwidth]{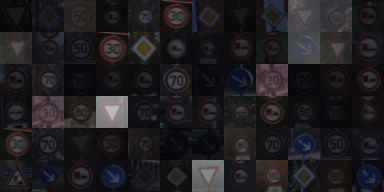}
            \caption{\textbf{Low contrast samples.}}
            \label{fig:gtsrb-low-contrast}
        \end{subfigure}\vspace{5pt}
        \begin{subfigure}{\textwidth}
            \centering
            \includegraphics[width=0.8\textwidth]{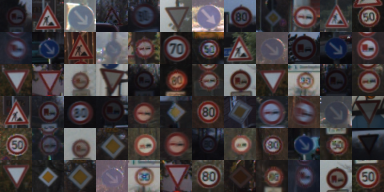}
            \caption{\textbf{Medium contrast samples.}}
            \label{fig:gtsrb-medium-contrast}
        \end{subfigure}\vspace{5pt}
        \begin{subfigure}{\textwidth}
            \centering
            \includegraphics[width=0.8\textwidth]{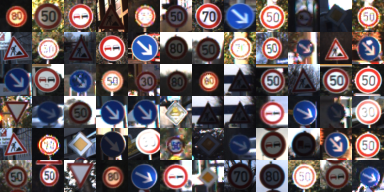}
            \caption{\textbf{High contrast samples.}}
            \label{fig:gtsrb-high-contrast}
        \end{subfigure}
    \end{subfigure}
    \caption[SVHN contrast thresholding overview]{\textbf{SVHN contrast thresholding overview.}  }
    \label{fig:gtsrb-contrast-data}
\end{figure}

We repeat this analysis for the brightness and contrast thresholding operations for GTSRB in Figures \ref{fig:gtsrb-brightness-data} and \ref{fig:gtsrb-contrast-data}.  Again, the difference between high- and low-brightness samples is remarkable, as is the difference in the samples corresponding to high- and low-contrast.  However, an interesting difference between the distributions of brightness and contrast on GTSRB vis-a-vis SVHN is that the distributions for GTSRB are skewed, whereas the distributions for SVHN are close to being symmetric. 

\begin{table}[]
    \centering
    \begin{tabular}{|c|c|c|c|c|c|c|}
        \hline
        \multirow{2}{*}{} & \multicolumn{3}{c|}{\bfseries SVHN} & \multicolumn{3}{c|}{\bfseries GTSRB}\\ \cline{2-7}
         & \thead{Low} & \thead{Medium} & \thead{High} & \thead{Low} & \thead{Medium} & \thead{High} \\ \hline
         Brightness & $\mathcal{B} < 60$ & $160 < \mathcal{B} < 170$ & $\mathcal{B} > 180$ & $\mathcal{B} < 40$ & $85 < \mathcal{B} < 125$ & $\mathcal{B} > 170$ \\ \hline
         
         Contrast & $\mathcal{C} < 80$ & $90 < \mathcal{C} < 100$ & $\mathcal{C} > 190$ & $\mathcal{C} < 80$ & $140 < \mathcal{C} < 200$ & $\mathcal{C} > 230$ \\ \hline
    \end{tabular}
    \caption[Thresholds for SVHN and GTSRB]{\textbf{Brightness and contrast thresholds.}  This table shows the thresholds we chose to represent low, medium, and high values of contrast and brightness for SVHN and GTSRB.}
    \label{tab:svhn-gtsrb-nuisances}
\end{table}


\subsubsection{Artificially-generated variation}

The remainder of the experiments, including those on MNIST, CURE-TSR, and ImageNet, use artifically-generated variation.  Indeed, one challenge in addressing deep learning's lack of robustness to natural variation is that relatively few datasets contain labeled forms of naturally-occurring sources of variation.  To this end, an important research challenge is to curate datasets with naturally-occurring variation; we plan to pursue this goal in future work.

When data with naturally-occurring variation is not available, artificially-generated variation can be used as an effective proxy for testing the robustness of deep learning against different forms of variation \cite{hendrycks2019benchmarking,hendrycks2020many}.  Indeed, the recently curated CURE-TSR \cite{temel2019traffic} and ImageNet-c \cite{hendrycks2019benchmarking} were created using pre-defined, artifical transformations of data.  While these transformations are synthetic, the images in Appendix \ref{app:gallery-of-learned-models} show that they are indeed quite realistic.

\subsection{Datasets introduced in this paper}

\newdataset{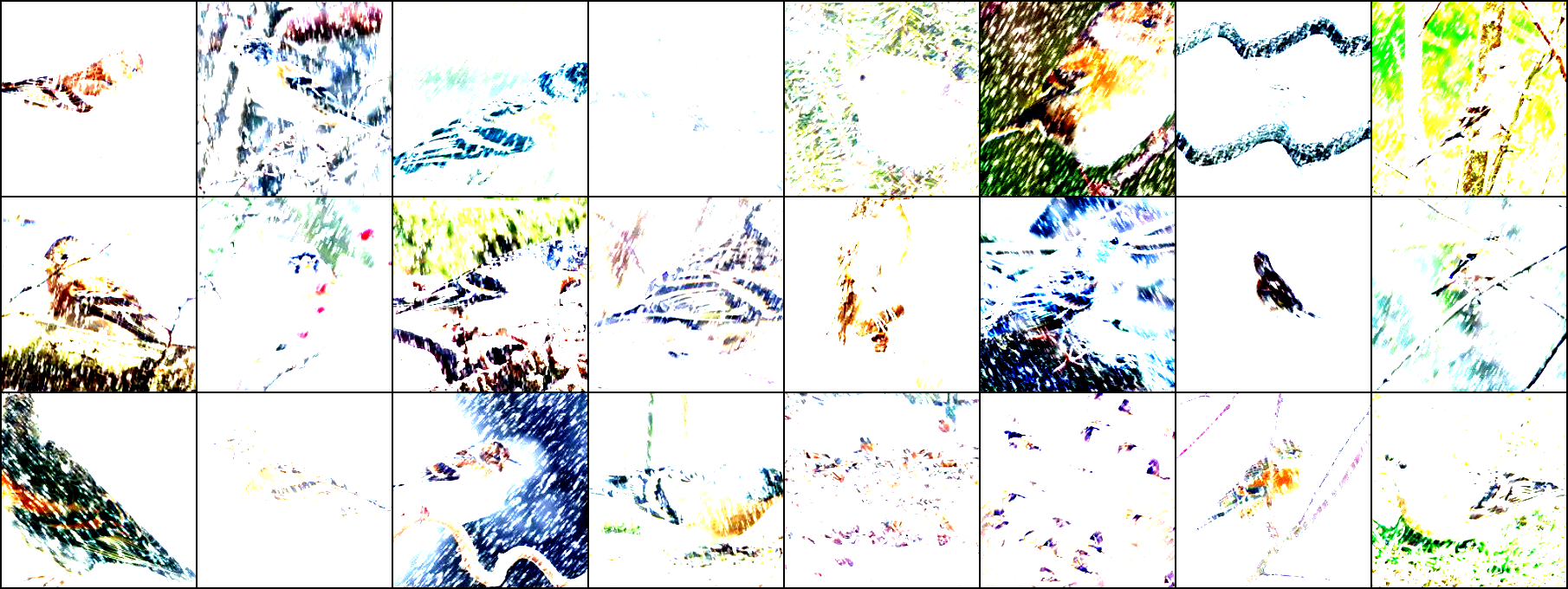}{\textbf{Brightness and snow.}  We use the challenge-level 1 transforms from ImageNet-c to generate an ImageNet test set with shifts in both brightness and snow.}{ImageNet dataset with brightness and snow}

\newdataset{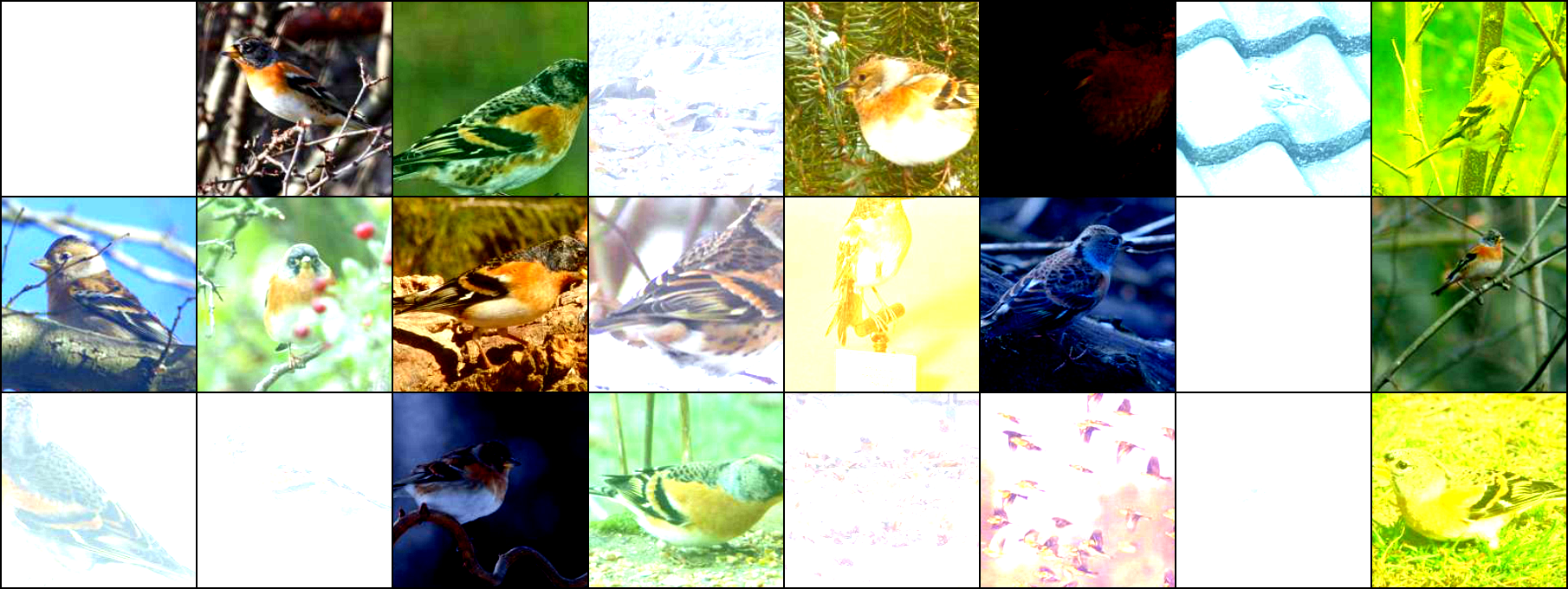}{\textbf{Brightness and contrast.}  We use the challenge-level 2 transforms from ImageNet-c to generate an ImageNet test set with shifts in both brightness and contrast.}{ImageNet dataset with brightness and contrast}

\newdataset{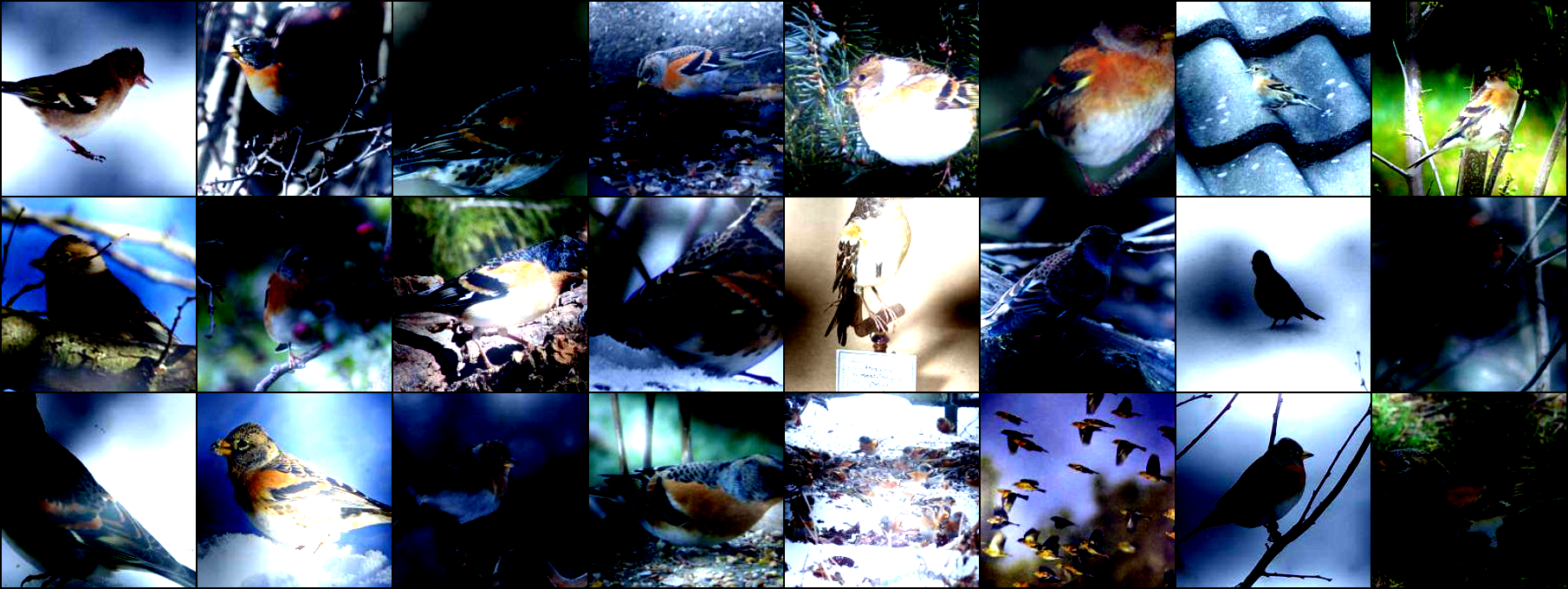}{\textbf{Brightness and fog.}  We use the challenge-level 1 transforms from ImageNet-c to generate an ImageNet test set with shifts in both brightness and fog.}{ImageNet dataset with brightness and fog}

\newdataset{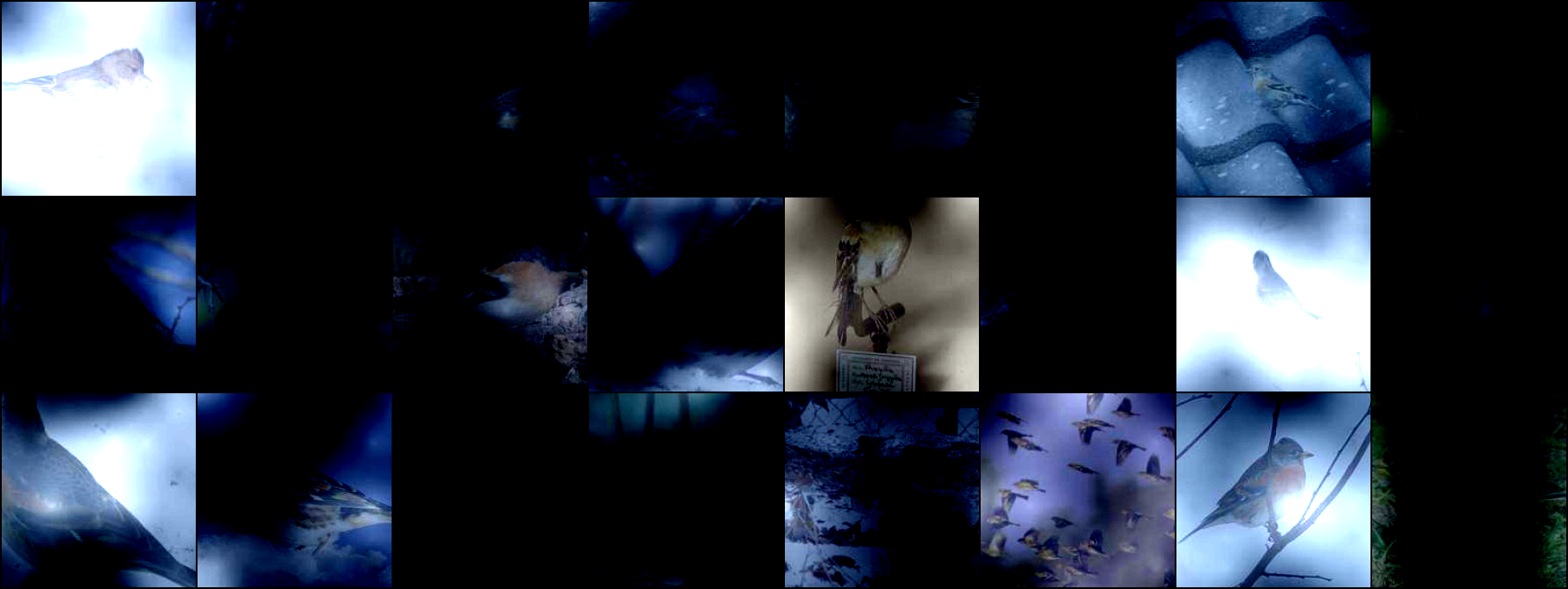}{\textbf{Contrast and fog.}  We use the challenge-level 1 transforms from ImageNet-c to generate an ImageNet test set with shifts in both contrast and fog.}{ImageNet dataset with contrast and fog}

In this paper, we introduced several new datasets which contain multiple simultaneous corruptions, including show, brightness, contrast, and fog.  In particular, we used the transforms used to create ImageNet-c to add multiple corruptions to the ImageNet test set.  To do so, we used the open-source code from \cite{hendrycks2019benchmarking}, which can be found at \url{https://github.com/hendrycks/robustness}.  Images of these datasets are shown in Figures 26-29.
\end{appendix}

\end{document}